\documentclass[journal]{IEEEtran}

\usepackage{cite}
\usepackage{graphicx}
\usepackage{amsmath}
\usepackage{amssymb}
\usepackage{multirow}
\usepackage{makecell}
\usepackage{booktabs}
\usepackage{subfigure}
\usepackage{url}
\usepackage{threeparttable}
\usepackage{verbatim}
\usepackage{todonotes}
\usepackage{algorithm}
\usepackage{algpseudocode}
\usepackage{amsmath}
\usepackage{verbatim}
\usepackage{color}
\usepackage{float}
\usepackage[colorlinks,
            linkcolor=blue,
            anchorcolor=blue,
            citecolor=blue]{hyperref}
\usepackage{booktabs}
\usepackage{multirow}
\usepackage[table]{xcolor}
\usepackage{colortbl}

\definecolor{rowgray}{gray}{0.96}
\definecolor{oursbg}{RGB}{235,242,250} 
\definecolor{bestbg}{RGB}{255,242,204}  
\definecolor{secondbg}{RGB}{226,239,249}

\ifCLASSINFOpdf
\else
\fi
\definecolor{gainblue}{RGB}{92,107,192} 
\definecolor{gainblueA}{RGB}{30,136,229}
\definecolor{gainblueB}{RGB}{51,153,255}
\definecolor{gainblueC}{RGB}{90,169,255}
\definecolor{gainblueD}{RGB}{102,178,255}

\begin{document}

\title{CoLR-Det: Collaborative Latent Restoration for Small Object Detection in Low-Resolution Remote Sensing Images}

\author{Ruo Qi, Linhui Dai$^\ast$, Yusong Qin, Chaolei Yang, Yanshan Li 
        
\thanks{This work was partially supported by National Natural Science Foundation of China (No.62471317), Natural Science Foundation of Shenzhen (No. JCYJ20240813141331042), Guangdong Provincial Key Laboratory (Grant 2023B1212060076), and the Scientific Foundation for Youth Scholars of Shenzhen University, China.
(\textit{$^\ast$Corresponding author: Linhui Dai}).
}
\thanks{Ruo Qi, Linhui Dai, Yusong Qin, Chaolei Yang, Yanshan Li
are with the Institute of Intelligent Information Processing, Shenzhen University, Guangdong Key Laboratory of Intelligent Information Processing, Shenzhen and Shenzhen Key Laboratory of Modern Communications and Information Processing, Shenzhen University, Shenzhen, China
518000, China (email: qiruo2023@email.szu.edu.cn; dailinhui@szu.edu.cn; 2450042012@mails.szu.edu.cn; yangchaolei2022@email.szu.edu.cn; lys@szu.edu.cn )}
}

\markboth{Journal of \LaTeX\ Class Files,~Vol.~14, No.~8, August~2021}%
{Shell \MakeLowercase{\textit{et al.}}: A Sample Article Using IEEEtran.cls for IEEE Journals}

\maketitle

\begin{abstract}
Remote sensing small object detection under low-resolution observations is fundamentally limited not only by missing visual details, but also by the ambiguity of how such details should be used for detection. Existing super-resolution-assisted detectors usually follow a restoration-first paradigm, where images, regions, or features are explicitly enhanced before being fed to a detector. However, this paradigm implicitly assumes that visually faithful reconstruction is beneficial for recognition, while in practice super-resolution favors dense texture recovery and edge fidelity, whereas object detection depends on sparse, instance-level semantic evidence. As a result, restoration may amplify background textures that are visually plausible but semantically irrelevant to small objects. In this paper, we propose CoLR-Det, a Collaborative Latent-Restoration-Assisted Small Object Detection framework that incorporates super-resolution supervision as detection-oriented latent regularization rather than explicit image-level enhancement. Instead of reconstructing a high-resolution image for inference, CoLR-Det uses a training-only restoration branch to impose auxiliary reconstruction constraints on a shared multiscale representation, while the inference pathway remains purely detection-driven. To ensure that restoration cues benefit object recognition rather than texture hallucination, we further introduce a saliency-guided object-preserving token routing mechanism, which prioritizes high-saliency tokens for attention-based refinement while bypassing the remaining tokens without permanently discarding their information. Moreover, we develop a detection-prioritized two-stage optimization strategy that first establishes stable object-level semantics and subsequently introduces restoration supervision. A smaller learning rate is assigned to the SR decoder to keep restoration-branch updates conservative and reduce abrupt perturbations during collaborative refinement.
Through this design, CoLR-Det turns restoration from an explicit visual enhancement operator into an implicit semantic regularizer for low-resolution small object detection. Experiments on resolution-degraded NWPU VHR-10-Split, DOTAv1.5-Split, and HRSSD-Split show that the proposed CoLR-Det outperforms state-of-the-art methods. Our code is available at \url{https://github.com/qiruo-ya/CoLR-Det}.
\end{abstract}


\section{INTRODUCTION}
\label{sec:Introduction}
\IEEEPARstart{S}{mall} object detection in remote sensing images is a fundamental yet challenging task in Earth observation, with broad applications in traffic monitoring, aircraft surveillance, and maritime monitoring. Although modern aerial and satellite sensors can provide high-resolution imagery, high image resolution does not necessarily guarantee sufficient object-level resolution. Many targets of interest still occupy only a small number of pixels due to large imaging distance, wide-area coverage, and limited ground sampling density. This issue becomes more severe under low-resolution (LR) or resolution-degraded inputs, where object-level structures such as boundaries, shapes, and discriminative textures are further weakened. As a result, small objects are easily submerged by homogeneous land-cover regions, repetitive man-made patterns, and cluttered background textures, making LR remote sensing small object detection particularly difficult~\cite{Nikouei2025, Gui2024}.

A straightforward way to compensate for the loss of object-level details is to introduce single-image super-resolution (SR)~\cite{Su2025,Lepcha2023,Wang2025,Alsaedi2025}. Existing SR-assisted detection methods aim to recover missing auxiliary reconstruction details for better object recognition, and have evolved from independent preprocessing to joint or selective optimization. Despite these efforts, most methods still exploit SR through explicit image- or region-level enhancement, where restored pixels, patches, or features are produced before detection. This restoration-centered interface is not necessarily detection-optimal. SR is usually driven by dense pixel-level fidelity or perceptual quality, whereas object detection relies on sparse, instance-level semantic evidence. Therefore, the central issue is not simply whether SR can recover more visual details, but whether the recovered cues are aligned with the evidence required by detection. This motivates a detection-centered principle: \emph{restoration should assist detection, rather than dominate it}.

\begin{figure}[!t]
    \centering
    \includegraphics[width=1.0\linewidth]{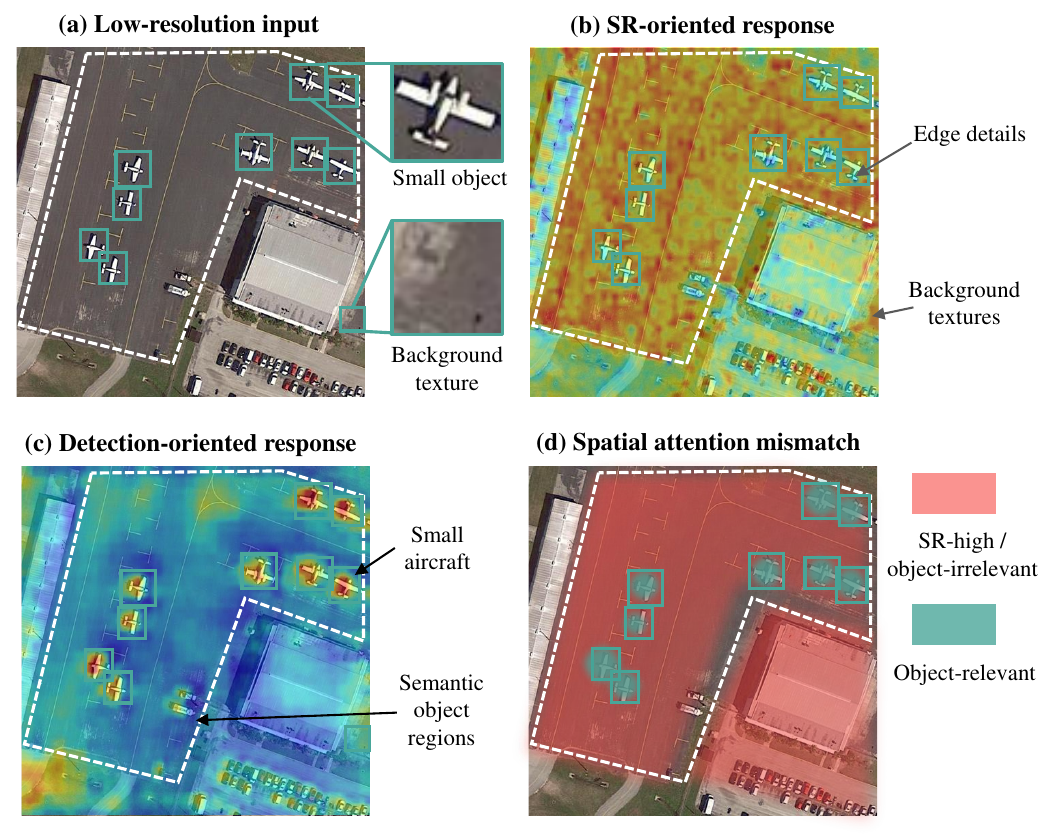}
    \caption{\textbf{Spatial response mismatch between super-resolution and detection.} (a) In resolution-degraded remote sensing inputs, small objects exhibit weak visual evidence while background textures remain prominent.
    (b) SR-oriented response map, generated from intermediate feature responses and overlaid on the input image, emphasizes texture-rich and edge-dominant regions, including pavement markings and structural boundaries. (c) Detection-oriented response map, generated from CAM-head responses, is sparse and object-centric, concentrating on target regions. (d) The mismatch between SR- and detection-oriented responses produces SR-high but object-irrelevant regions, which may cause background texture pollution when restoration cues are injected into detection features. The cyan boxes are manually marked to roughly indicate object locations, and the dashed and colored regions are manually highlighted for visual illustration.}
    \label{fig:activation_maps}
\end{figure}

Fig.~\ref{fig:activation_maps} provides visual evidence for this restoration-detection mismatch. Under resolution degradation, small objects become weak and blurred, whereas background textures remain visually prominent (Fig.~\ref{fig:activation_maps}(a)). SR supervision tends to emphasize texture-rich background structures and structural edges (Fig.~\ref{fig:activation_maps}(b)), while detection requires sparse object-centric responses around target regions (Fig.~\ref{fig:activation_maps}(c)). The discrepancy produces SR-high but object-irrelevant regions, indicating that restoration cues are not uniformly beneficial for detection (Fig.~\ref{fig:activation_maps}(d)). This observation exposes a key conflict: dense texture reconstruction may improve visual fidelity, but it can also introduce texture-biased interference into detection-oriented representations.

These observations lead to a central question: \emph{how can restoration cues assist LR remote sensing small object detection without dominating the detection objective?}
To answer this question, three coupled challenges must be addressed. First, restoration should assist detection without becoming an additional image-level inference stage. This suggests that SR cues should be used as latent supervision during training rather than as explicit preprocessing during inference. Second, restoration cues are dense and texture-oriented, while detection cues are sparse and object-oriented. Therefore, the model should selectively preserve object-relevant restoration cues and suppress background-dominated responses. Third, the tasks have different optimization preferences; dense pixel-level reconstruction supervision may interfere with the sparse, localized cues required for semantic abstraction, necessitating a mechanism to improve SR--detection compatibility during joint learning.

To address these challenges, we propose \textbf{CoLR-Det}, a \textbf{Co}llaborative \textbf{L}atent-\textbf{R}estoration Assisted Small Object \textbf{Det}ection framework for remote sensing images. CoLR-Det reformulates SR-assisted detection from explicit image reconstruction to latent restoration assistance. First, a training-only restoration branch is attached to a parameter-shared multiscale encoder to inject auxiliary reconstruction cues into the latent feature space. During inference, this branch is removed, and detection is performed directly from the learned latent representation without explicit SR reconstruction. Second, to prevent background textures from contaminating object semantics, CoLR-Det introduces a saliency-guided object-preserving token routing mechanism. Considering that learned saliency can be unreliable in early training, the proposed routing is designed to be non-destructive. Instead of permanently discarding low-saliency tokens, CoLR-Det bypasses them from expensive self-attention layers while retaining their information flow, reducing the risk of missing weak small objects. Third, we design a detection-prioritized optimization strategy. The detector is first optimized to establish stable object-level semantic representations, after which restoration supervision is introduced. A smaller learning rate is assigned to the SR decoder to keep its parameter updates conservative and avoid abrupt perturbations during collaborative refinement.

Our contributions are summarized as follows:
\begin{enumerate}
    \item We propose CoLR-Det, a novel collaborative latent-restoration framework that reformulates SR-assisted remote sensing detection as latent regularization rather than explicit SR preprocessing, enhancing detection-oriented representations without requiring explicit SR reconstruction at inference.
    \item We introduce an object-preserving collaboration mechanism that pairs a training-only restoration branch with a non-destructive token routing module. By bypassing rather than discarding low-saliency tokens, this mechanism suppresses background-dominated responses while preserving potential cues for weak small objects.
    \item We design a detection-prioritized two-stage optimization strategy with decoupled learning-rate scaling for the restoration branch. This strategy stabilizes object-level semantic representations before incorporating conservative restoration updates, reducing abrupt perturbations during collaborative refinement.
    \item Extensive experiments on the NWPU VHR-10-Split~\cite{Cheng2016}, DOTAv1.5-Split~\cite{DOTA2018}, and HRSSD-Split datasets~\cite{Zhang2019} demonstrate that CoLR-Det achieves state-of-the-art performance, achieving substantial accuracy gains for low-resolution small objects.
\end{enumerate}

\section{RELATED WORK}
\label{sec:Related Work}

\subsection{Remote Sensing Small Object Detection}

Image super-resolution (SR) is important for small object detection by reconstructing fine details from low-resolution (LR) inputs and improving feature discriminability for downstream recognition tasks. By architecture, SR methods are grouped into CNN-based and Transformer-based approaches.

CNN-based SR methods, following the seminal SRCNN~\cite{Dong2016}, have substantially advanced image reconstruction~\cite{Lei2017,Jiang2018a,Jiang2019,Xiao2023}. They often improve feature representation with attention mechanisms. Channel attention~\cite{Zhang2018} and holistic attention~\cite{Niu2020} enhance feature selection, whereas self-similarity attention~\cite{Lei2022} and non-local sparse attention~\cite{Mei2021} introduce global dependencies for context modeling. Although non-local modeling can ease local receptive-field constraints, it often incurs high computation, leaving a trade-off between global semantic modeling and reconstruction efficiency in remote sensing scenarios.

Transformer-based SR methods exploit self-attention for long-range dependency modeling and effective feature representation~\cite{Kim2023,He2022,Xiao2023b}. Liang et al.~\cite{Liang2021} proposed SwinIR, combining hierarchical feature extraction with reconstruction modules for super-resolution, denoising, and compression artifact removal. Xiao et al.~\cite{Xiao2024} designed TTST with dynamic token selection and multiscale feed-forward layers, reducing redundant computation while improving reconstruction accuracy. Kang et al.~\cite{Kang2024} proposed ESTNet, reducing computational complexity through efficient channel attention and residual group mechanisms while maintaining reconstruction performance. Overall, Transformer architectures balance global feature modeling and efficiency, offering useful directions for remote sensing image super-resolution.

Traditional CNN-based object detectors include two-stage and one-stage approaches. Two-stage methods, such as Faster R-CNN~\cite{Ren2015}, achieve high accuracy via region proposal generation and classification refinement, whereas one-stage methods, such as the YOLO series~\cite{Redmon2017}, directly regress bounding boxes for faster inference. In remote sensing images, small object sizes, arbitrary orientations, and complex backgrounds still limit their performance.

Numerous CNN-based frameworks address remote sensing object detection. Some methods enhance feature representation and multiscale fusion, such as MS-OPN~\cite{MSOPN2019} and FSANet~\cite{FSANet2023}, which improve multi-level feature extraction and alignment to capture small, densely distributed objects. Others use context modeling and attention mechanisms to suppress background interference and enhance discriminative regions, including FE-CenterNet~\cite{FECenterNet2023} and FFCA-YOLO~\cite{FFCAYOLO2024}. Meanwhile, lightweight detectors such as LSODNet~\cite{LSODNet2024} optimize efficiency via depthwise and deformable convolutions. In this context, LEGNet~\cite{LEGNet2025} proposes a lightweight backbone incorporating edge priors and Gaussian-based modeling to improve feature robustness in LR remote sensing images.

Recently, Transformer-based detectors attract increasing attention for their global modeling capability. PR-Deformable DETR~\cite{PRDETR2024} combines adaptive feature fusion with deformable attention for multiscale and rotation-robust detection, achieving state-of-the-art results on several benchmark datasets. Drone-DETR~\cite{DroneDETR2024} further optimizes the Transformer architecture for UAV onboard applications and balances accuracy with real-time inference efficiency. Despite these advances, small object detection in remote sensing remains challenging, especially under low resolution and limited computational resources. Therefore, lightweight, context-aware, and saliency-driven detection frameworks remain important for improving robustness and efficiency.

\subsection{SR-Assisted Remote Sensing Object Detection}

Recent studies on joint SR and object detection in remote sensing images follow three paradigms: stepwise learning~\cite{Shermeyer2019,Ji2019}, joint optimization~\cite{Rabbi2020,Kim2024}, and selective optimization~\cite{Wu2021}. These methods aim to improve image quality and detection performance under weak feature responses and complex backgrounds.

Stepwise learning treats SR as independent preprocessing before object detection in remote sensing images~\cite{Shermeyer2019,Ji2019}. This approach is simple to train but lacks closed-loop optimization between SR and detection. Consequently, SR may improve visual clarity without improving detection performance and may introduce artifacts such as pseudo-textures that degrade detection results.

Joint optimization strategies integrate super-resolution and object detection into a unified end-to-end framework~\cite{Rabbi2020,Kim2024}. Unlike independent stepwise training, this paradigm constructs a serial SR-detection pipeline, allowing gradients from detection loss to propagate back to the upstream SR module. Consequently, image reconstruction is guided by both pixel fidelity and downstream semantic needs, encouraging object-relevant detail recovery. However, the intrinsic conflict between pixel-level fidelity and semantic-level invariance can cause gradient inconsistency, hindering convergence and limiting overall performance improvement.

Selective optimization applies super-resolution only to key regions through region extraction or attention mechanisms, balancing efficiency and performance. HSOD-Net, proposed by Wu et al.~\cite{Wu2021}, focuses super-resolution on potential object regions predicted by keypoints, achieving effective collaborative optimization of image enhancement and object detection. However, HSOD-Net still follows a stepwise pipeline: region selection is predicted on low-resolution features and then used to crop patches for subsequent super-resolution. Once first-stage keypoint prediction misses or mislocalizes an object, the subsequent super-resolution module can no longer compensate for this error, causing irreversible detection failures. Moreover, hard region cropping and loosely coupled feature sharing make the interaction between enhancement and detection relatively shallow, so detection loss provides only indirect feedback to the super-resolution branch.

Overall, existing methods achieve different degrees of collaborative optimization between SR and object detection, but still suffer from inefficient feature sharing and optimization conflicts. Enabling tighter detection-driven collaboration between SR and object detection within an end-to-end framework, while improving small object detection in remote sensing images and maintaining computational efficiency, remains an important research direction.

 \begin{figure*}[htbp] 
 \centering 
 \includegraphics[width=1.0\textwidth]{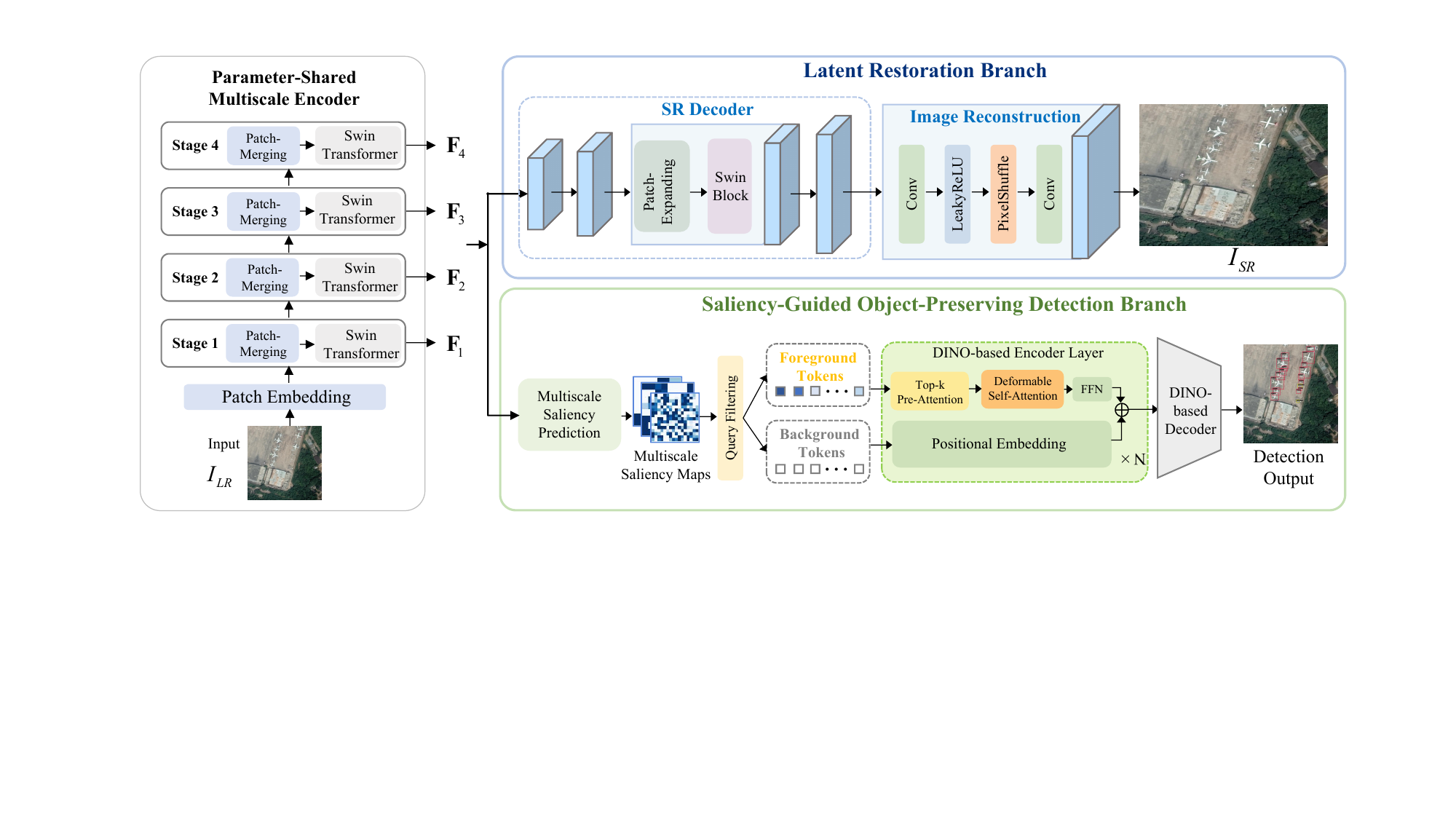} 
 \caption{ \textbf{Overall architecture of CoLR-Det.} Given an LR image, a parameter-shared multiscale encoder extracts latent features for both detection and training-time restoration supervision. The restoration branch injects auxiliary reconstruction cues during training and is removed during inference. The detection branch predicts saliency responses and performs object-preserving token routing before DINO-style decoding. CoLR-Det is optimized with a detection-prioritized two-stage strategy, where object-level semantics are first stabilized and restoration supervision is later incorporated under a decoupled learning-rate schedule.
} 
\label{fig:CoLR-Det_structure} 
\end{figure*}

\section{METHODS}
\label{sec:methods}

\subsection{Overview}

Guided by the principle that \emph{restoration should assist detection rather than dominate it}, we propose CoLR-Det, a Collaborative Latent-Restoration Assisted Small Object Detection framework for remote sensing images. Different from explicit SR-before-detection pipelines, CoLR-Det does not use super-resolution as an image-level preprocessing step. Instead, restoration is introduced only as training-time latent regularization to enhance detection-oriented representations under LR observations.

As shown in Fig.~\ref{fig:CoLR-Det_structure}, CoLR-Det consists of three components: a parameter-shared multiscale encoder, a training-only latent restoration branch, and a saliency-guided object-preserving detection branch. Given an LR image $I_{\mathrm{LR}}$, the shared encoder extracts multiscale latent features $\{F_s\}_{s=1}^{4}$. During training, the restoration branch reconstructs an auxiliary high-resolution image $I_{\mathrm{SR}}$ from the shared features and provides auxiliary reconstruction supervision to the encoder. During inference, this branch is removed, and detection is performed directly from the learned latent representation without explicit SR reconstruction.

The detection branch is built upon a DINO-style detector. To prevent restoration-enhanced background textures from contaminating object semantics, CoLR-Det predicts multiscale saliency responses and performs object-preserving token routing before Transformer decoding. Foreground-relevant tokens are emphasized, while background-dominated tokens are bypassed from expensive attention computation rather than permanently discarded. This non-destructive design reduces texture-biased interference while retaining potential cues for weak small objects.

Finally, CoLR-Det is trained with a detection-prioritized two-stage optimization strategy. The first stage optimizes the encoder and detector to establish stable object-level semantic representations. The second stage introduces restoration supervision under a decoupled learning-rate schedule, enabling collaborative refinement while keeping restoration-branch updates conservative.

\subsection{Parameter-Shared Multiscale Encoder}

We construct a cross-task parameter-shared encoder based on the Swin Transformer Tiny (Swin-T) architecture~\cite{Liu2021}. As illustrated in the left part of Fig.~\ref{fig:CoLR-Det_structure}, the encoder contains four hierarchical stages and outputs multiscale features for subsequent SR reconstruction and object detection. The inset details the Swin block, where window-based multi-head self-attention (W-MSA) and shifted-window multi-head self-attention (SW-MSA) are alternately combined with layer normalization and MLP layers. This encoder provides a common latent space where restoration supervision can regularize detection-oriented representations during training. Given an input low-resolution image $\mathbf{I}_{\text{LR}} \in \mathbb{R}^{H \times W \times 3}$, where $H$ and $W$ denote the spatial dimensions, the shared encoder processes it through four progressive stages to generate a multiscale feature pyramid. Specifically, at stage $s$ ($s \in \{1, 2, 3, 4\}$), the encoder outputs feature maps $\mathbf{F}_s \in \mathbb{R}^{H_s \times W_s \times C_s}$, where the spatial resolution is downsampled by a factor of $2^{s+1}$ relative to the input image, and the channel dimensions are $C_s \in \{96, 192, 384, 768\}$, respectively. The progressive feature extraction process can be formulated as:
\begin{equation}
\mathbf{F}_s = \text{SwinBlock}_s(\mathbf{F}_{s-1}), \quad s \in \{1, 2, 3, 4\},
\end{equation}
where $\mathbf{F}_0 = \mathbf{I}_{\text{LR}}$ represents the input image, and $\text{SwinBlock}_s(\cdot)$ denotes the $s$-th stage Swin block that consists of patch merging operations followed by stacked Swin Transformer layers.
During training, the encoder parameters $\theta_{\text{enc}}$ simultaneously receive backpropagated gradients from both the SR reconstruction loss $\mathcal{L}_{\text{SR}}$ and the detection losses ($\mathcal{L}_{\text{cls}}$, $\mathcal{L}_{\text{bbox}}$, $\mathcal{L}_{\text{giou}}$, $\mathcal{L}_{\text{sa}}$). The detailed formulation and optimization strategy for these loss functions are presented in Section~\ref{sec:optimization}.

 \begin{figure}[!t]     
 \centering     
 \includegraphics[width=1.0\linewidth]{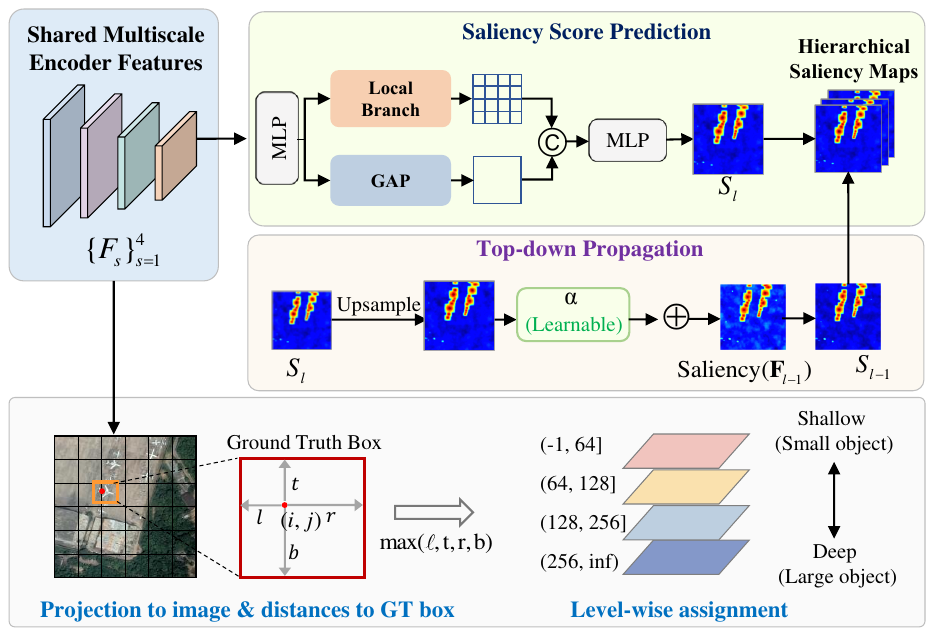}     
 \caption{ \textbf{Hierarchical saliency prediction and supervision framework.} Multiscale features are integrated with local and global contextual information to generate hierarchical saliency maps, while scale-aware supervision encourages object-centered responses across different feature levels. }     \label{fig:Saliency_maps} 
 \end{figure} 

\subsection{Latent Restoration Branch}

The super-resolution decoder adopts a hierarchical reconstruction design, as shown in the upper-right panel of Fig.~\ref{fig:CoLR-Det_structure}, that combines the progressive upsampling path of U-Net~\cite{UNet2015} with the feature modeling capability of the Swin Transformer. To avoid ambiguity in subsequent formulations, the input feature of a specific module is denoted as $\mathbf{Z}_{in}$. The decoder consists of cascaded patch-expanding modules and residual Swin blocks. Each patch-expanding module restores spatial resolution through transposed convolution and reshapes feature dimensions for fine-grained recovery:
\begin{equation}
\mathbf{Z}_{\text{expand}} =
\mathcal{R}_{\text{out}}\!\Big(
    \operatorname{TransConv}\!\Big(
        \operatorname{LN}\!\Big(
            \mathcal{R}_{\text{in}}\!\left(\mathbf{Z}_{\text{in}}\right)
        \Big)
    \Big)
\Big),
\end{equation}
where $\mathbf{Z}_{\text{in}}$ represents the input feature map from the preceding level, and $\mathbf{Z}_{\text{expand}}$ is the upsampled output feature. $\mathcal{R}_{\text{in}}(\cdot)$ rearranges the input tensor from the spatial layout to a sequence format compatible with the transposed convolution. $LN(\cdot)$ denotes layer normalization applied along the channel dimension. $\operatorname{TransConv}(\cdot)$ represents a transposed convolution with stride 2, which doubles the spatial resolution. Finally, $\mathcal{R}_{\text{out}}(\cdot)$ restores the expanded feature tensor back to the standard spatial layout for subsequent Swin Blocks.

To enhance feature representation, each Swin Transformer block adopts a pre-norm residual formulation that sequentially applies window-based multi-head self-attention ($\operatorname{Attn}$) and a multi-layer perceptron (MLP). The processing flow for a feature token sequence $\mathbf{T}_{\text{in}}$ is defined as:
\begin{equation}
\begin{aligned}
\mathbf{T}' &= \mathbf{T}_{\text{in}} + 
\operatorname{Attn}\!\left(\operatorname{LN}\!\left(\mathbf{T}_{\text{in}}\right)\right), \\
\mathbf{T}_{\text{out}} &= \mathbf{T}' + 
\operatorname{MLP}\!\left(\operatorname{LN}\!\left(\mathbf{T}'\right)\right).
\end{aligned}
\end{equation}
where $\mathbf{T}'$ denotes the intermediate token sequence after the attention residual connection.

After progressive multi-stage upsampling, the decoder produces the final decoded feature map $\mathbf{F}_{\text{dec}}$. The image reconstruction module then maps these deep features to the image domain. It uses lightweight convolutions with LeakyReLU ($\sigma$) for nonlinear mapping and a PixelShuffle layer for upsampling. The final high-resolution output $I_{SR}$ is obtained as follows:
\begin{equation}
I_{SR}=\operatorname{Conv}\left(\operatorname{PixelShuffle}\left(\sigma\left(\operatorname{Conv}\left(\mathbf{F}_{\text{dec}}\right)\right)\right)\right),
\end{equation}
where $\mathbf{F}_{\text{dec}}$ denotes the final output features of the decoder. During training, this auxiliary reconstruction branch provides latent restoration supervision, encouraging the shared encoder to preserve restoration-related structural cues that are beneficial for identifying small objects under resolution degradation. During inference, the restoration branch is discarded, and detection is performed directly from the shared latent features.

\subsection{Saliency-Guided Object-Preserving Token Routing}

Although DINO provides effective denoising training and fast convergence, DETR-style encoders still process a large number of spatial tokens, many of which correspond to background-dominated regions in high-resolution remote sensing features. Under latent restoration supervision, these background tokens may carry texture-biased responses and interfere with object-centric decoding. To address this issue, we introduce a saliency-guided object-preserving token routing strategy, as illustrated by the Saliency-Guided Detection Branch in Fig.~\ref{fig:CoLR-Det_structure}. The detection branch first predicts multiscale saliency maps by integrating local details and global context, and then ranks encoder tokens according to their saliency scores. High-ranked tokens are routed to the attention-based refinement path, whereas the remaining low-ranked tokens bypass the computationally intensive attention operations while continuing through the feature stream. This mechanism acts as a spatial attention prior, allocating more computation to potentially object-relevant regions. Here, “object-preserving” refers to the non-destructive routing property rather than a guarantee that every object token is selected into the active set. Low-ranked tokens are bypassed from attention-based refinement but remain in the feature stream, thereby avoiding irreversible information loss.

The extracted shared multiscale features are denoted as $\left\{\mathbf{F}_l\right\}_{l=1}^L$, where $\mathbf{F}_l$ is the feature map of the $l$-th layer with a specific spatial resolution and number of channels. These features contain rich semantic information for multiscale object detection.

 \begin{figure*}[!t]     
 \centering     
 \includegraphics[width=0.86\textwidth]{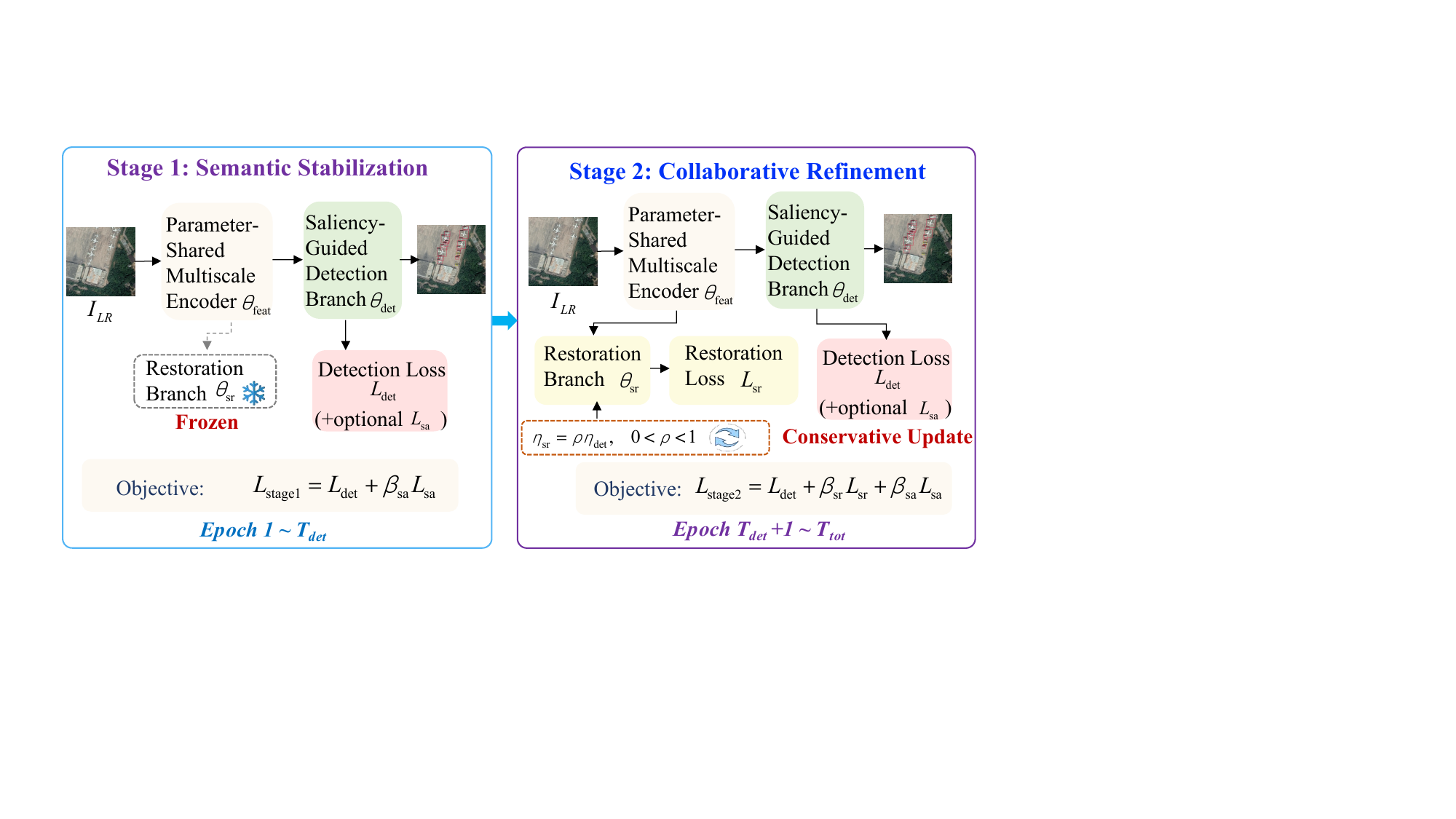}   
 \caption{ \textbf{Detection-prioritized two-stage optimization strategy.} The first stage establishes stable detection-oriented representations, whereas the second stage introduces restoration supervision for collaborative feature refinement. }     \label{fig:optimization} 
 \end{figure*}
 
\subsubsection{Saliency Score Prediction}

To model spatial importance, we design a lightweight prediction module that estimates a saliency map from the feature pyramid, as illustrated in the upper part of Fig.~\ref{fig:Saliency_maps}. Because saliency depends on both local contrast and global context, the input features are split along the channel dimension. The local branch preserves spatial details, whereas the global branch uses global average pooling and broadcasting to capture scene-level context. The two branches are concatenated and fused by a dimensionality-reducing MLP to produce a single-channel saliency score map $S_l$. The prediction process is defined as:
\begin{equation}
S_l = \text{MLP}_2\left(\mathcal{F}_{\text{fuse}}\left(\text{MLP}_1(\mathbf{F}_l)\right)\right),
\end{equation}
where $\mathcal{F}_{\text{fuse}}(\cdot)$ denotes the feature fusion operation that concatenates the local branch features $\mathbf{F}_{\text{local}}$ and the globally pooled-and-broadcasted features $\mathbf{F}_{\text{global}}$ along the channel dimension.

To ensure multiscale consistency, we implement a top-down propagation pathway. Saliency priors from coarse, high-level features are upsampled and fused with finer, low-level predictions. For the $l-1$-th layer, the refined saliency map is:
\begin{equation}
S_{l-1} = \alpha \cdot \text{UP}(S_l) + \text{Saliency}(\mathbf{F}_{l-1}),
\label{eq:saliency_propagation}
\end{equation}
where $\alpha$ is a learnable modulation coefficient that dynamically weights the contribution of high-level semantics.

 \begin{algorithm}[t] \caption{Detection-Prioritized Two-Stage Optimization} \label{alg:optimization} \begin{algorithmic}[1] \Require Training set $\mathcal{D} = \{(x_n, y_n)\}_{n=1}^N$; shared multi-level feature encoder parameters $\theta_{\text{feat}}$; detection Transformer encoder-decoder and head parameters $\theta_{\text{det}}$; SR decoder parameters $\theta_{\text{sr}}$; Stage-One learning rates $\eta_{\text{feat}}^{(1)}, \eta_{\text{det}}^{(1)}$; Stage-Two learning rates $\eta_{\text{feat}}^{(2)}, \eta_{\text{det}}^{(2)}, \eta_{\text{sr}}^{(2)}$; scaling factor $\rho$; detection pretraining epochs $T_{\text{det}}$; total epochs $T_{\text{tot}}$.
\Ensure Optimized parameters $\theta_{\text{feat}}, \theta_{\text{det}}, \theta_{\text{sr}}$.

\State \textbf{Notation:} 
$\mathcal{L}_{\text{det}} \gets 
\beta_{\text{cls}} \mathcal{L}_{\text{cls}} +
\beta_{\text{bbox}} \mathcal{L}_{\text{bbox}} +
\beta_{\text{giou}} \mathcal{L}_{\text{giou}}$, 
i.e., the standard detection loss excluding the saliency constraint.

\State \textbf{// Stage One: Semantic Stabilization}
\State Set $\text{requires\_grad}(\theta_{\text{sr}}) \leftarrow \text{False}$
\For{$t = 1$ to $T_{\text{det}}$}
    \State Sample mini-batch $\mathcal{B} \subset \mathcal{D}$
    \State Construct low-resolution inputs and forward through shared encoder and detection Transformer encoder-decoder
    \State Compute $\mathcal{L}_{\text{det}}$ and $\mathcal{L}_{\text{sa}}$ on $\mathcal{B}$
\State $\mathcal{L}_{\text{stage1}} \gets 
\mathcal{L}_{\text{det}} + 
\beta_{\text{sa}} \mathcal{L}_{\text{sa}}$
    \State Update $\theta_{\text{feat}}$ with $\eta_{\text{feat}}^{(1)}$ and $\theta_{\text{det}}$ with $\eta_{\text{det}}^{(1)}$ by back-propagating $\mathcal{L}_{\text{stage1}}$
\EndFor

\State \textbf{// Stage Two: Collaborative Refinement}
\State Set $\text{requires\_grad}(\theta_{\text{sr}}) \leftarrow \text{True}$
\State Set $\eta_{\text{sr}}^{(2)} \gets \rho \cdot \eta_{\text{det}}^{(2)}$, with $\rho < 1$
\For{$t = T_{\text{det}}+1$ to $T_{\text{tot}}$}
    \State Sample mini-batch $\mathcal{B} \subset \mathcal{D}$
    \State Forward through the shared encoder, the detection Transformer encoder-decoder, and the SR decoder
    \State Compute $\mathcal{L}_{\text{det}}$, $\mathcal{L}_{\text{sa}}$, and $\mathcal{L}_{\text{sr}}$ on $\mathcal{B}$
\State $\mathcal{L}_{\text{stage2}} \gets 
\beta_{\text{sr}} \mathcal{L}_{\text{sr}} + 
\mathcal{L}_{\text{det}} + 
\beta_{\text{sa}} \mathcal{L}_{\text{sa}}$
    \State Update $\theta_{\text{feat}}$ with $\eta_{\text{feat}}^{(2)}$,
           $\theta_{\text{det}}$ with $\eta_{\text{det}}^{(2)}$,
           and $\theta_{\text{sr}}$ with $\eta_{\text{sr}}^{(2)}$
           by back-propagating $\mathcal{L}_{\text{stage2}}$
\EndFor \end{algorithmic} 
 \end{algorithm}

\subsubsection{Saliency Confidence Calculation Based on Relative Distance}

Predicting only a binary foreground-background mask is insufficient for dense small object detection because such a mask does not provide gradient information about object centrality. Therefore, as illustrated in the lower part of Fig.~\ref{fig:Saliency_maps}, we design a scale-agnostic saliency confidence metric as a soft supervisory signal to measure the spatial alignment between a query position $(i,j)$ and the ground truth.

For a query located at $(i,j)$ in the $l$-th feature map, we map it back to the input image via global coordinates and compute its distances to the four borders of its enclosing ground-truth box $B_{GT}$, denoted as $\ell$, $t$, $r$, and $b$ for the left, top, right, and bottom sides, respectively. Considering the correspondence between feature stride and receptive field, each query is only assigned to objects whose $\max(\ell, t, r, b)$ falls within a predefined range $(\tau_{l-1}, \tau_l]$, which routes small objects to shallow layers and large objects to deeper layers, eliminating cross-scale interference at the supervision stage. Upon this assignment, the centrality of the query is jointly characterized by the normalized border-imbalance ratios $\delta_x = (\ell - r)/(\ell + r)$ and $\delta_y = (t - b)/(t + b)$, both of which vanish at the object center and approach $\pm 1$ near the borders, irrespective of object size. The saliency confidence is then defined as:
\begin{equation}
C_l^{(i,j)} =
\begin{cases}
1 - \dfrac{1}{2}\sqrt{\delta_x^{2} + \delta_y^{2}}, 
& \text{if } (i,j) \in B_{\text{GT}} \cap \Omega_l, \\
0, & \text{otherwise},
\end{cases}
\label{eq:saliency_confidence}
\end{equation}
where $\Omega_l$ denotes the set of query positions that satisfy the level-wise range constraint. When a query simultaneously falls inside multiple overlapping boxes, the maximum confidence across all candidates is retained to preserve the dominant object response.

Based on the predicted saliency map and the scale-aware confidence target, the saliency constraint loss is formulated as:
\begin{equation} 
\mathcal{L}_{\text{sa}} = \frac{1}{\max(N_{\text{pos}},1)} \sum_{l=1}^{L}\sum_{(i,j)} \operatorname{FL}\!\left(S_l^{(i,j)},\, C_l^{(i,j)}\right), 
\label{eq:loss_saliency} 
\end{equation}
where $L$ denotes the number of feature levels, and $(i,j)$ indexes a spatial query position on the $l$-th feature map. $S_l^{(i,j)}$ represents the predicted saliency score at position $(i,j)$, while $C_l^{(i,j)}$ is the corresponding soft saliency confidence target defined in Eq.~\eqref{eq:saliency_confidence}. $\operatorname{FL}(\cdot,\cdot)$ denotes the sigmoid focal loss with soft targets, where the sigmoid activation is applied internally. $N_{\text{pos}} = \sum_{l=1}^{L}\left|\{(i,j) : C_l^{(i,j)} > 0\}\right|$ is the total number of positive samples across all feature levels. We use $\max(N_{\text{pos}},1)$ for normalization to avoid division by zero when no positive sample is present in a mini-batch.

\subsubsection{Hierarchical Query Routing and Refinement}

Leveraging the predicted saliency scores, we implement a hierarchical query routing mechanism to optimize the computational budget allocated to the Transformer encoder. As introduced in the motivation, processing the entirety of high-resolution remote sensing features is redundant. Therefore, we dynamically select the top $\beta_t \gamma_l N$ queries based on their saliency ranking to form the active set $\Phi_t$, while routing the remaining tokens through a bypass path:
\begin{equation}
q_i =
\begin{cases}
\mathcal{A}(q_i + \text{pos}_i, q), & q_i \in \Phi_t, \\
q_i, & \text{otherwise},
\end{cases}
\end{equation}
where $q_i$ denotes the query tokens at the $i$-th spatial position, and $\text{pos}_i$ represents the corresponding positional embedding. $\beta_t$ denotes the pyramid-level token retention ratio, which controls the number of queries retained across different feature scales, with values set to $(0.6,\,0.8,\,1.0,\,1.0)$; and $\gamma_l$ represents the Transformer layer-wise token retention ratio, indicating the proportion of queries involved in computation at layer $l$, with values $(1.0,\,0.8,\,0.6,\,0.6,\,0.4,\,0.2)$. The routing strength progressively increases as the network depth grows. Both $\beta_t$ and $\gamma_l$ take values within $(0,1]$ and jointly determine the effective active-query budget of the model. Therefore, the routing decision determines whether a token receives attention-based refinement, rather than whether it is retained in the representation.

As depicted in the query-handling module of Fig.~\ref{fig:CoLR-Det_structure}, foreground and background tokens are processed differently within the encoder layer. The selected foreground tokens are refined through top-$k$ pre-attention, deformable self-attention, and an FFN, whereas the background-dominated tokens routed through the bypass path are supplemented with learnable positional embeddings. By focusing Deformable Attention $\mathcal{A}(\cdot)$ only on $\Phi_t$, the computational complexity is reduced from $O(NHK)$ to $O(\beta_t \gamma_l NHK)$, where \(N\) is the number of spatial tokens, \(H\) is the number of attention heads, and \(K\) is the number of sampled keys in deformable attention. This strategy preserves consistent background priors while directing computation toward salient small objects, achieving a balance between detection accuracy and inference efficiency.

\subsection{Model Optimization}
\label{sec:optimization}

Optimizing CoLR-Det is challenging because dense restoration supervision and sparse detection supervision may introduce different optimization preferences. As shown in Fig.~\ref{fig:optimization}, we propose a Detection-Prioritized Two-Stage Optimization Strategy to improve SR--detection compatibility. Let $\theta_{\text{feat}}$, $\theta_{\text{det}}$, and $\theta_{\text{sr}}$ denote the parameters of the shared encoder, detection head, and SR branch, respectively.

\textbf{Stage One}: The SR branch is frozen, and only $\theta_{\text{feat}}$ and $\theta_{\text{det}}$ are updated using detection-related losses: classification loss $\mathcal{L}_{\text{cls}}$, bounding box loss $\mathcal{L}_{\text{bbox}}$, GIoU loss $\mathcal{L}_{\text{giou}}$, and saliency constraint loss $\mathcal{L}_{\text{sa}}$ defined in Eq.~\eqref{eq:loss_saliency}. This anchors the encoder in a detection-oriented semantic space.

\textbf{Stage Two}: $\theta_{\text{sr}}$ is unfrozen for joint optimization with task-specific learning rates:
\begin{equation}
\eta_{\text{sr}}^{(2)} = \rho \, \eta_{\text{det}}^{(2)}, \quad 0 < \rho < 1,
\end{equation}
where $\eta_{\text{sr}}^{(2)}$ denotes the learning rates for the SR branch. The scaling factor $\rho \in (0,1)$ keeps the parameter updates of the newly activated SR decoder conservative, thereby reducing abrupt restoration-branch changes during collaborative refinement.

\textbf{Overall Loss Functions}.
The overall training process is summarized in Algorithm~\ref{alg:optimization}.
For Stage One:
\begin{equation}
\mathcal{L}_{\text{stage1}} = \beta_{\text{cls}} \mathcal{L}_{\text{cls}} + \beta_{\text{bbox}} \mathcal{L}_{\text{bbox}} + \beta_{\text{giou}} \mathcal{L}_{\text{giou}} + \beta_{\text{sa}} \mathcal{L}_{\text{sa}},
\end{equation}
where $\beta_{\text{cls}}$, $\beta_{\text{bbox}}$, $\beta_{\text{giou}}$, and $\beta_{\text{sa}}$ are weighting coefficients for each loss term.

For Stage Two:
\begin{equation}
\mathcal{L}_{\text{stage2}} = \beta_{\text{sr}} \mathcal{L}_{\text{sr}} + \beta_{\text{cls}} \mathcal{L}_{\text{cls}} + \beta_{\text{bbox}} \mathcal{L}_{\text{bbox}} + \beta_{\text{giou}} \mathcal{L}_{\text{giou}} + \beta_{\text{sa}} \mathcal{L}_{\text{sa}},
\end{equation}
where $\beta_{\text{sr}}$ controls the SR reconstruction loss contribution. Here, $\mathcal{L}_{\text{sr}}$ is defined as the $\ell_1$ loss between the reconstructed image $I_{\text{SR}}$ and the high-resolution target image $I_{\text{HR}}$.


\begin{table*}[!t]
\centering
\caption{\textbf{Performance comparison of CoLR-Det with representative object detectors on the NWPU VHR-10-Split dataset.} All detectors are trained and evaluated on low-resolution images obtained by $2\times$ bicubic downsampling of the original high-resolution inputs. The best and second-best results are marked in \textcolor{red}{\textbf{bold}} and \textcolor{blue}{\underline{underlined}}.}
\label{tab:nwpu_comparison}
\resizebox{\textwidth}{!}{%
\begin{tabular}{c|c|c|c|cccccc}
\toprule
Type & Method & Venue & Backbone & $\text{AP}$ & $\text{AP}_{50}$ & $\text{AP}_{75}$ & AP$_s$ & AP$_m$ & AP$_l$ \\
\midrule
\multirow{12}{*}{\makecell{General\\Object\\Detectors}}
 & Faster RCNN~\cite{Ren2015} & TPAMI2017 & Swin-T & 0.741 & 0.972 & 0.890 & 0.363 & 0.726 & 0.796 \\
 & RetinaNet~\cite{FocalLoss2017} & ICCV2017 & Swin-T & 0.747 & 0.969 & 0.865 & 0.320 & 0.729 & 0.811 \\
 & Cascade R-CNN~\cite{CascadeRCNN2018} & CVPR2018 & Swin-T & 0.778 & 0.975 & 0.918 & 0.418 & 0.767 & 0.843 \\
 & YOLOX-s~\cite{YOLOX2021} & arXiv2021 & CSPDarknet & 0.766 & 0.964 & 0.892 & 0.450 & 0.750 & 0.802 \\
 & Sparse R-CNN~\cite{SparseRCNN2023} & TPAMI2023 & ResNet50 & 0.381 & 0.678 & 0.356 & 0.025 & 0.316 & 0.617 \\
 & SR4IR~\cite{Kim2024} & CVPR2024 & MobileNet-V3 & 0.743 & 0.967 & 0.878 & 0.308 & 0.725 & 0.844 \\
 & D-FINE~\cite{peng2025dfine} & ICLR2025 & HGNetv2-B0 & 0.634 & 0.913 & 0.721 & 0.275 & 0.591 & 0.670 \\
 & DAB-DETR~\cite{DABDETR2022} & ICLR2022 & ResNet50 & 0.656 & 0.935 & 0.756 & 0.342 & 0.609 & 0.791 \\
 & DN-DETR~\cite{DNDETR2022} & CVPR2022 & ResNet50 & 0.681 & 0.936 & 0.782 & 0.299 & 0.652 & 0.800 \\
 & Deformable-DETR~\cite{DeformableDETR2020} & ICLR2021 & ResNet50 & 0.712 & 0.980 & 0.847 & 0.390 & 0.687 & 0.777 \\
 & DINO~\cite{Zhang2022} & ICLR2023 & ResNet50 & 0.793 & 0.980 & 0.933 & 0.534 & 0.769 & 0.832 \\
 & DINO~\cite{Zhang2022} & ICLR2023 & Swin-T & 0.766 & 0.986 & 0.911 & 0.487 & 0.744 & 0.830 \\
\midrule
\multirow{9}{*}{\makecell{Remote\\Sensing\\Object\\Detectors}}
 & EESRGAN~\cite{Rabbi2020} & Remote Sens. 2020 & ResNet50 & 0.749 & 0.971 & 0.891 & 0.448 & 0.724 & 0.802 \\
 & FSANet~\cite{FSANet2023} & TGRS2022 & Swin-T & 0.671 & 0.956 & 0.791 & 0.306 & 0.649 & 0.754 \\
 & SuperYOLO (RGB)~\cite{SuperYOLO2023} & TGRS2023 & CSPDarknet & 0.700 & 0.868 & 0.825 & 0.554 & 0.704 & 0.773 \\
 & FFCA-YOLO~\cite{FFCAYOLO2024} & TGRS2024 & CSPDarknet & 0.689 & 0.946 & 0.836 & 0.427 & 0.688 & 0.778 \\
 & LEGNet-T~\cite{LEGNet2025} & ICCVW2025 & LEGNet-T & 0.788 & 0.979 & 0.921 & 0.449 & 0.769 & 0.841 \\
 & MGAM~\cite{MGAM2025} & TGRS2025 & ResNet50 & 0.754 & 0.969 & 0.883 & 0.349 & 0.730 & 0.816 \\
 & Strip-RCNN~\cite{yuan2026strip} & AAAI2026 & StripNet-T & 0.657 & 0.961 & 0.774 & 0.310 & 0.639 & 0.719 \\
 \rowcolor{oursbg}&  \textbf{CoLR-Det (Ours)} & - & ResNet50 & \textcolor{blue}{\underline{0.828}} & \textcolor{blue}{\underline{0.987}} & \textcolor{blue}{\underline{0.946}} & \textcolor{blue}{\underline{0.607}} & \textcolor{blue}{\underline{0.813}} & \textcolor{red}{\textbf{0.910}} \\
 \rowcolor{oursbg} &  \textbf{CoLR-Det (Ours)} & - & Swin-T & \textcolor{red}{\textbf{0.836}} & \textcolor{red}{\textbf{0.991}} & \textcolor{red}{\textbf{0.954}} & \textcolor{red}{\textbf{0.649}} & \textcolor{red}{\textbf{0.816}} & \textcolor{blue}{\underline{0.890}} \\
\bottomrule
\end{tabular}
}
\end{table*}

\section{EXPERIMENTS AND ANALYSIS}
\label{sec:Experiment}

\subsection{Implementation Details}

\subsubsection{Datasets}

To comprehensively evaluate the proposed method, we conduct experiments on three benchmark remote sensing datasets: NWPU VHR-10~\cite{Cheng2016}, DOTAv1.5~\cite{DOTA2018}, and HRSSD~\cite{Zhang2019}. Our method focuses on horizontal bounding box (HBB) detection. We adopt a standard sliding-window cropping strategy to satisfy network input requirements and reduce information loss caused by direct resizing.

\textbf{DOTAv1.5 Dataset}~\cite{DOTA2018}: The DOTAv1.5 dataset contains large-scale high-resolution remote sensing images with sizes reaching up to 4000$\times$4000 pixels. To facilitate processing, these images are cropped into 1024$\times$1024 patches with a 200-pixel overlap to ensure that objects near patch boundaries remain intact. For images smaller than 1024$\times$1024 pixels, zero-padding is applied. The resulting DOTAv1.5-Split dataset consists of 11{,}828 training images and 4{,}055 testing images.

\textbf{NWPU VHR-10 Dataset}~\cite{Cheng2016}: The NWPU VHR-10 dataset is a collection of high-resolution remote sensing images with a focus on various object categories, including vehicles and buildings. For this dataset, we apply a similar preprocessing strategy by cropping the original images into 512$\times$512 patches with a 100-pixel overlap. This results in the NWPU VHR-10-Split dataset, which contains 2{,}723 training images and 908 testing images.

\textbf{HRSSD Dataset}~\cite{Zhang2019}: The HRSSD dataset includes high-resolution remote sensing images that cover a wide range of small-object categories. Similar to DOTAv1.5, these images are cropped into 1024$\times$1024 patches with a 200-pixel overlap. The resulting HRSSD-Split dataset contains 26{,}215 training images and 26{,}841 testing images.

For all datasets, we follow the official train/test split and perform cropping after the split to prevent patches from the same original image appearing in both training and testing sets. 


\subsubsection{Training and Optimization Settings}

In the SR setup, we generate low-resolution inputs using a fixed bicubic downsampling protocol and evaluate the model's downstream detection performance. The cropped 1024$\times$1024 and 512$\times$512 patches are treated as HR ground truth. Each HR image is downsampled by a factor of 2 using bicubic interpolation to generate the corresponding LR input. During training, the network only receives these LR images (512$\times$512 or 256$\times$256), which forces the model to recover fine details and detect objects from significantly degraded visual evidence.

\begin{table*}[htbp]
\centering
\caption{Performance comparison of CoLR-Det with representative object detectors on the DOTAv1.5-Split dataset. All detectors are trained and evaluated on low-resolution images obtained by $2\times$ bicubic downsampling of the original high-resolution inputs. The best and second-best results are marked in \textcolor{red}{\textbf{bold}} and \textcolor{blue}{\underline{underlined}}.}
\label{tab:dotav1.5_comparison}
\resizebox{\textwidth}{!}{%
\begin{tabular}{c|c|c|c|cccccc}
\toprule
Type & Method & Venue & Backbone & $\text{AP}$ & $\text{AP}_{50}$ & $\text{AP}_{75}$ & AP$_s$ & AP$_m$ & AP$_l$ \\
\midrule
\multirow{12}{*}{\makecell{General\\Object\\Detectors}}
 & Faster RCNN~\cite{Ren2015} & TPAMI2017 & Swin-T & 0.336 & 0.541 & 0.364 & 0.123 & 0.376 & 0.475 \\
 & RetinaNet~\cite{FocalLoss2017} & ICCV2017 & Swin-T & 0.307 & 0.530 & 0.309 & 0.077 & 0.323 & 0.465 \\
 & Cascade R-CNN~\cite{CascadeRCNN2018} & CVPR2018 & Swin-T & 0.357 & 0.537 & 0.391 & 0.142 & 0.396 & 0.513 \\
 & YOLOX-s~\cite{YOLOX2021} & arXiv2021 & CSPDarknet & 0.269 & 0.498 & 0.277 & 0.119 & 0.292 & 0.347 \\
 & Sparse R-CNN~\cite{SparseRCNN2023} & TPAMI2023 & ResNet50 & 0.266 & 0.446 & 0.279 & 0.060 & 0.263 & 0.422 \\
 & SR4IR~\cite{Kim2024} & CVPR2024 & MobileNet-V3 & 0.329 & 0.511 & 0.340 & 0.147 & 0.358 & 0.495 \\
 & D-FINE~\cite{peng2025dfine} & ICLR2025 & HGNetv2-B0 & 0.203 & 0.381 & 0.196 & 0.066 & 0.209 & 0.309 \\
 & DAB-DETR~\cite{DABDETR2022} & ICLR2022 & ResNet50 & 0.233 & 0.440 & 0.215 & 0.035 & 0.211 & 0.394 \\
 & DN-DETR~\cite{DNDETR2022} & CVPR2022 & ResNet50 & 0.274 & 0.462 & 0.254 & 0.030 & 0.289 & 0.425 \\
 & Deformable-DETR~\cite{DeformableDETR2020} & ICLR2021 & Swin-T & 0.248 & 0.464 & 0.229 & 0.044 & 0.256 & 0.434 \\
 & DINO~\cite{Zhang2022} & ICLR2023 & ResNet50 & 0.367 & 0.597 & 0.397 & 0.160 & 0.394 & 0.506 \\
 & DINO~\cite{Zhang2022} & ICLR2023 & Swin-T & \textcolor{blue}{\underline{0.402}} & \textcolor{blue}{\underline{0.659}} & \textcolor{blue}{\underline{0.427}} & 0.192 & \textcolor{blue}{\underline{0.434}} & \textcolor{blue}{\underline{0.540}} \\
\midrule
\multirow{9}{*}{\makecell{Remote\\Sensing\\Object\\Detectors}}
 & EESRGAN~\cite{Rabbi2020} & Remote Sens. 2020 & ResNet50 & 0.314 & 0.512 & 0.342 & 0.160 & 0.360 & 0.481 \\
 & FSANet~\cite{FSANet2023} & TGRS2022 & Swin-T & 0.283 & 0.557 & 0.271 & 0.103 & 0.312 & 0.384 \\
 & SuperYOLO (RGB)~\cite{SuperYOLO2023} & TGRS2023 & CSPDarknet & 0.307 & 0.495 & 0.397 & 0.191 & 0.335 & 0.429 \\
 & FFCA-YOLO~\cite{FFCAYOLO2024} & TGRS2024 & CSPDarknet & 0.399 & 0.642 & 0.425 & \textcolor{blue}{\underline{0.221}} & 0.425 & 0.470 \\
 & LEGNet-T~\cite{LEGNet2025} & ICCVW2025 & LEGNet-T & 0.316 & 0.508 & 0.343 & 0.084 & 0.361 & 0.434 \\
 & MGAM~\cite{MGAM2025} & TGRS2025 & ResNet50 & 0.300 & 0.488 & 0.326 & 0.079 & 0.326 & 0.435 \\
 & Strip-RCNN~\cite{yuan2026strip} & AAAI2026 & StripNet-T & 0.312 & 0.526 & 0.321 & 0.099 & 0.358 & 0.444 \\
 \rowcolor{oursbg}& \textbf{CoLR-Det (Ours)} & - & ResNet50 & 0.373 & 0.613 & 0.403 & 0.201 & 0.403 & 0.512 \\
 \rowcolor{oursbg}& \textbf{CoLR-Det (Ours)} & - & Swin-T & \textcolor{red}{\textbf{0.419}} & \textcolor{red}{\textbf{0.674}} & \textcolor{red}{\textbf{0.455}} & \textcolor{red}{\textbf{0.250}} & \textcolor{red}{\textbf{0.451}} & \textcolor{red}{\textbf{0.562}} \\
\bottomrule
\end{tabular}
}
\end{table*}

The network is implemented in PyTorch and trained on a single NVIDIA RTX~3090 GPU with a batch size of 2. We use the AdamW optimizer with an initial learning rate of $1 \times 10^{-4}$ and a weight decay of $1 \times 10^{-4}$. Gradient clipping is applied with max\_norm=0.1, and a 0.1$\times$ learning-rate multiplier is used for the backbone. The learning rate follows a multi-step decay schedule (MultiStepLR), in which it is reduced by a factor of 0.1 at predefined milestones within an epoch-based training loop. We adopt the proposed two-stage training strategy: the detection network is trained for the first 16 epochs, and the full model is then jointly optimized until $T_{\text{tot}}=52$ under the same learning-rate policy. During joint optimization, the SR branch uses a learning-rate scaling factor of $\rho=0.1$, and the loss weights are set to $\beta_{\text{cls}}=1.0$, $\beta_{\text{bbox}}=5.0$, $\beta_{\text{giou}}=2.0$, $\beta_{\text{sa}}=2.0$, and $\beta_{\text{sr}}=1.0$. The modulation coefficient $\alpha$ is learnable and initialized from $\mathcal{U}(-0.3,\,0.3)$. The input resolution is kept fixed during both training and testing. For a controlled and reproducible evaluation, we only apply random flipping as data augmentation, while disabling random cropping, random resizing, and other additional augmentations.

\subsection{Comparison with State-of-the-Art Models}

To evaluate the performance of CoLR-Det, we compare it with representative state-of-the-art object detectors. Quantitative results on NWPU VHR-10-Split, DOTAv1.5-Split, and HRSSD-Split are summarized in Tables~\ref{tab:nwpu_comparison}--\ref{tab:hrssd_comparison}. All methods are retrained under identical experimental settings.

As shown in Tables~\ref{tab:nwpu_comparison}--\ref{tab:hrssd_comparison}, these methods cover CNN-based one-stage and two-stage detectors, Transformer-based DETR variants, a representative task-driven SR-assisted recognition method, and specialized approaches developed for remote sensing and tiny-object detection. In addition, we include an architectural variant in which the shared encoder is replaced with ResNet-50 and the restoration branch adopts a symmetric U-Net structure, serving as a supplementary architecture-level comparison.
Among them, EESRGAN, SuperYOLO, FSANet, FFCA-YOLO, LEGNet-T, MGAM, and Strip R-CNN are included to provide comparisons with methods that specifically address challenges such as resolution degradation, multiscale feature representation, tiny-object detection, and complex object geometries in remote sensing imagery. LEGNet-T is a lightweight and effective backbone designed for remote sensing object detection. We integrate it with Faster R-CNN and adapt it to horizontal bounding box detection for evaluation. Since SuperYOLO is originally designed as a multimodal fusion detector, only RGB images are used for a fair comparison. CoLR-Det achieves the best overall AP and small-object AP on all three benchmarks, demonstrating consistent advantages in small-object detection.

Compared with the same-backbone Swin-T DINO baseline~\cite{Zhang2022}, CoLR-Det achieves clear performance gains. On NWPU VHR-10-Split, the absolute improvements in AP, $\text{AP}_{50}$, and AP$_s$ are 7.0, 0.5, and 16.2 percentage points, respectively. The corresponding gains are 1.7, 1.5, and 5.8 percentage points on DOTAv1.5-Split, and 1.0, 1.1, and 5.6 percentage points on HRSSD-Split. The large improvement in AP$_s$ shows that the proposed saliency-driven SR--detection collaboration can generate discriminative features for dense small objects in cluttered backgrounds while preserving strong overall detection accuracy. To further demonstrate the advantage of this collaborative framework beyond accuracy, Section~\ref{subsec:model_efficiency} analyzes the accuracy--efficiency trade-off against two-stage serial SR--detection pipelines. This category-level analysis further confirms that the proposed saliency-driven collaborative mechanism improves the discriminability of challenging small targets that are otherwise prone to missed detections.

\begin{figure*}[htbp]
    \centering
    \setlength{\tabcolsep}{1pt}
    \renewcommand{\arraystretch}{0.5}
    \begin{tabular}{c ccccccc}
        & \scriptsize GT
        & \scriptsize Faster R-CNN
        & \scriptsize YOLOX-s
        & \scriptsize \shortstack{DeformableDETR}
        & \scriptsize MGAM
        & \scriptsize DINO
        & \scriptsize \shortstack{CoLR-Det(Ours)} \\[2pt]

        \multirow{2}{*}[4ex]{\rotatebox{90}{\scriptsize NWPU VHR-10-Split}}
        & \includegraphics[width=0.125\textwidth]{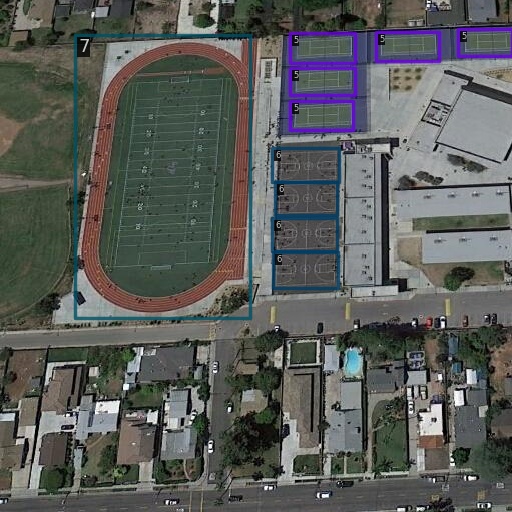}
        & \includegraphics[width=0.125\textwidth]{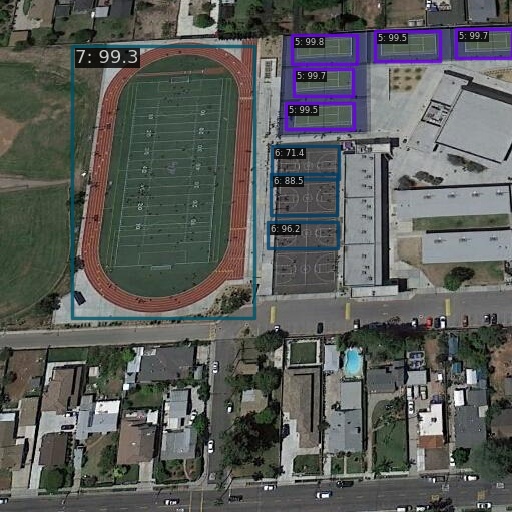}
        & \includegraphics[width=0.125\textwidth]{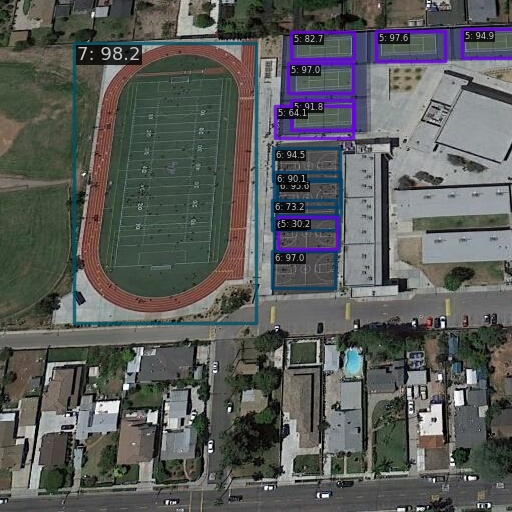}
        & \includegraphics[width=0.125\textwidth]{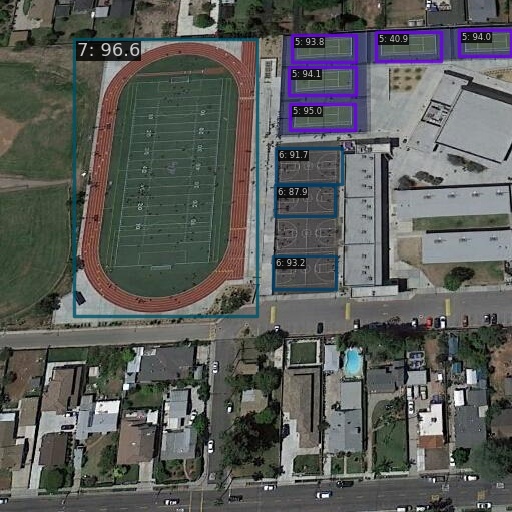}
        & \includegraphics[width=0.125\textwidth]{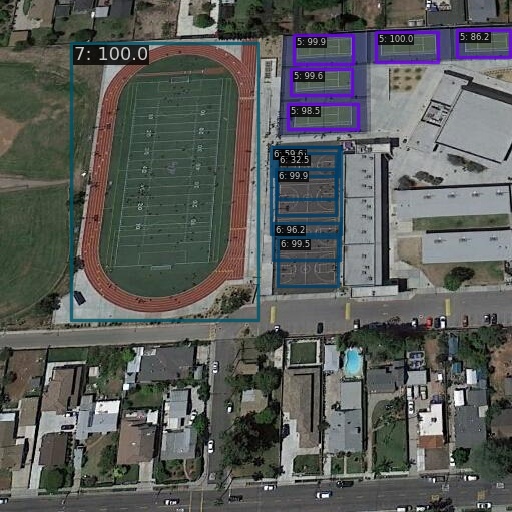}
        & \includegraphics[width=0.125\textwidth]{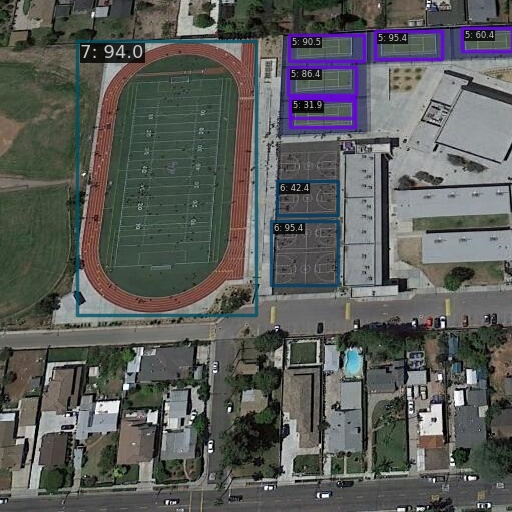}
        & \includegraphics[width=0.125\textwidth]{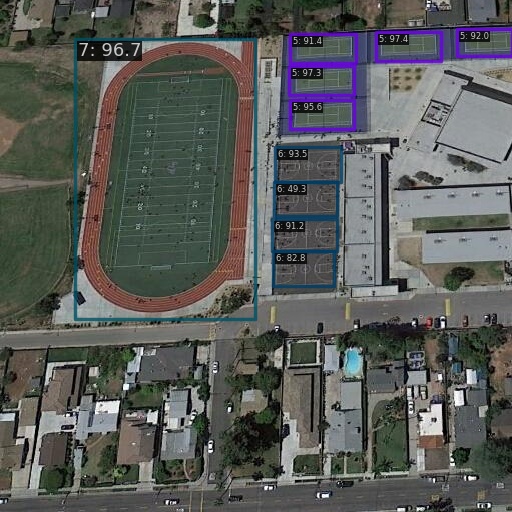} \\

        & \includegraphics[width=0.125\textwidth]{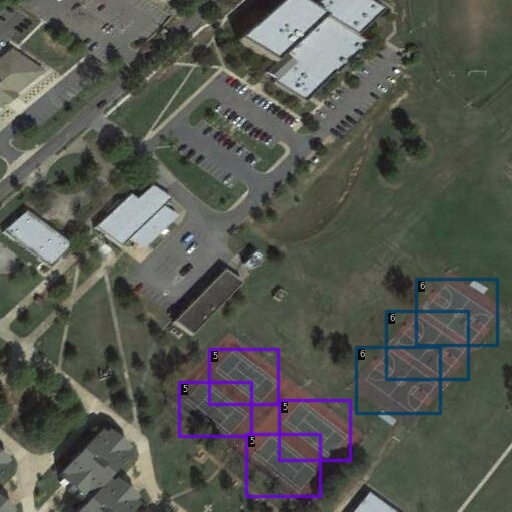}
        & \includegraphics[width=0.125\textwidth]{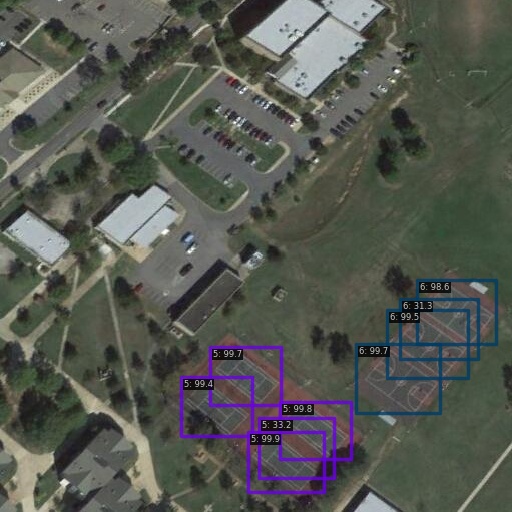}
        & \includegraphics[width=0.125\textwidth]{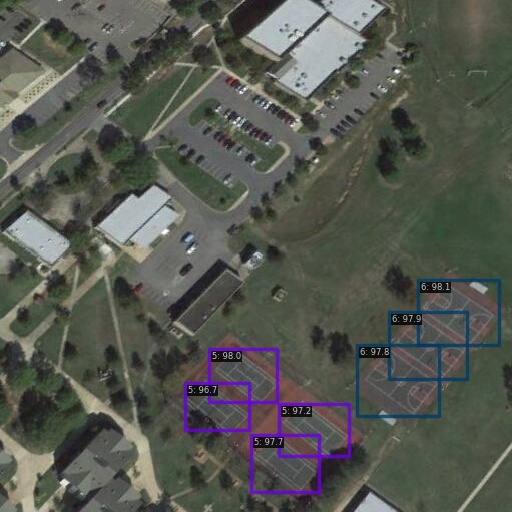}
        & \includegraphics[width=0.125\textwidth]{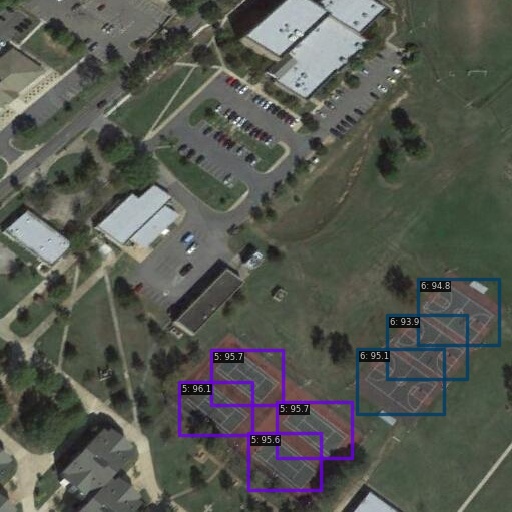}
        & \includegraphics[width=0.125\textwidth]{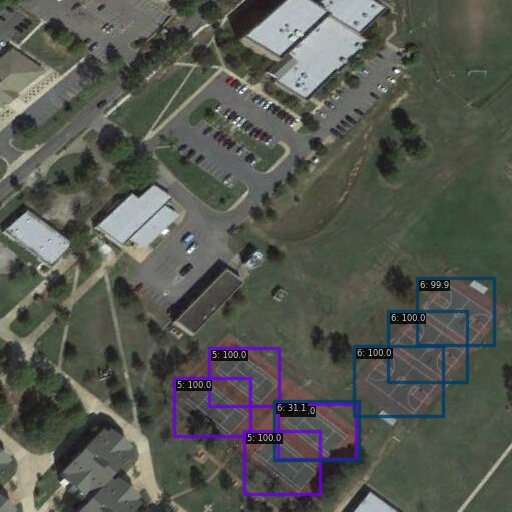}
        & \includegraphics[width=0.125\textwidth]{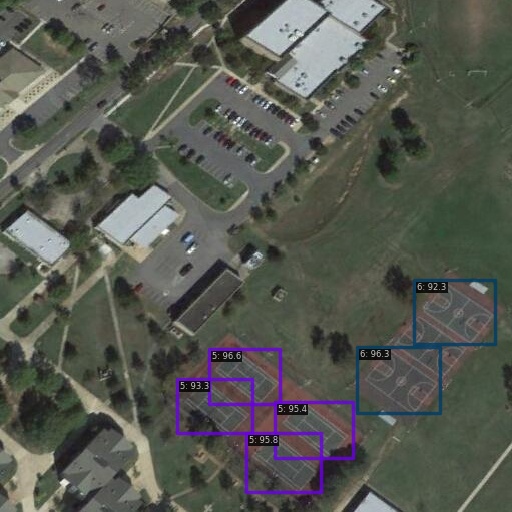}
        & \includegraphics[width=0.125\textwidth]{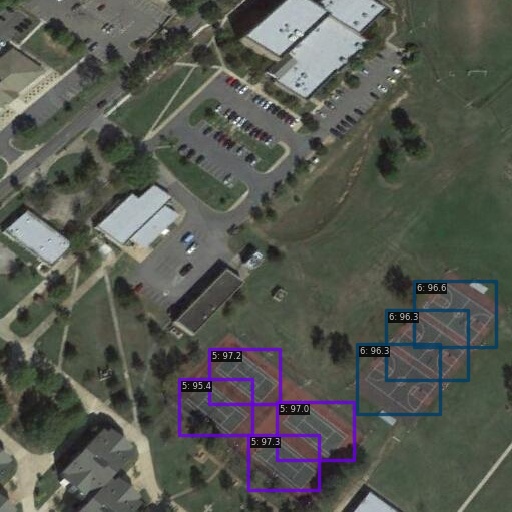} \\[2pt]

        \multirow{2}{*}[3ex]{\rotatebox{90}{\scriptsize DOTAv1.5-Split}}
        & \includegraphics[width=0.125\textwidth]{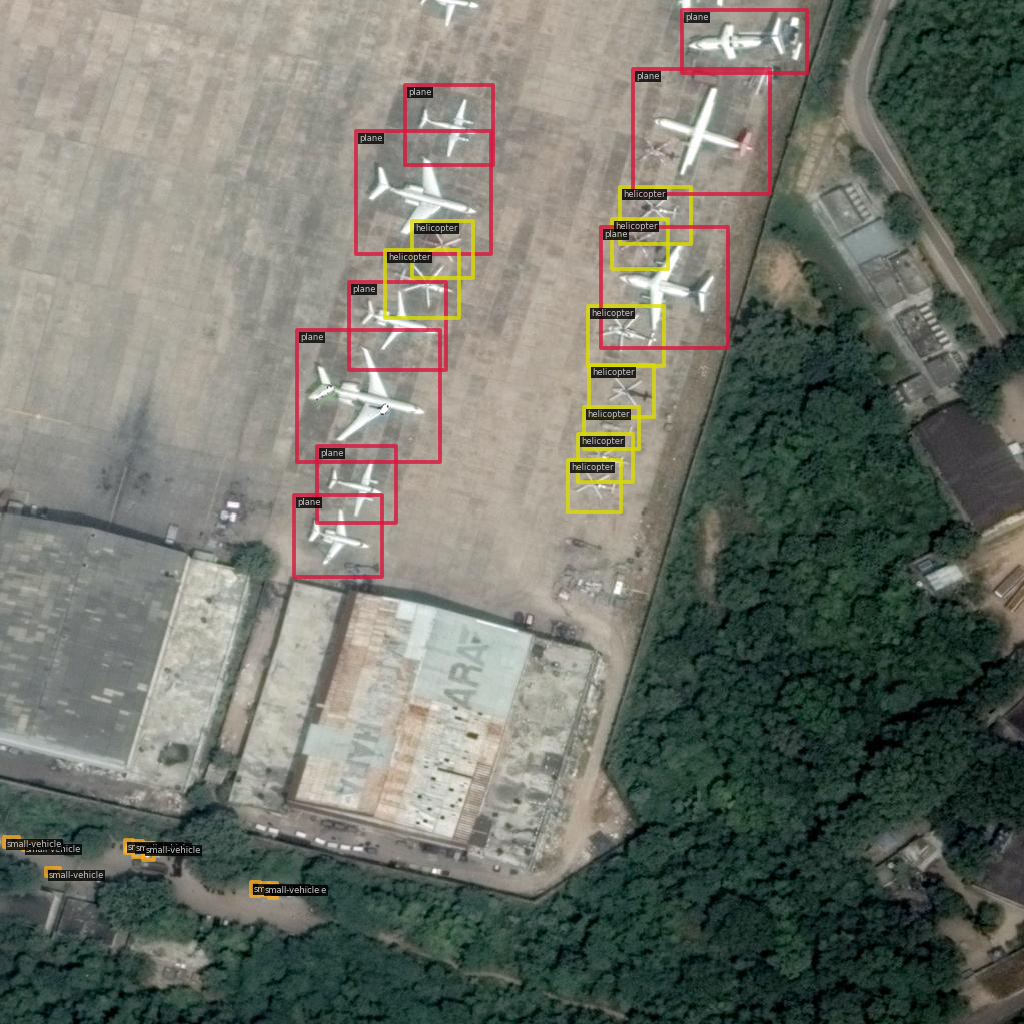}
        & \includegraphics[width=0.125\textwidth]{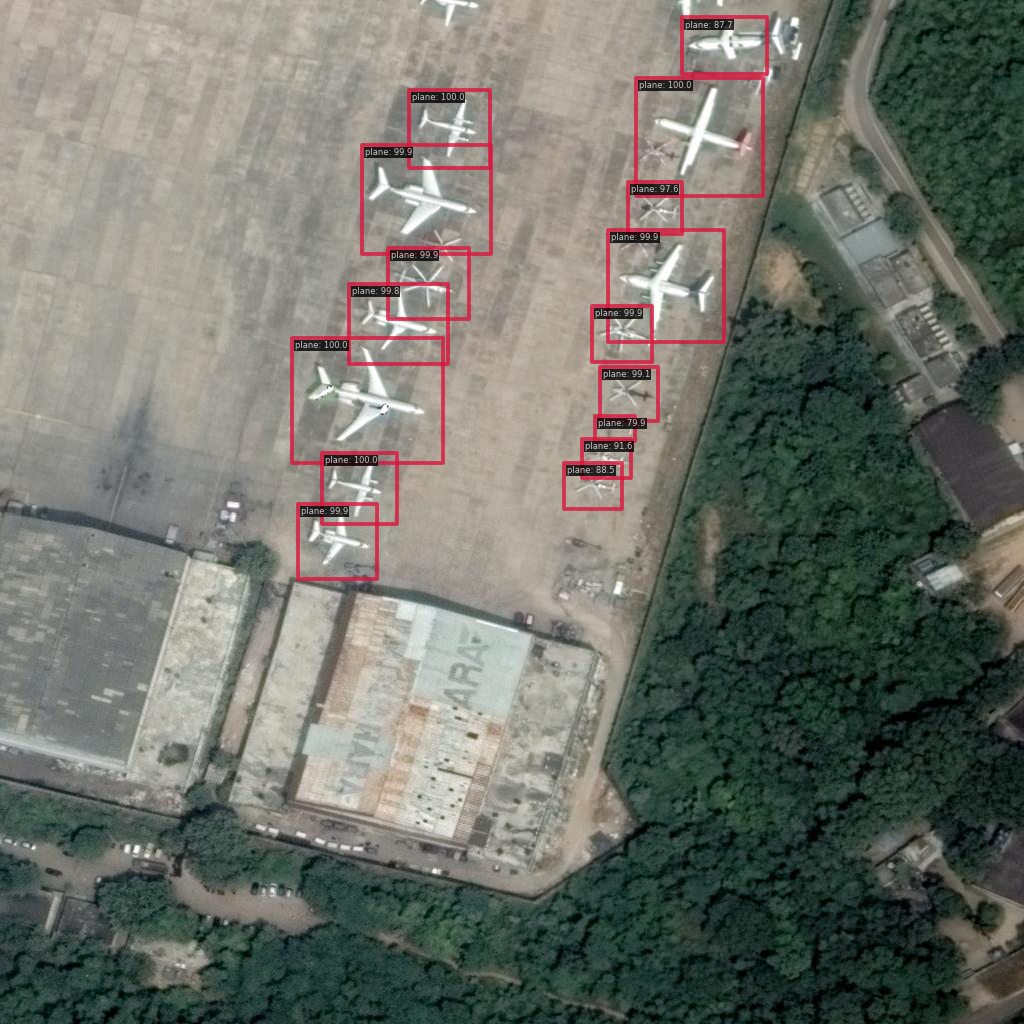}
        & \includegraphics[width=0.125\textwidth]{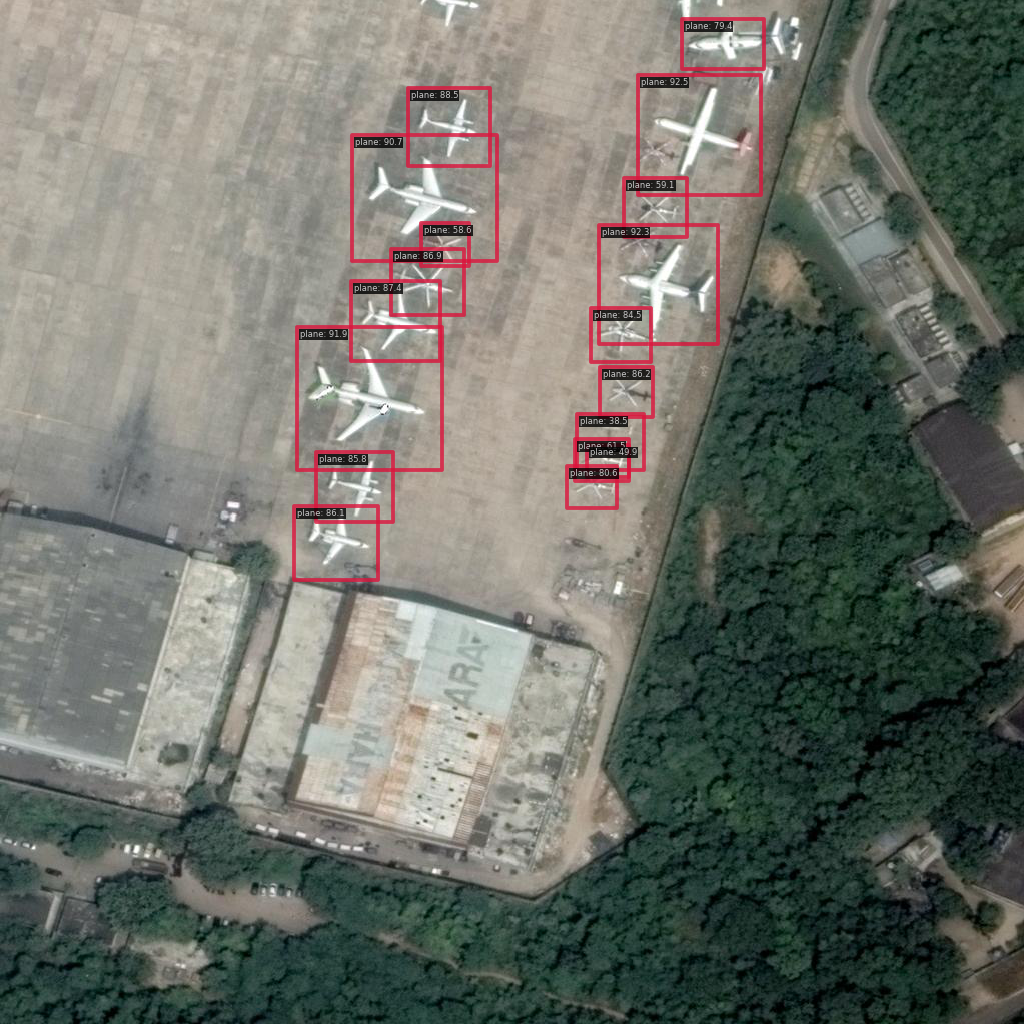}
        & \includegraphics[width=0.125\textwidth]{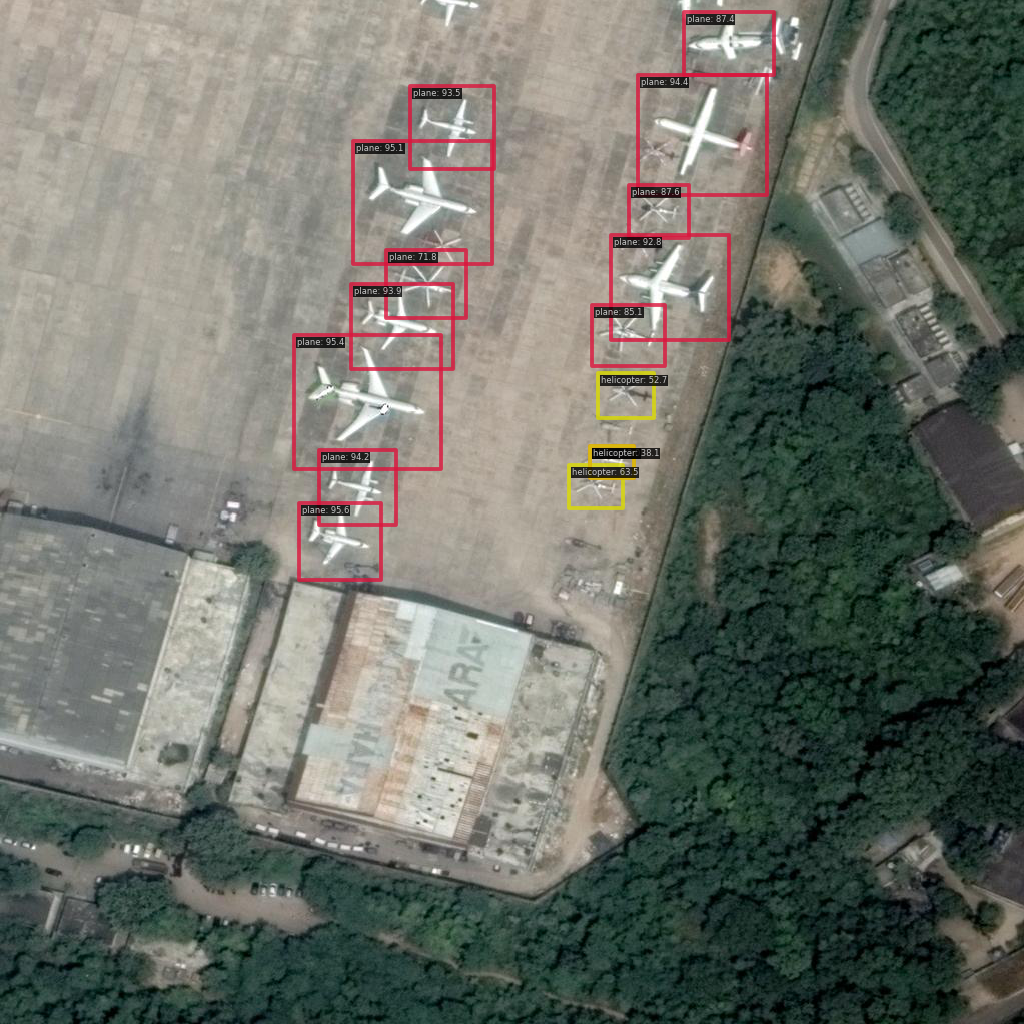}
        & \includegraphics[width=0.125\textwidth]{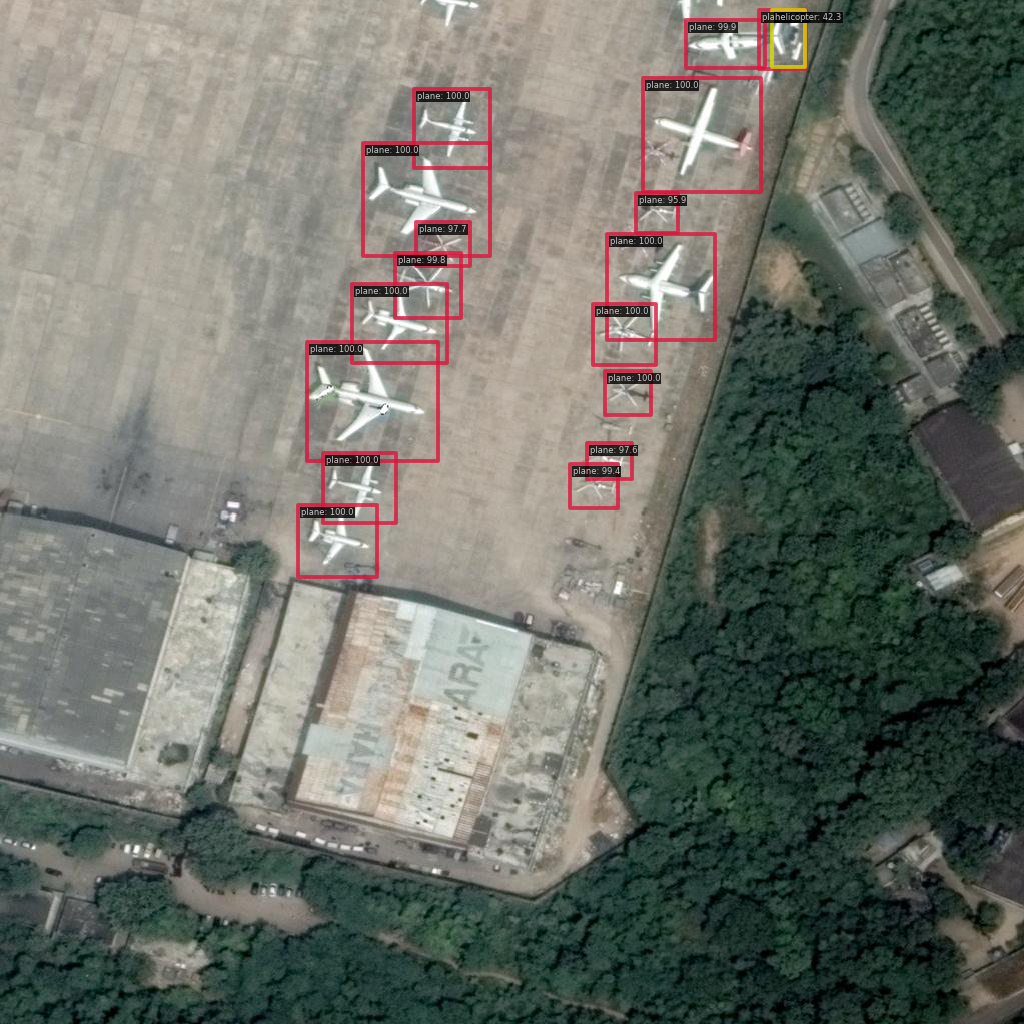}
        & \includegraphics[width=0.125\textwidth]{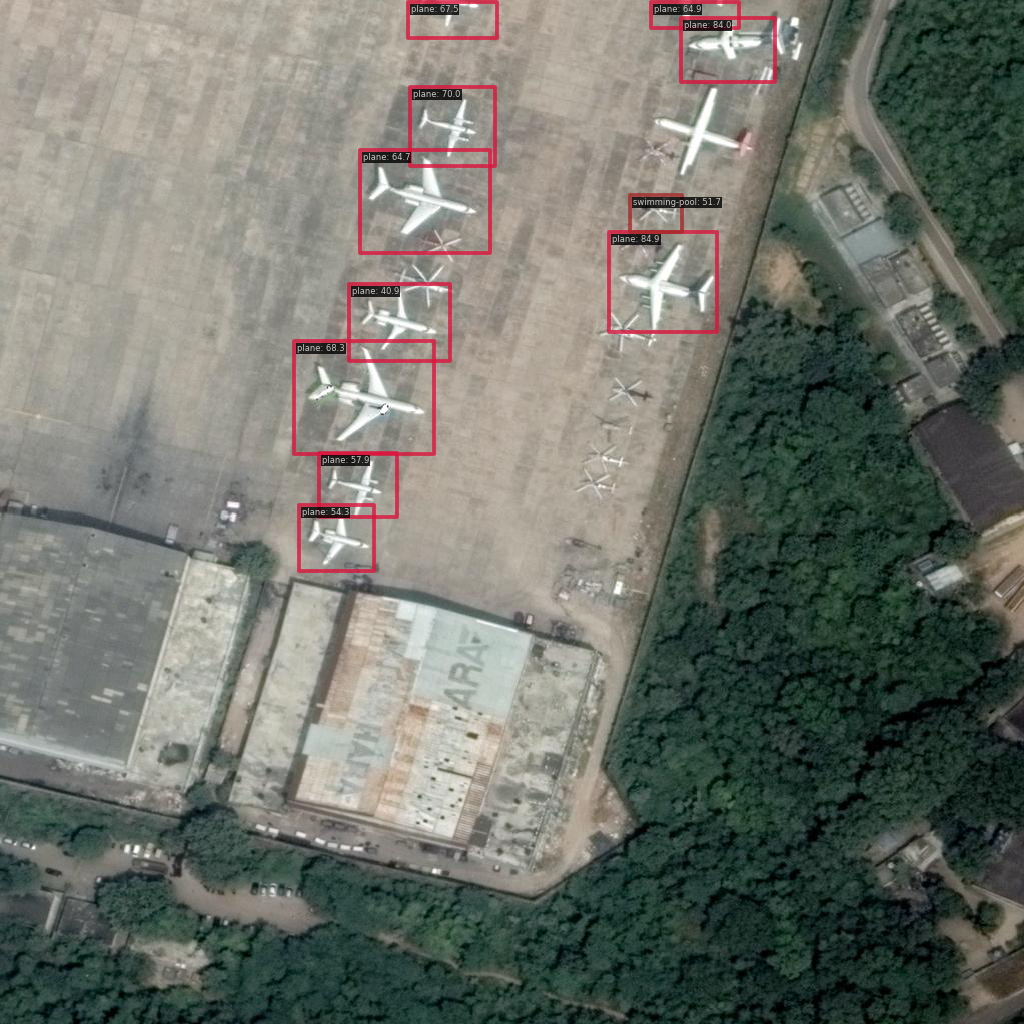}
        & \includegraphics[width=0.125\textwidth]{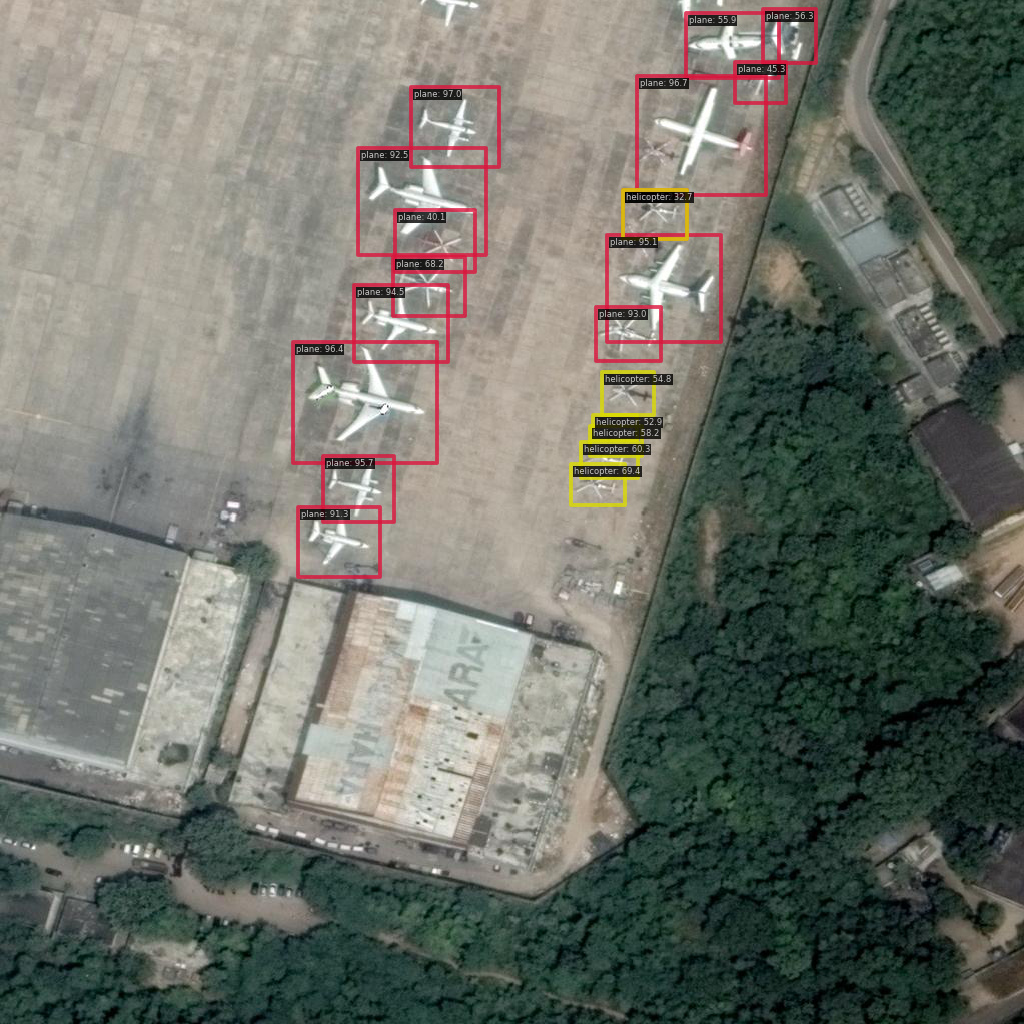} \\

        & \includegraphics[width=0.125\textwidth]{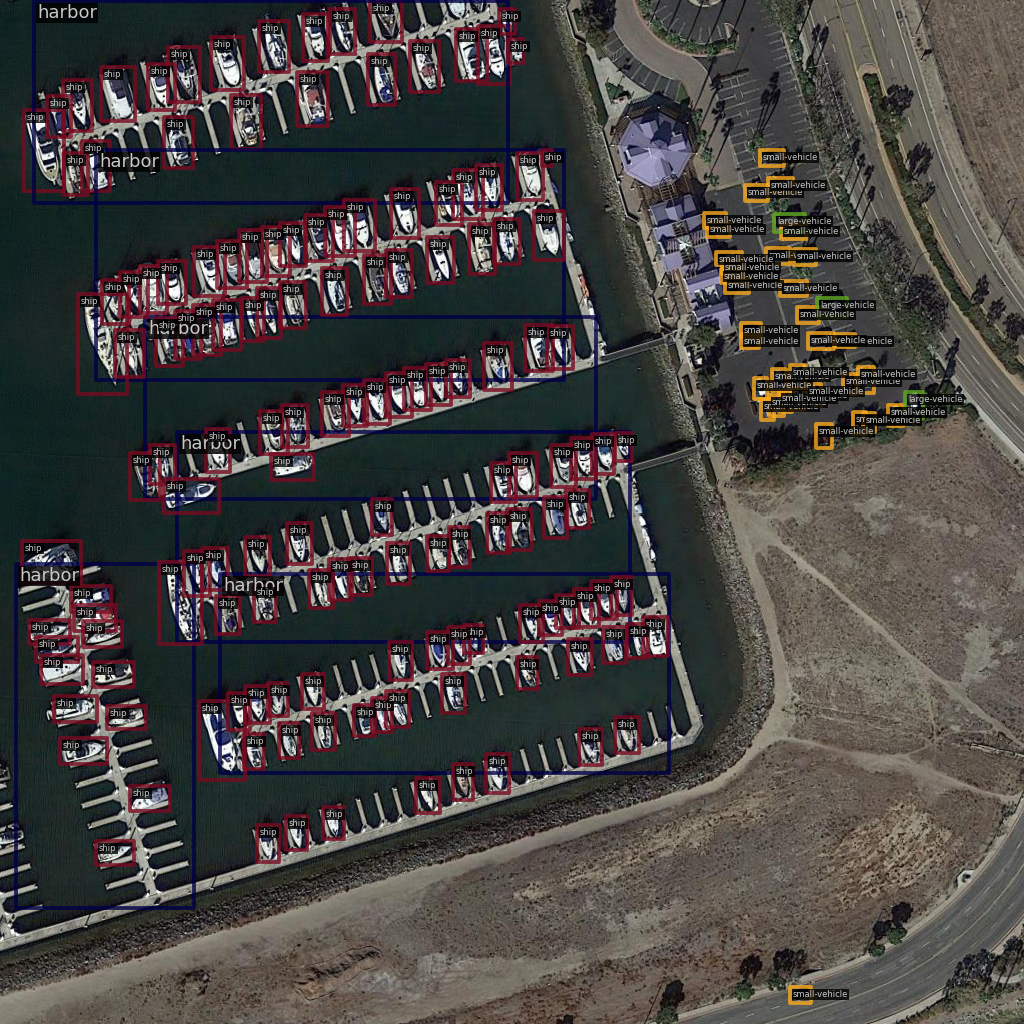}
        & \includegraphics[width=0.125\textwidth]{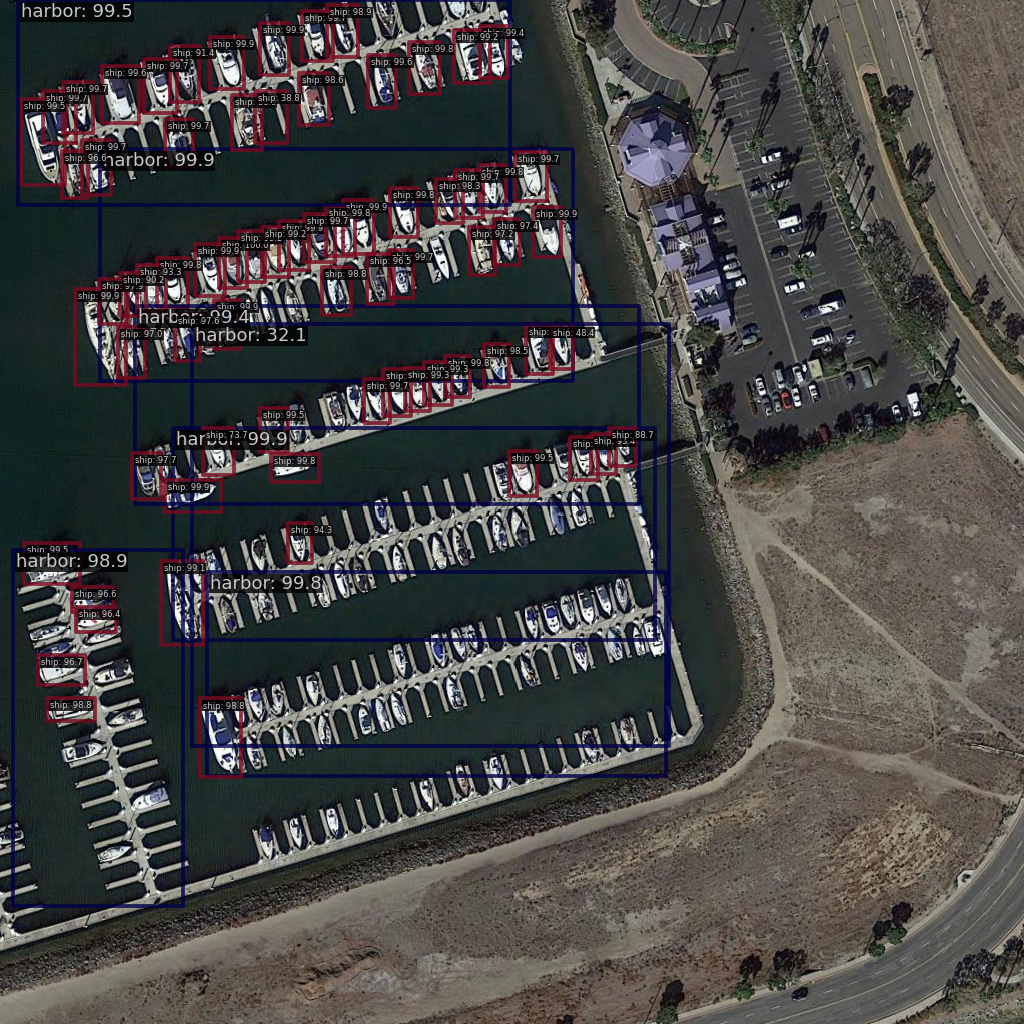}
        & \includegraphics[width=0.125\textwidth]{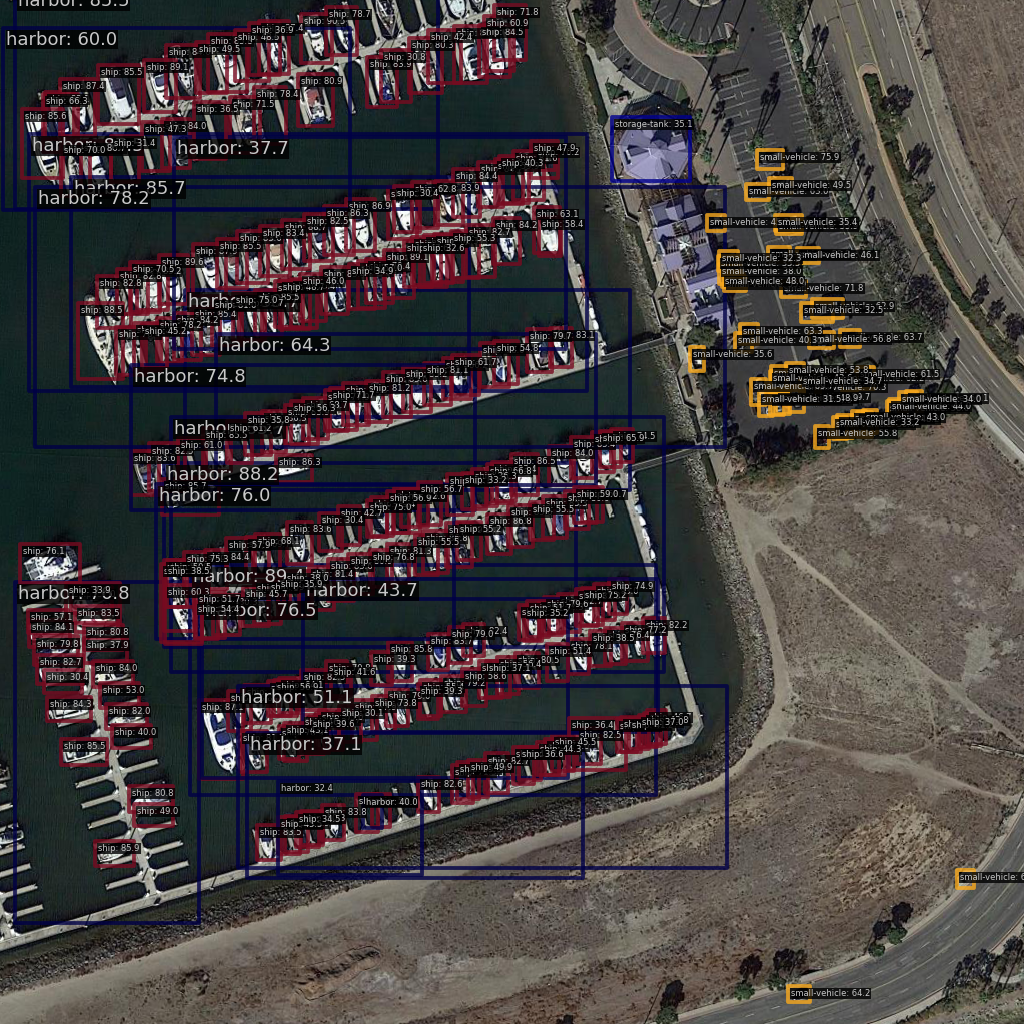}
        & \includegraphics[width=0.125\textwidth]{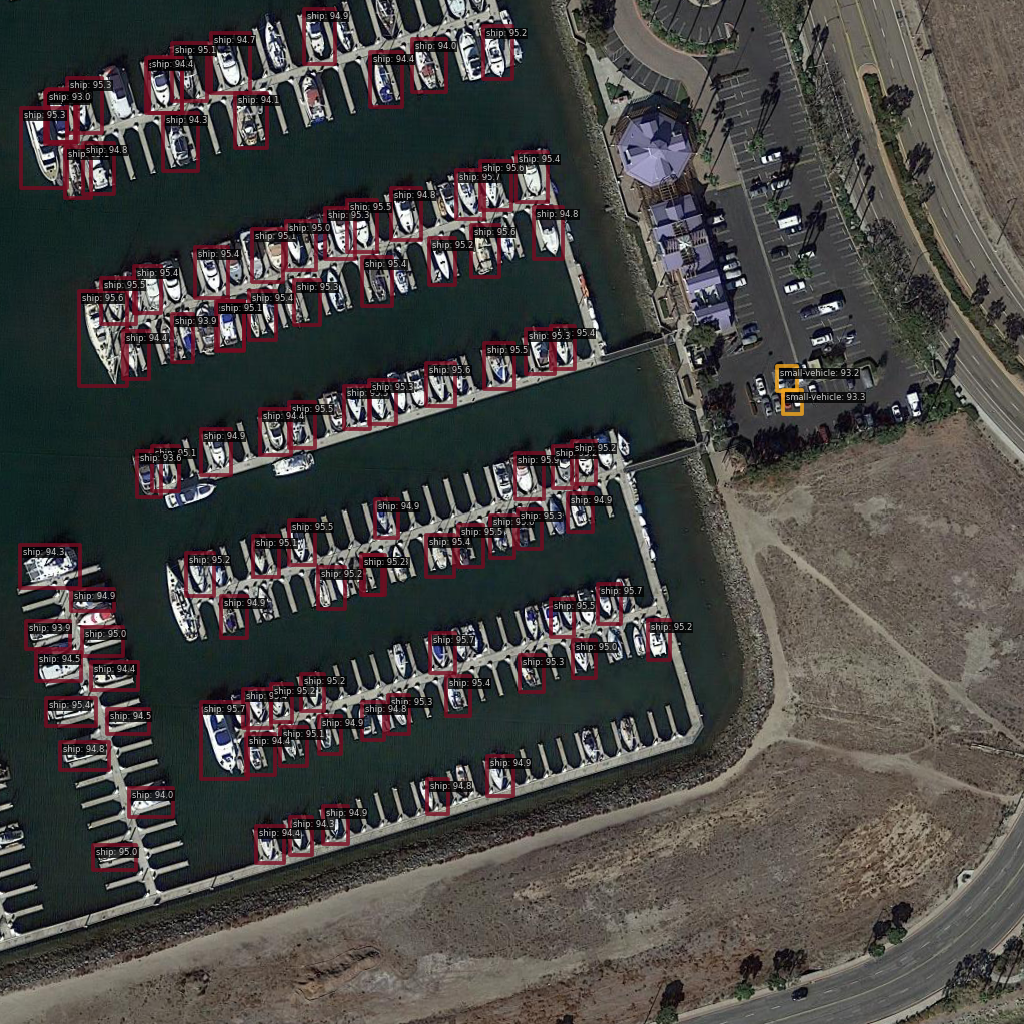}
        & \includegraphics[width=0.125\textwidth]{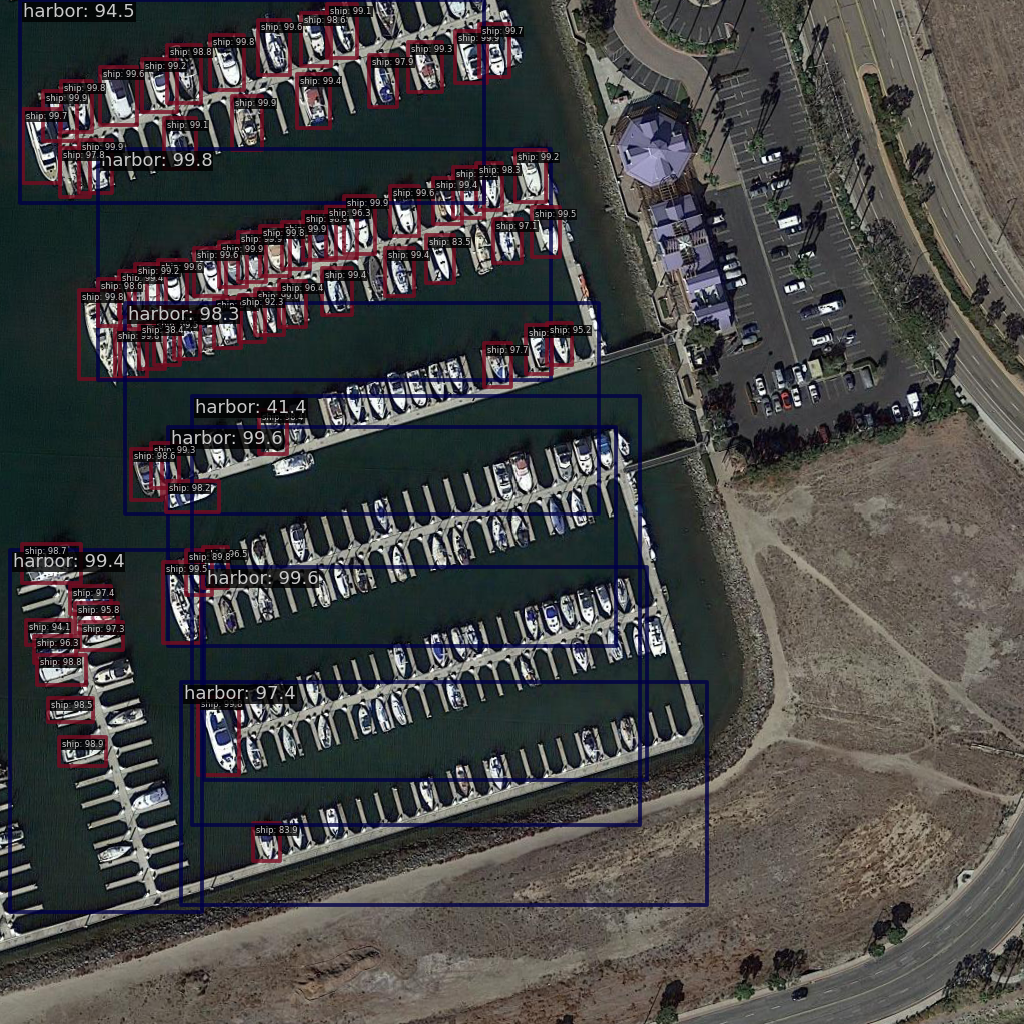}
        & \includegraphics[width=0.125\textwidth]{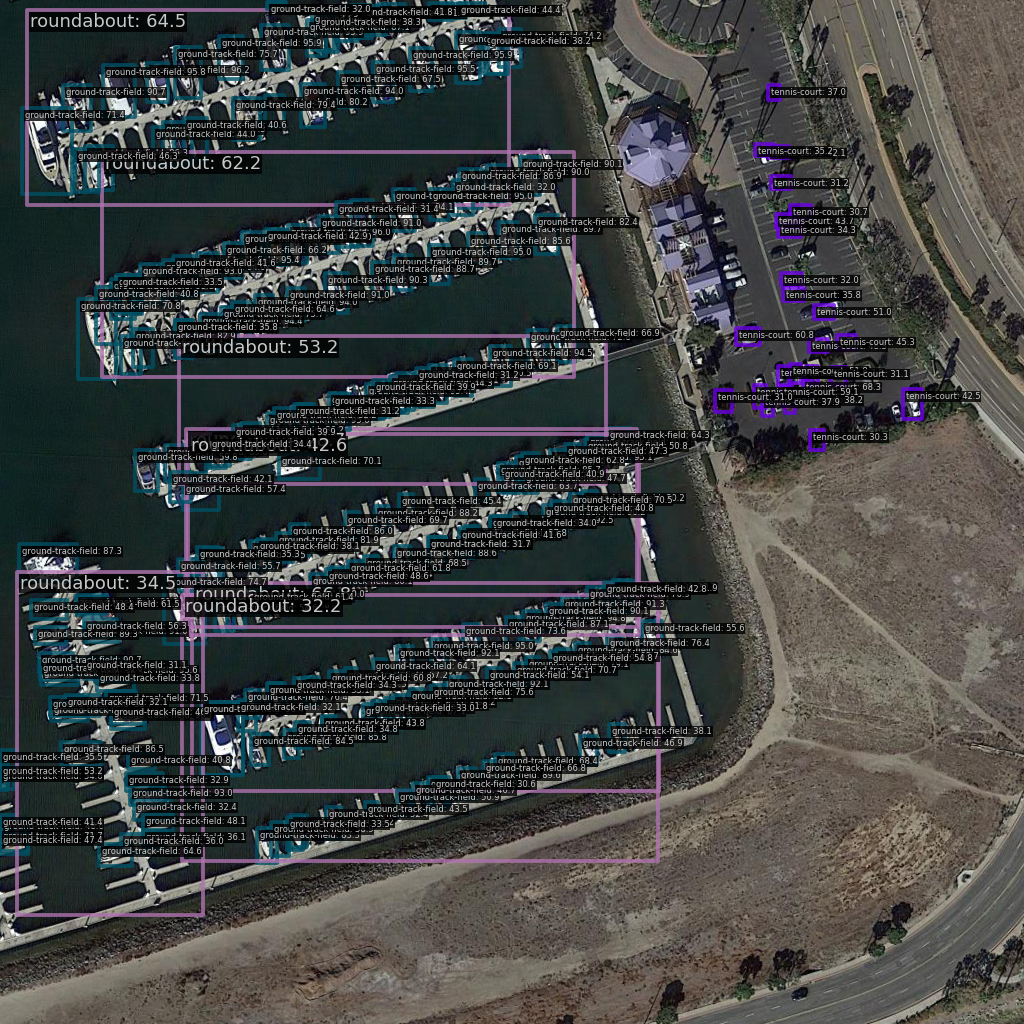}
        & \includegraphics[width=0.125\textwidth]{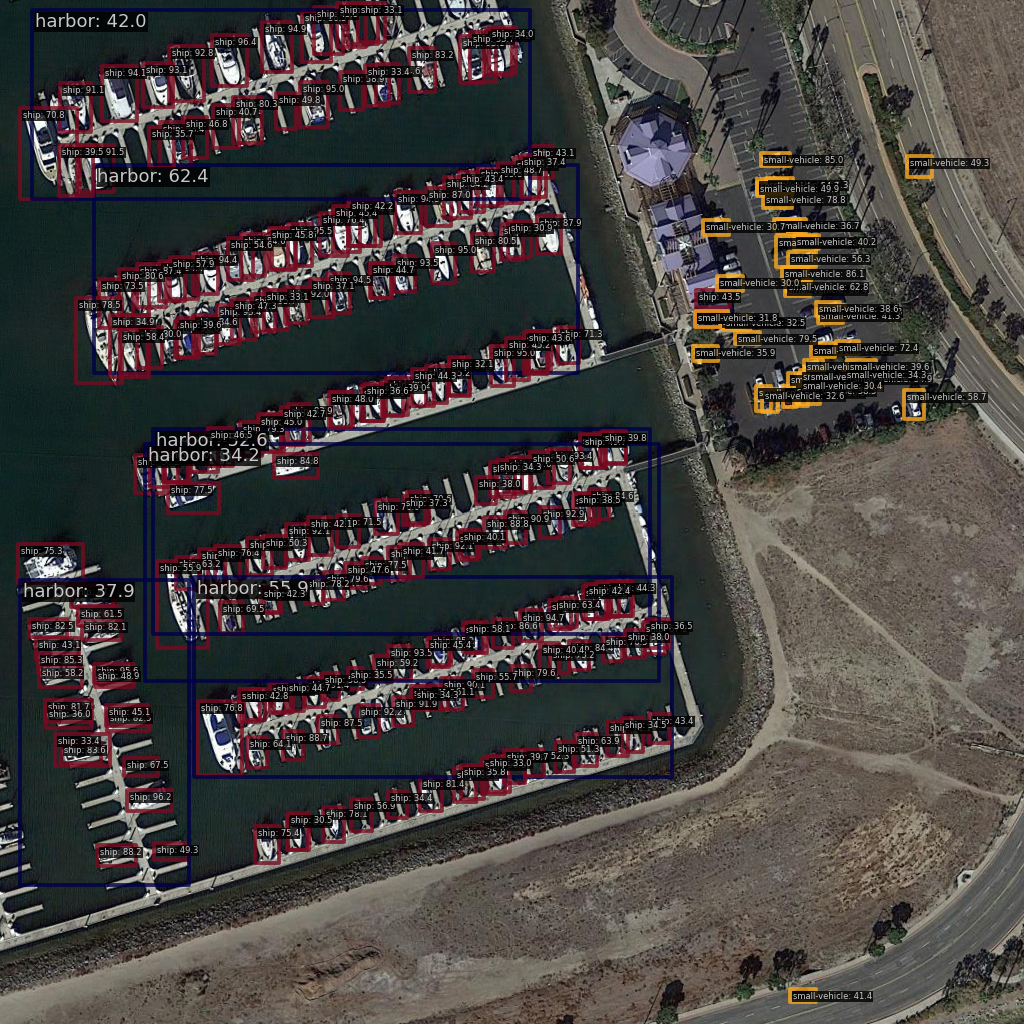} \\[2pt]

        \multirow{2}{*}[3ex]{\rotatebox{90}{\scriptsize HRSSD-Split}}
        & \includegraphics[width=0.125\textwidth]{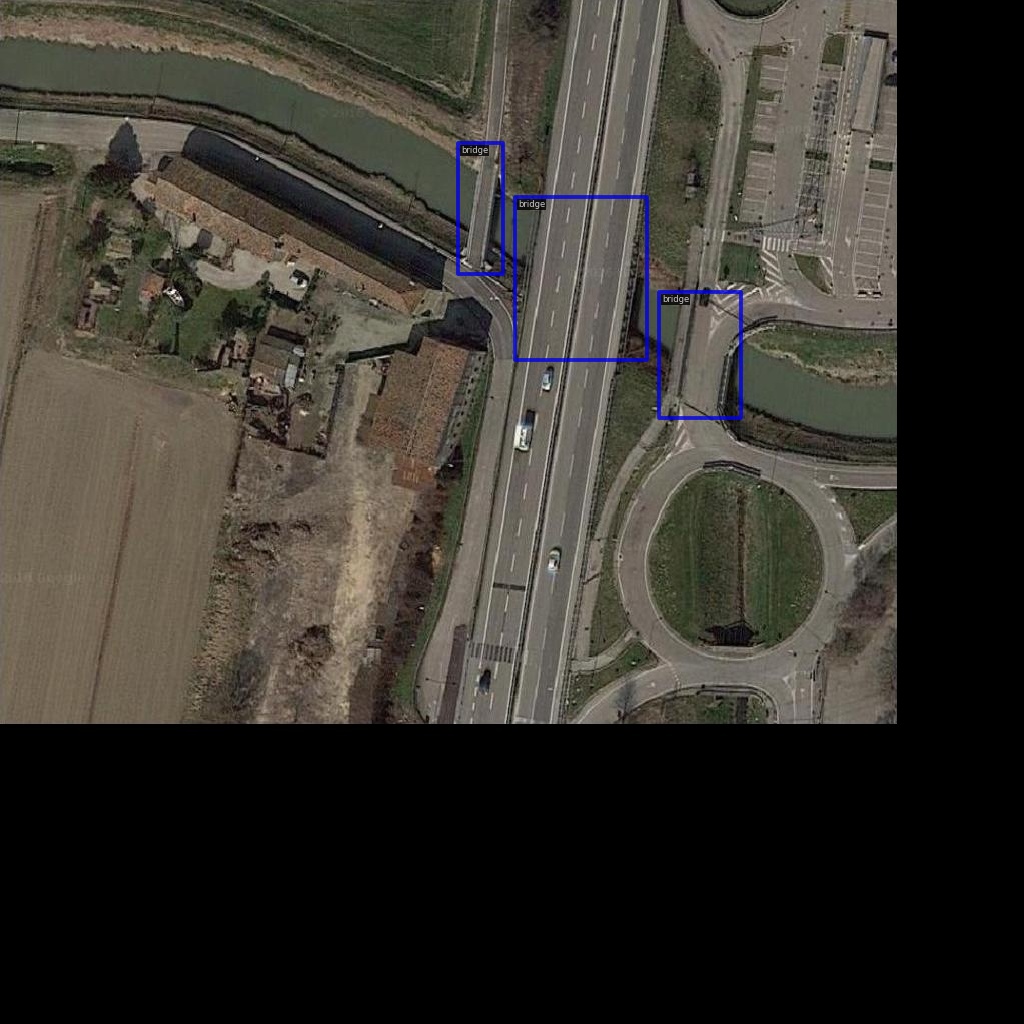}
        & \includegraphics[width=0.125\textwidth]{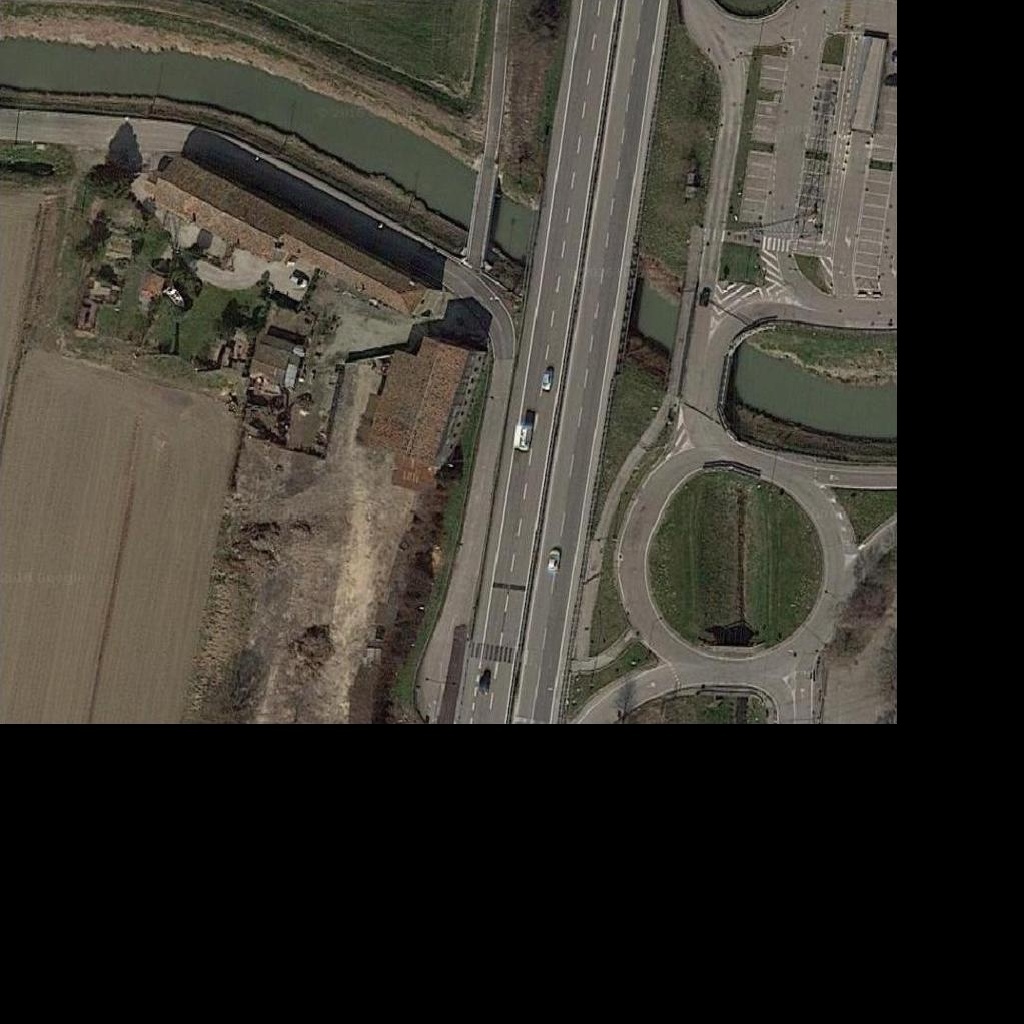}
        & \includegraphics[width=0.125\textwidth]{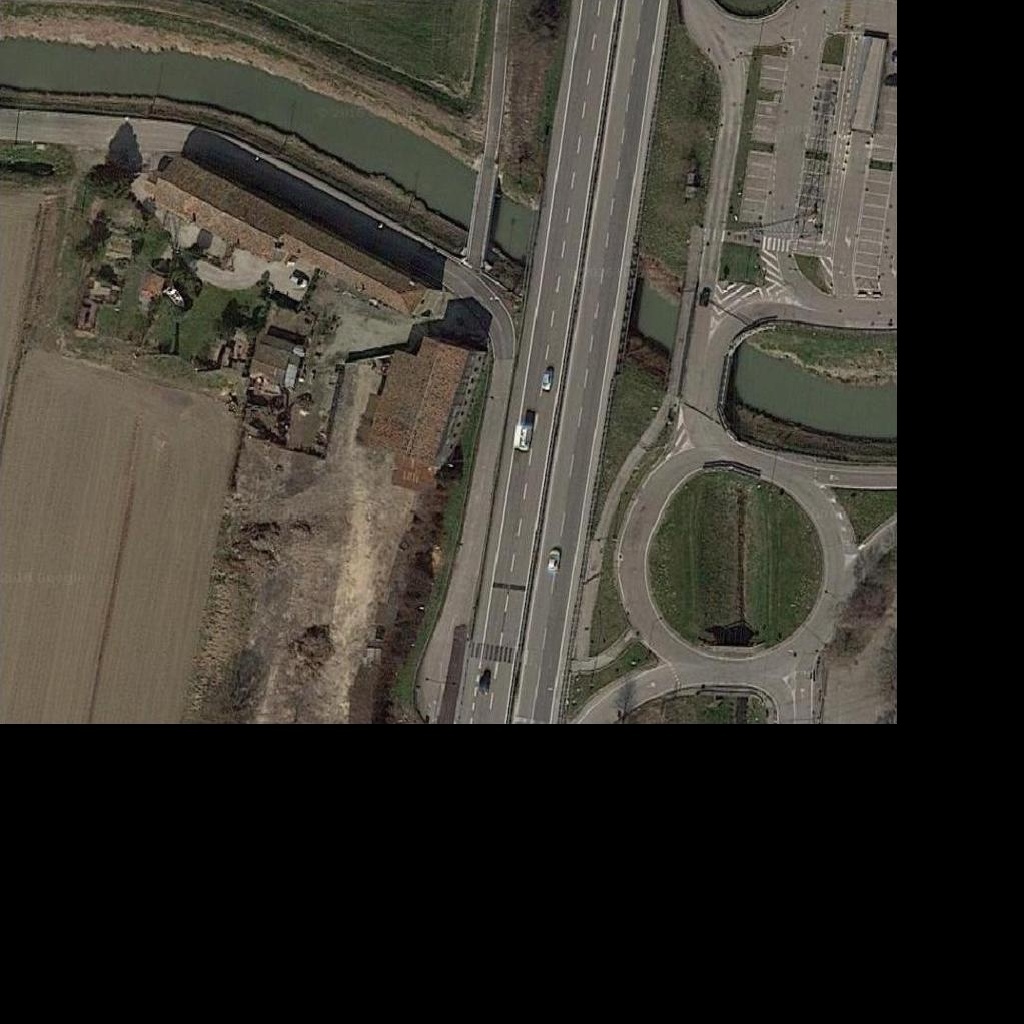}
        & \includegraphics[width=0.125\textwidth]{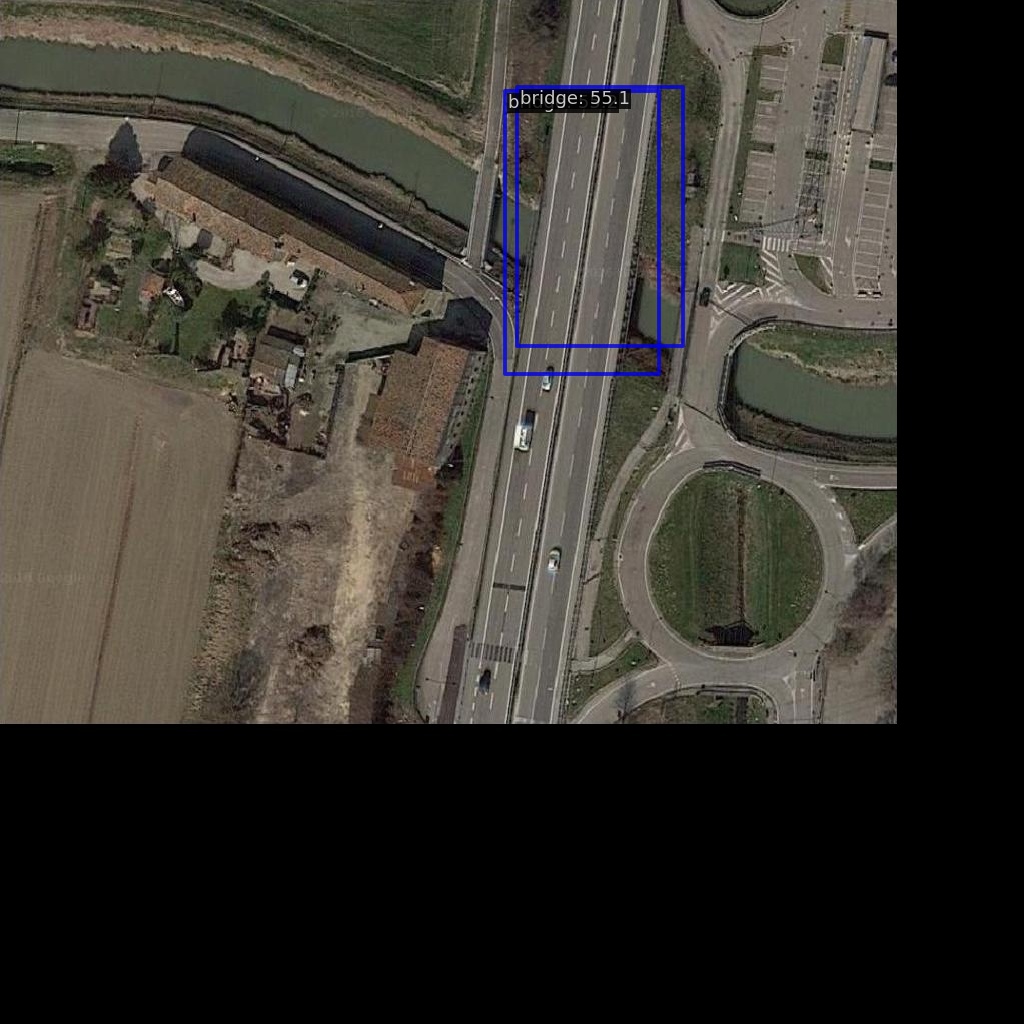}
        & \includegraphics[width=0.125\textwidth]{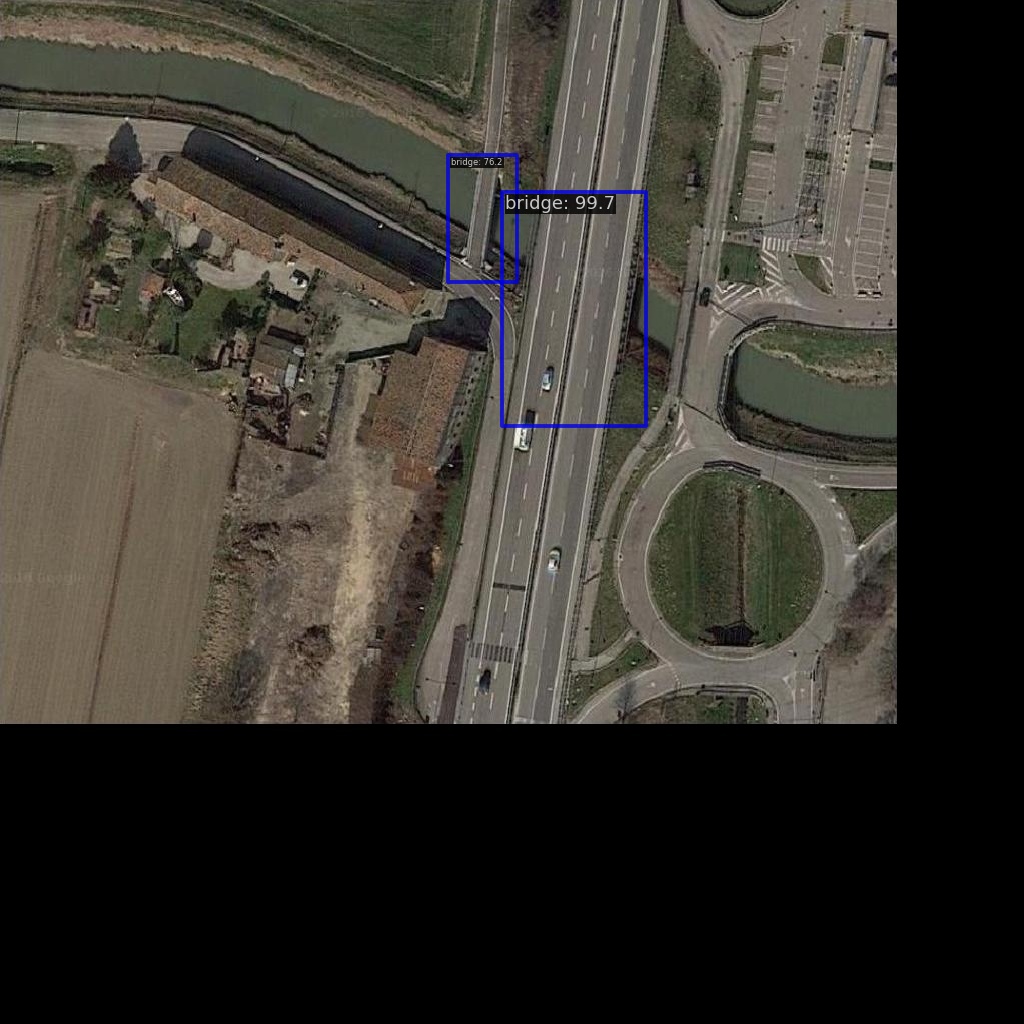}
        & \includegraphics[width=0.125\textwidth]{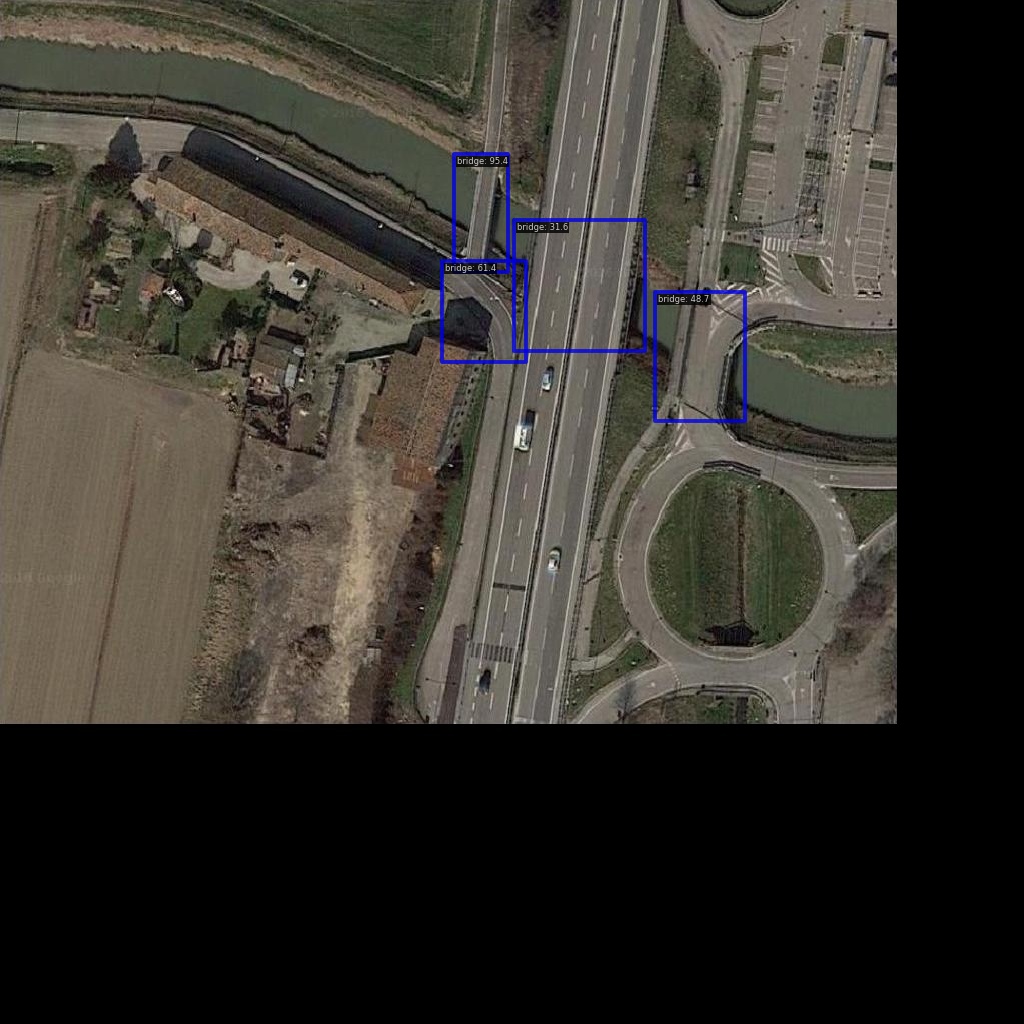}
        & \includegraphics[width=0.125\textwidth]{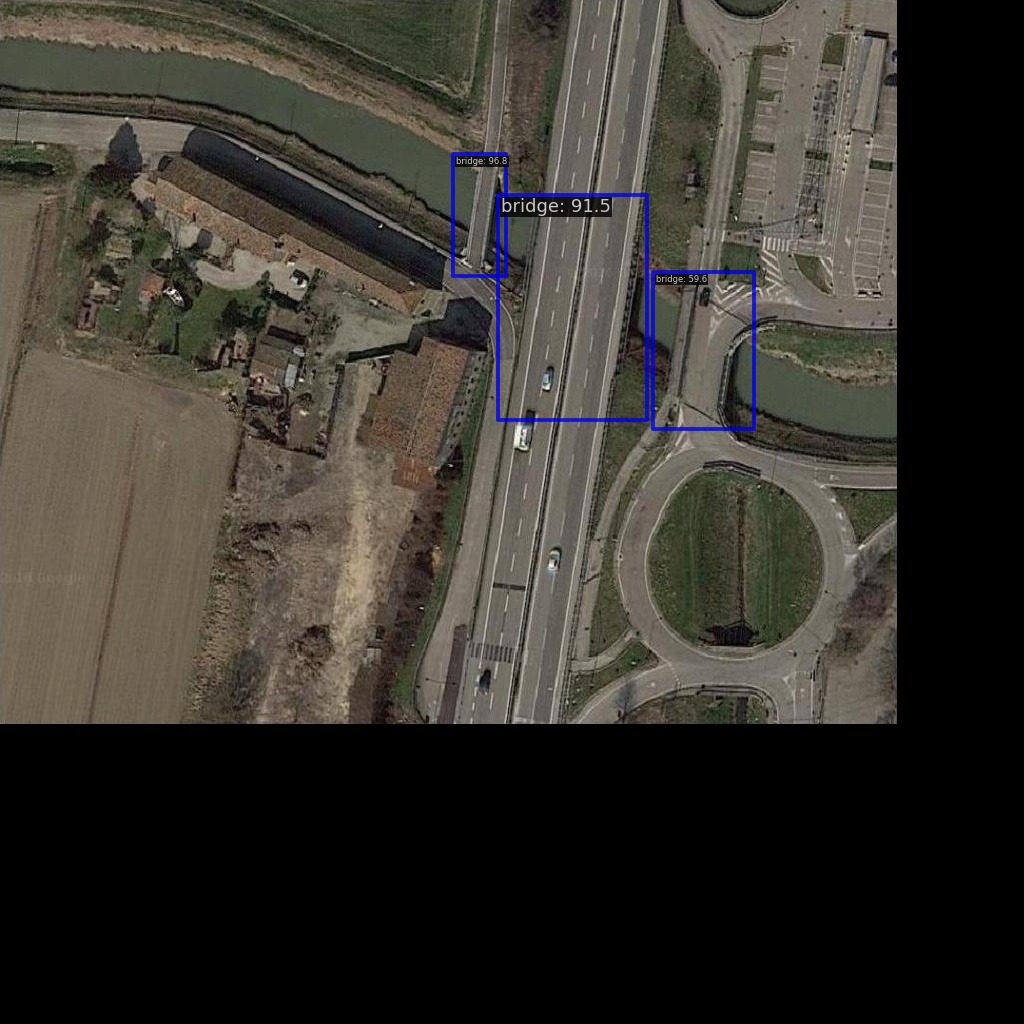} \\

        & \includegraphics[width=0.125\textwidth]{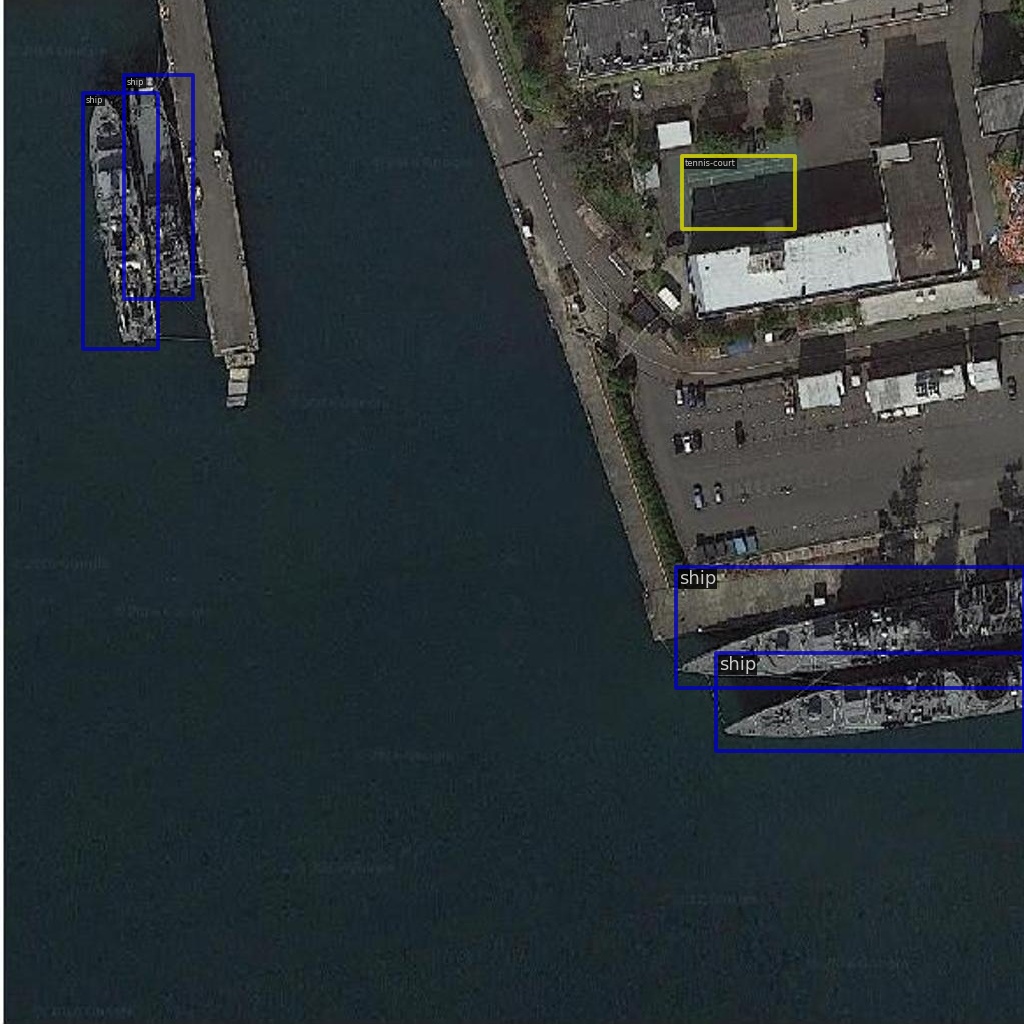}
        & \includegraphics[width=0.125\textwidth]{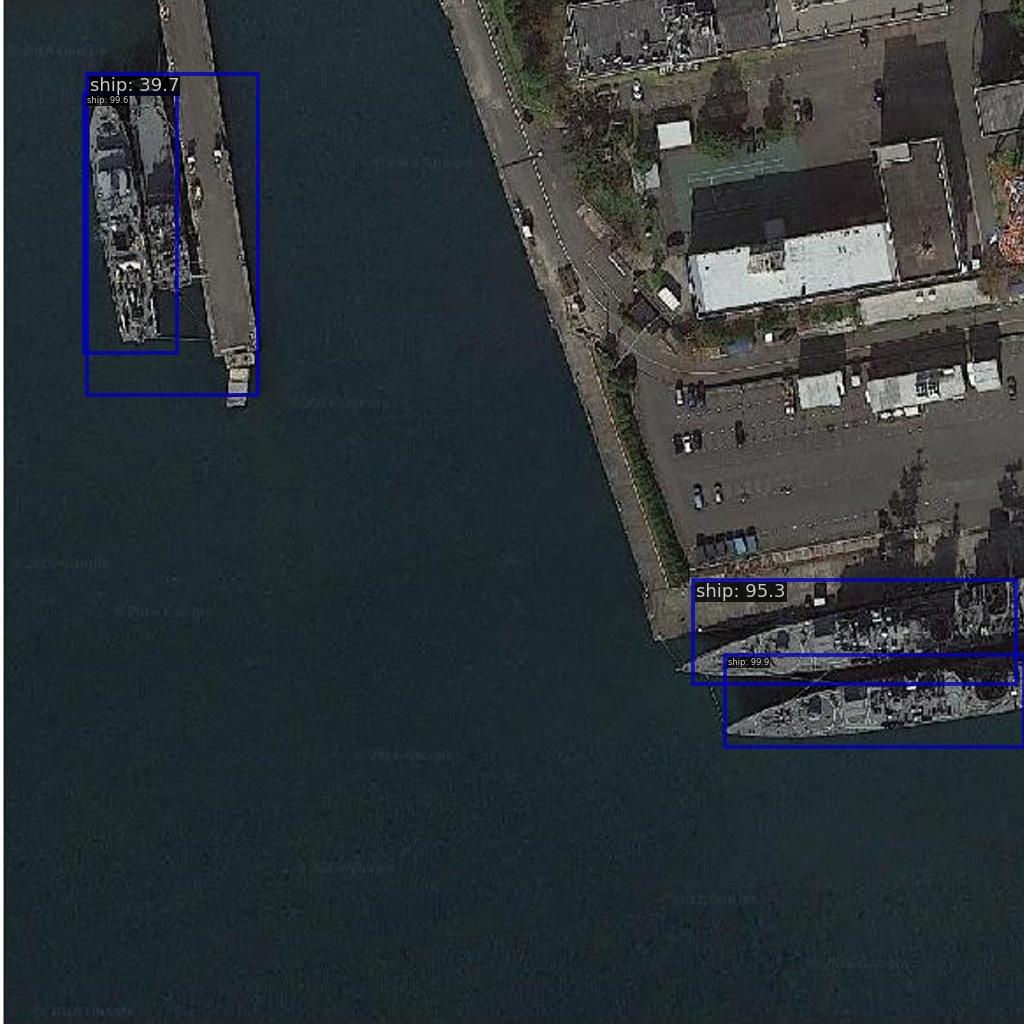}
        & \includegraphics[width=0.125\textwidth]{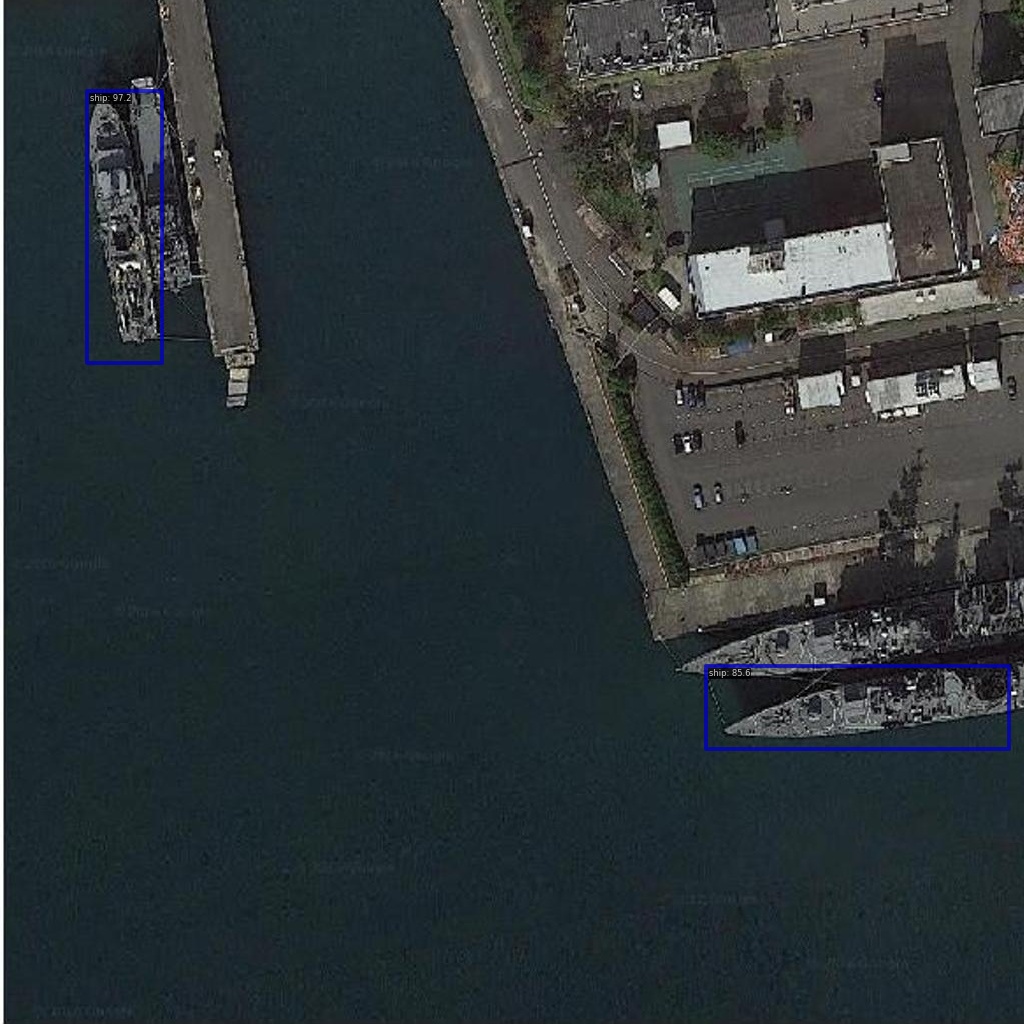}
        & \includegraphics[width=0.125\textwidth]{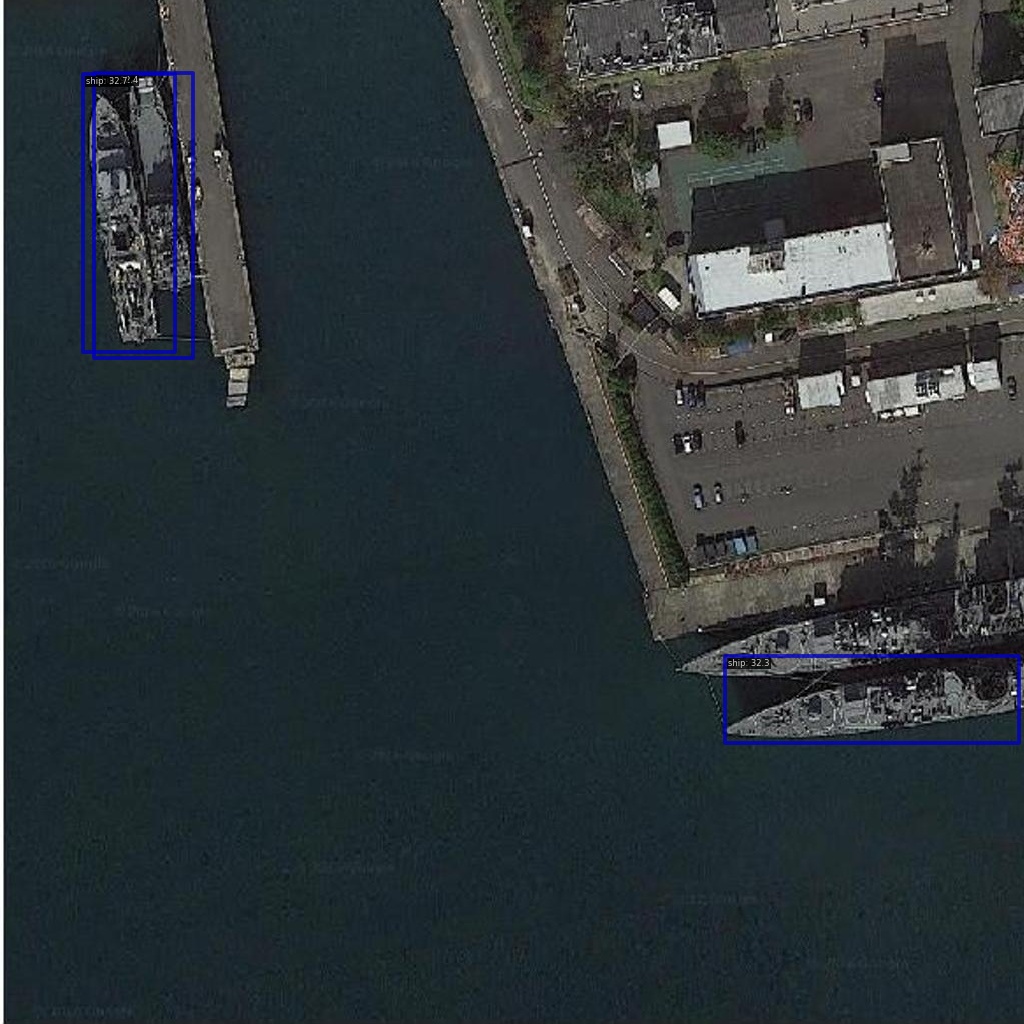}
        & \includegraphics[width=0.125\textwidth]{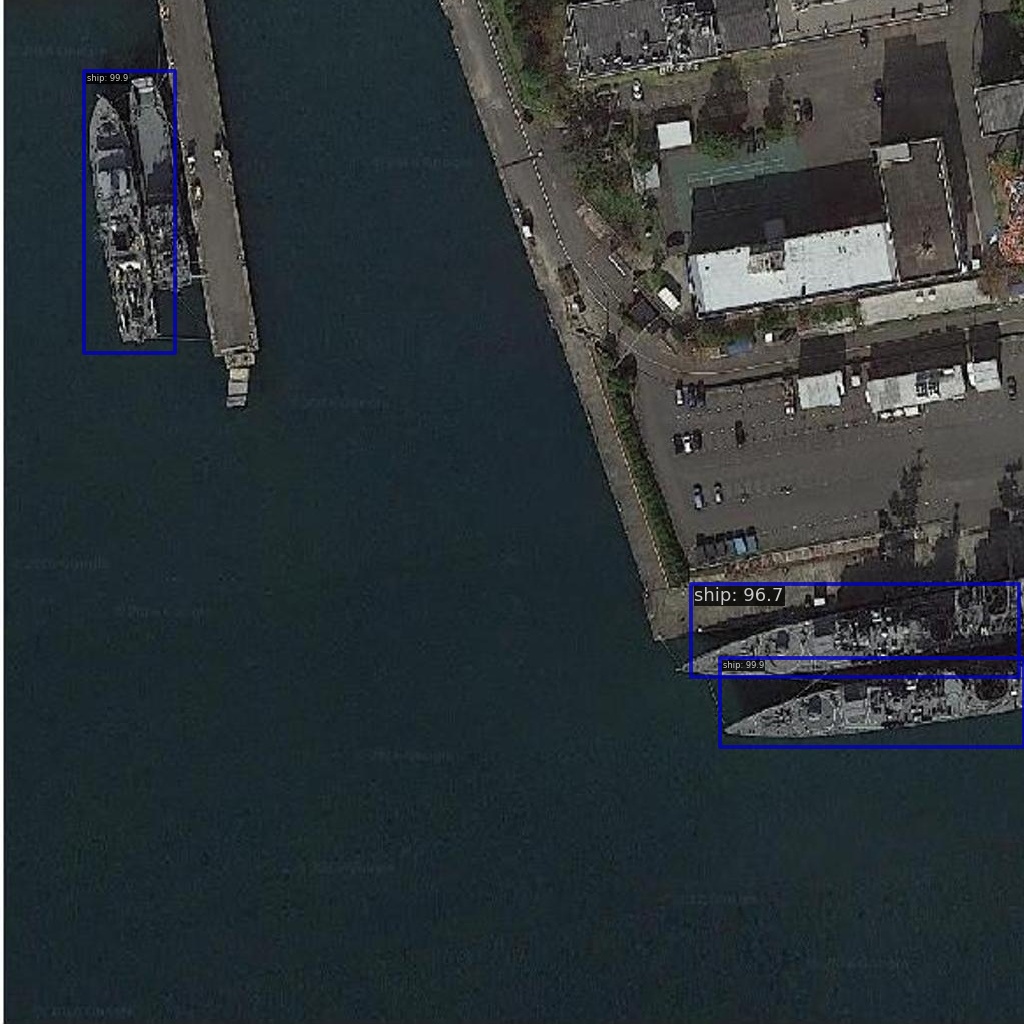}
        & \includegraphics[width=0.125\textwidth]{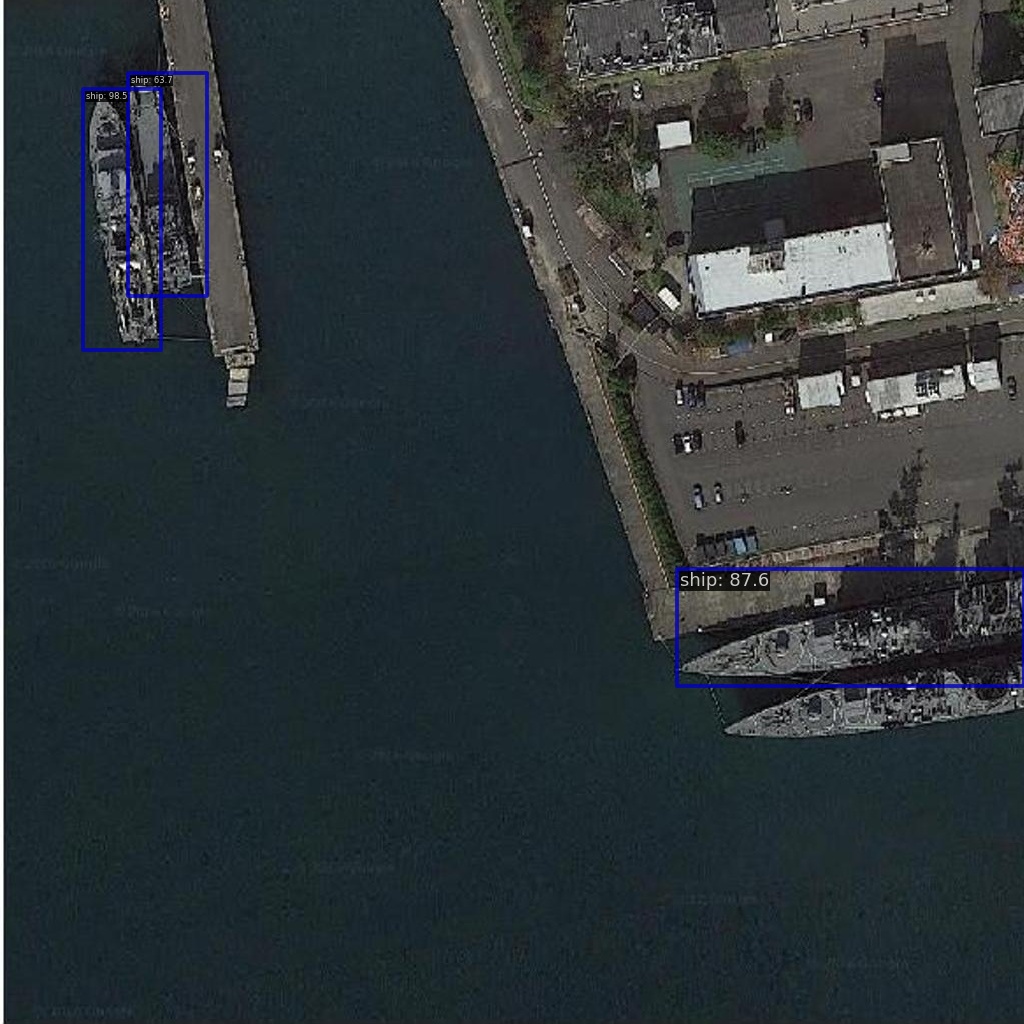}
        & \includegraphics[width=0.125\textwidth]{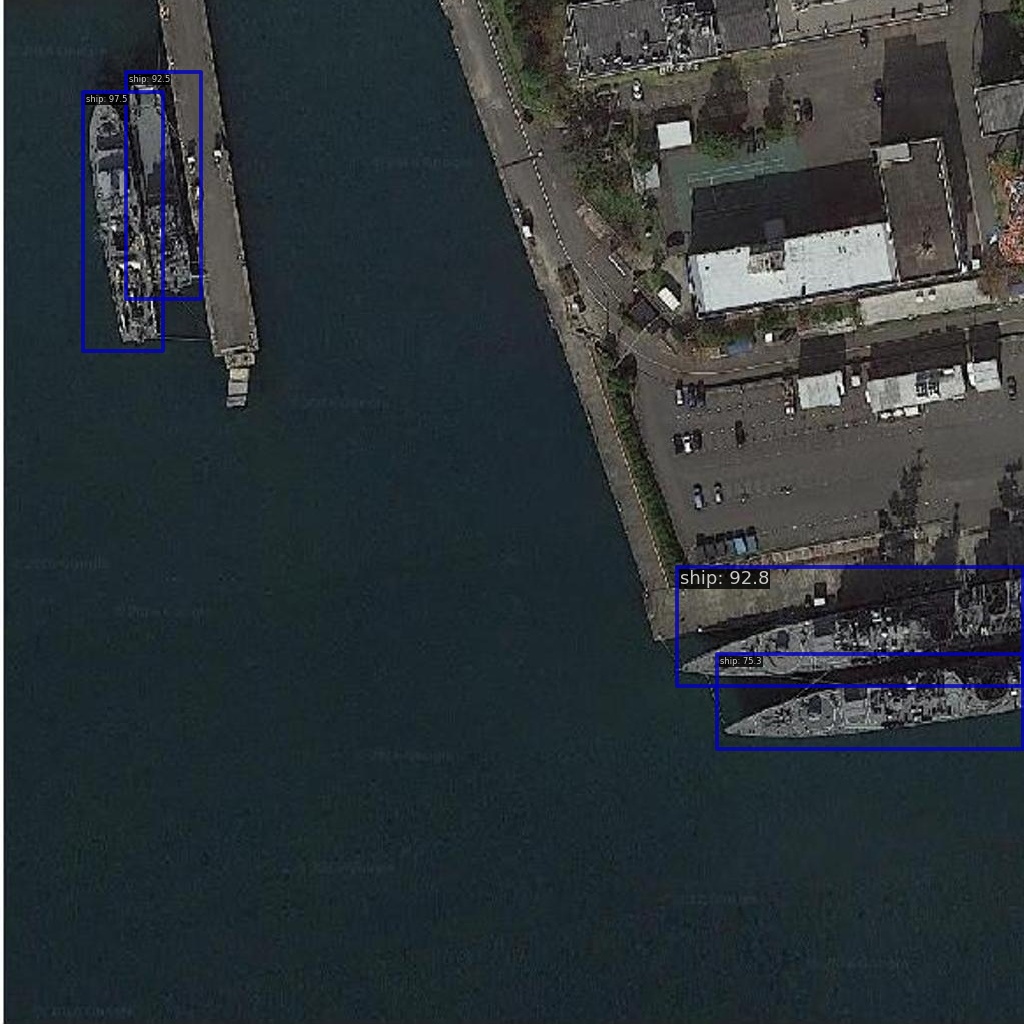} \\
    \end{tabular}
    \caption{\textbf{Qualitative detection results of the proposed CoLR-Det compared with representative detectors.} From left to right, the columns correspond to GT, Faster R-CNN, YOLOX-s, Deformable DETR, MGAM, DINO, and CoLR-Det, where GT denotes ground truth annotations. From top to bottom, the rows show visualization results on the NWPU VHR-10-Split, DOTAv1.5-Split, and HRSSD-Split datasets, respectively.}
    \label{fig:visualization}
\end{figure*}

The visualization results in Fig.~\ref{fig:visualization} are consistent with the quantitative findings. Compared with representative CNN-based, Transformer-based, and specialized tiny-object detectors, CoLR-Det produces fewer missed detections and false positives in dense small-object scenes. On NWPU VHR-10-Split, CoLR-Det detects densely distributed tennis courts and basketball courts more reliably. On DOTAv1.5-Split, it better distinguishes visually similar categories, such as airplanes and helicopters, while recovering small ships and vehicles in port scenes with fewer false alarms. On HRSSD-Split, it provides more accurate localization of shore-side ships and bridges, reducing the category confusion observed in competing detectors. These qualitative results further demonstrate that the proposed saliency-guided multi-task collaborative mechanism effectively suppresses background clutter, improves small-object coverage, and enhances the discrimination of visually similar categories in complex remote sensing scenes.

\subsection{Ablation on the Super-Resolution Branch}

To isolate the contribution of the proposed super-resolution (SR) branch, we conduct controlled ablation studies on NWPU VHR-10-Split~\cite{Cheng2016} and DOTAv1.5-Split~\cite{DOTA2018}. Cascade R-CNN~\cite{CascadeRCNN2018} and DINO~\cite{Zhang2022} are adopted as representative CNN-based and Transformer-based detectors, respectively. Unless otherwise specified, the SR branch is used only during training to regularize the shared encoder and is discarded during inference. The results are reported in Table~\ref{tab:sr_impact}.
\begin{table*}[htbp]
\centering
\caption{Performance comparison of CoLR-Det with representative object detectors on the HRSSD-Split dataset. All detectors are trained and evaluated on low-resolution images obtained by $2\times$ bicubic downsampling of the original high-resolution inputs. The best and second-best results are marked in \textcolor{red}{\textbf{bold}} and \textcolor{blue}{\underline{underlined}}.}
\label{tab:hrssd_comparison}
\resizebox{\textwidth}{!}{%
\begin{tabular}{c|c|c|c|cccccc}
\toprule
Type & Method & Venue & Backbone & $\text{AP}$ & $\text{AP}_{50}$ & $\text{AP}_{75}$ & AP$_s$ & AP$_m$ & AP$_l$ \\
\midrule
\multirow{12}{*}{\makecell{General\\Object\\Detectors}}
 & Faster RCNN~\cite{Ren2015} & TPAMI2017 & Swin-T & 0.631 & 0.891 & 0.737 & 0.228 & 0.537 & 0.681 \\
 & RetinaNet~\cite{FocalLoss2017} & ICCV2017 & Swin-T & 0.638 & 0.895 & 0.734 & 0.251 & 0.548 & 0.664 \\
 & Cascade R-CNN~\cite{CascadeRCNN2018} & CVPR2018 & Swin-T & 0.656 & 0.899 & 0.737 & 0.254 & 0.548 & 0.668 \\
 & YOLOX-s~\cite{YOLOX2021} & arXiv2021 & CSPDarknet & 0.586 & 0.834 & 0.664 & 0.185 & 0.441 & 0.599 \\
 & Sparse R-CNN~\cite{SparseRCNN2023} & TPAMI2023 & ResNet50 & 0.340 & 0.552 & 0.361 & 0.004 & 0.223 & 0.385 \\
 & SR4IR~\cite{Kim2024} & CVPR2024 & MobileNet-V3 & 0.637 & 0.871 & 0.702 & 0.204 & 0.528 & 0.665 \\
 & D-FINE~\cite{peng2025dfine} & ICLR2025 & HGNetv2-B0 & 0.599 & 0.849 & 0.673 & 0.210 & 0.482 & 0.603 \\
 & DAB-DETR~\cite{DABDETR2022} & ICLR2022 & ResNet50 & 0.585 & 0.868 & 0.655 & 0.126 & 0.430 & 0.605 \\
 & DN-DETR~\cite{DNDETR2022} & CVPR2022 & ResNet50 & 0.604 & 0.870 & 0.687 & 0.244 & 0.486 & 0.621 \\
 & Deformable-DETR~\cite{DeformableDETR2020} & ICLR2021 & ResNet50 & 0.601 & 0.868 & 0.688 & 0.180 & 0.514 & 0.600 \\
 & DINO~\cite{Zhang2022} & ICLR2023 & ResNet50 & 0.663 & 0.893 & 0.767 & 0.356 & 0.579 & 0.648 \\
 & DINO~\cite{Zhang2022} & ICLR2023 & Swin-T & \textcolor{blue}{\underline{0.691}} & \textcolor{blue}{\underline{0.926}} & \textcolor{blue}{\underline{0.806}} & 0.315 & 0.591 & \textcolor{red}{\textbf{0.681}} \\
\midrule
\multirow{9}{*}{\makecell{Remote\\Sensing\\Object\\Detectors}}
 & EESRGAN~\cite{Rabbi2020} & Remote Sens. 2020 & ResNet50 & 0.612 & 0.870 & 0.693 & 0.187 & 0.508 & 0.595 \\
 & FSANet~\cite{FSANet2023} & TGRS2022 & Swin-T & 0.562 & 0.876 & 0.635 & 0.053 & 0.485 & 0.562 \\
 & SuperYOLO (RGB)~\cite{SuperYOLO2023} & TGRS2023 & CSPDarknet & 0.581 & 0.883 & 0.694 & 0.263 & 0.494 & 0.587 \\
 & FFCA-YOLO~\cite{FFCAYOLO2024} & TGRS2024 & CSPDarknet & 0.622 & 0.893 & 0.713 & 0.099 & 0.543 & 0.616 \\
 & LEGNet-T~\cite{LEGNet2025} & ICCVW2025 & LEGNet-T & 0.614 & 0.900 & 0.715 & 0.176 & 0.539 & 0.620 \\
 & MGAM~\cite{MGAM2025} & TGRS2025 & ResNet50 & 0.613 & 0.863 & 0.707 & 0.162 & 0.524 & 0.628 \\
 & Strip-RCNN~\cite{yuan2026strip} & AAAI2026 & StripNet-T & 0.610 & 0.902 & 0.694 & 0.205 & 0.510 & 0.600 \\
 \rowcolor{oursbg}& \textbf{CoLR-Det (Ours)} & - & ResNet50 & 0.671 & 0.904 & 0.781 & \textcolor{red}{\textbf{0.391}} & \textcolor{blue}{\underline{0.592}} & 0.650 \\
 \rowcolor{oursbg}& \textbf{CoLR-Det (Ours)} & - & Swin-T & \textcolor{red}{\textbf{0.701}} & \textcolor{red}{\textbf{0.937}} & \textcolor{red}{\textbf{0.820}} & \textcolor{blue}{\underline{0.371}} & \textcolor{red}{\textbf{0.619}} & \textcolor{blue}{\underline{0.680}} \\
\bottomrule
\end{tabular}
}
\end{table*}

\begin{table}[!t]
\centering
\caption{\textbf{Effect of the SR branch on detection performance.} ``SR Branch'' indicates whether the SR branch is enabled.}
\label{tab:sr_impact}
\resizebox{\linewidth}{!}{%
\begin{tabular}{c |c|c| cccccc}
\toprule
Dataset & Method & \makecell{SR\\Branch} & $\text{AP}$ & $\text{AP}_{50}$ & $\text{AP}_{75}$ & AP$_s$ & AP$_m$ & AP$_l$ \\
\midrule
\multirow{4}{*}{\makecell{NWPU\\VHR-10-Split}}
    & \multirow{2}{*}{\makecell{Cascade\\R-CNN}}
    &
    & 0.778 & 0.975 & 0.918 & 0.418 & 0.767 & 0.843 \\
    &
    & $\checkmark$
    & 0.789 & 0.977 & 0.926 & \textbf{0.452} & 0.783 & 0.860 \\
\cmidrule(l){2-9}
    & \multirow{2}{*}{DINO}
    &
    & 0.766 & 0.986 & 0.911 & 0.487 & 0.744 & 0.830 \\
    &
    & $\checkmark$
    & 0.816 & 0.990 & 0.939 & \textbf{0.552} & 0.794 & 0.886 \\
\midrule
\multirow{4}{*}{\makecell{DOTA\\v1.5-Split}}
    & \multirow{2}{*}{\makecell{Cascade\\R-CNN}}
    &
    & 0.357 & 0.537 & 0.391 & 0.142 & 0.396 & 0.513 \\
    &
    & $\checkmark$
    & 0.362 & 0.541 & 0.404 & \textbf{0.183} & 0.410 & 0.525 \\
\cmidrule(l){2-9}
    & \multirow{2}{*}{DINO}
    &
    & 0.402 & 0.659 & 0.427 & 0.192 & 0.434 & 0.540 \\
    &
    & $\checkmark$
    & 0.412 & 0.663 & 0.439 & \textbf{0.221} & 0.445 & 0.560 \\
\bottomrule
\end{tabular}
}
\end{table}

\begin{table}[htbp]
\centering
\caption{Effect of training and inference configurations of the SR branch on detection performance.}
\label{tab:sr_disentangle}
\resizebox{\columnwidth}{!}{%
\begin{tabular}{l|ccc|ccc}
\toprule
Method & \makecell{Train\\SR} & \makecell{Encoder\\Grad} & \makecell{Infer\\SR} & $\text{AP}$ & $\text{AP}_{50}$ & AP$_s$ \\
\midrule
DINO (Baseline) & $\times$ & -- & $\times$ & 0.766 & 0.986 & 0.487 \\
DINO-SR (Detached) & $\checkmark$ & $\times$ & $\times$ & 0.770 & 0.987 & 0.491 \\
\textbf{DINO-SR Branch} & $\checkmark$ & $\checkmark$ & $\times$ & \textbf{0.816} & \textbf{0.990} & \textbf{0.552} \\
DINO-SR (Infer) & $\checkmark$ & $\checkmark$ & $\checkmark$ & \underline{0.815} & \textbf{0.990} & \underline{0.551} \\
\bottomrule
\end{tabular}
}
\end{table}

As shown in Table~\ref{tab:sr_impact}, enabling the SR branch consistently improves detection performance across both datasets and detector families. The improvement is particularly pronounced for small objects, with AP$_s$ gains of 0.034 and 0.065 on NWPU VHR-10-Split, and 0.041 and 0.029 on DOTAv1.5-Split for Cascade R-CNN and DINO, respectively. These results indicate that the SR branch provides an effective auxiliary training signal, encouraging the shared encoder to preserve fine-grained spatial details in LR remote sensing imagery while maintaining stable performance on medium and large objects.

To further identify the source of the improvement, we conduct two additional controlled experiments on NWPU VHR-10-Split using DINO as the baseline. In the detached variant, the SR branch is trained with reconstruction supervision, but its gradients are blocked from updating the shared encoder. In the inference variant, the model follows the default training procedure, but the SR head is additionally executed during inference. The results are summarized in Table~\ref{tab:sr_disentangle}.

The detached variant performs comparably to the DINO baseline, indicating that an isolated reconstruction head does not provide a consistent detection benefit when its gradients are blocked from the shared encoder. In contrast, allowing reconstruction supervision to update the shared encoder improves AP from 0.766 to 0.816 and AP$_s$ from 0.487 to 0.552. Moreover, additionally executing the SR head during inference produces essentially unchanged detection performance. These results indicate that the benefit of the restoration branch mainly arises from training-time regularization of the shared representation, while explicit SR reconstruction is unnecessary during inference.

\begin{table}[!t]
\centering
\caption{Comparison of saliency-driven and baseline configurations for object detection performance.}
\label{tab:saliency_impact}
\resizebox{\columnwidth}{!}{%
\begin{tabular}{c|cc|cccccc}
\toprule
Method & Saliency & \makecell{SR\\Branch} & $\text{AP}$ & $\text{AP}_{50}$ & $\text{AP}_{75}$ & AP$_s$ & AP$_m$ & AP$_l$ \\
\midrule
DINO (baseline) & & & 0.766 & 0.986 & 0.911 & 0.487 & 0.744 & 0.830 \\
DINO-Saliency & $\checkmark$ & & 0.776 & \textbf{0.991} & 0.918 & 0.474 & 0.759 & 0.841 \\
DINO-SR Branch & & $\checkmark$ & \underline{0.816} & \underline{0.990} & \underline{0.939} & \underline{0.552} & \underline{0.794} & \underline{0.886} \\
\textbf{CoLR-Det (Ours)} & $\checkmark$ & $\checkmark$ & \textbf{0.836} & \textbf{0.991} & \textbf{0.954} & \textbf{0.649} & \textbf{0.816} & \textbf{0.890} \\
\bottomrule
\end{tabular}
}
\end{table}

\begin{table}[!t]
\centering
\caption{Ablation on the training strategy with different Stage-One durations $T_{det}$ on the NWPU VHR-10-Split dataset.}
\label{tab:training_strategy}
\resizebox{\columnwidth}{!}{%
\begin{tabular}{l|cccccc}
\toprule
Setting & $\text{AP}$ & $\text{AP}_{50}$ & $\text{AP}_{75}$ & AP$_s$ & AP$_m$ & AP$_l$ \\
\midrule
Single-stage (EW) & 0.811 & \textbf{0.992} & 0.945 & 0.603 & 0.789 & 0.870 \\
Single-stage (UW) & 0.800 & \underline{0.991} & 0.949 & 0.586 & 0.780 & 0.858 \\
\midrule
$T_{det}=8$ & 0.815 & \underline{0.991} & 0.947 & 0.593 & 0.791 & 0.862 \\
$T_{det}=12$ & \underline{0.830} & \textbf{0.992} & \textbf{0.959} & 0.621 & \underline{0.810} & \underline{0.878} \\
\textbf{$T_{det}=16$} & \textbf{0.836} & \underline{0.991} & 0.954 & \underline{0.649} & \textbf{0.816} & \textbf{0.890} \\
$T_{det}=20$ & 0.819 & \textbf{0.992} & 0.956 & \textbf{0.662} & 0.804 & 0.876 \\
$T_{det}=24$ & 0.829 & \textbf{0.992} & \underline{0.957} & 0.586 & \underline{0.810} & 0.870 \\
\bottomrule
\end{tabular}
}
\end{table}

\begin{figure}[!t]
\centering
\includegraphics[width=\linewidth]{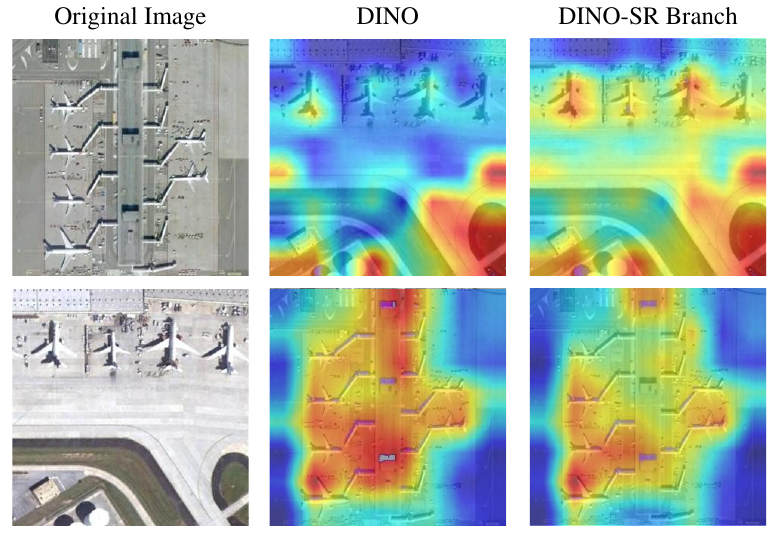}
\caption{ \textbf{Visualization of the third-stage backbone feature responses of DINO and DINO-SR Branch on NWPU VHR-10-Split~\cite{Cheng2016}.} From left to right, the columns show the original images and the corresponding feature responses produced by DINO and DINO-SR Branch, respectively. Each row represents a different example. }
\label{fig:feature_vis}
\end{figure}

We examine the representational effect of the SR branch by visualizing intermediate feature maps from the four stages of the shared encoder, as shown in Fig.~\ref{fig:feature_vis}. Compared with the DINO baseline, the SR-augmented model produces clearer foreground--background separation in shallow layers and more concentrated object-aware activations in deeper layers. In particular, the baseline often exhibits diffuse responses around cluttered airport regions, whereas the SR-augmented model focuses more precisely on small targets such as airplanes while suppressing irrelevant background responses. These observations are consistent with the quantitative results and suggest that the SR branch serves as a training-time representation regularizer: it injects auxiliary reconstruction structural cues into the shared encoder and converts them into more discriminative semantic features for small-object localization.

\begin{figure}[!t]
    \centering
    \includegraphics[width=\linewidth]{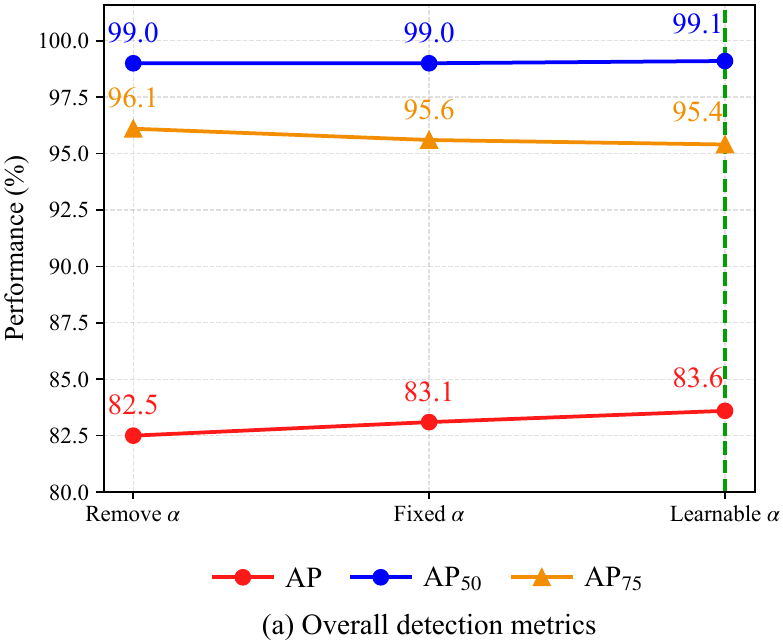}
    \vspace{3mm}
    \includegraphics[width=\linewidth]{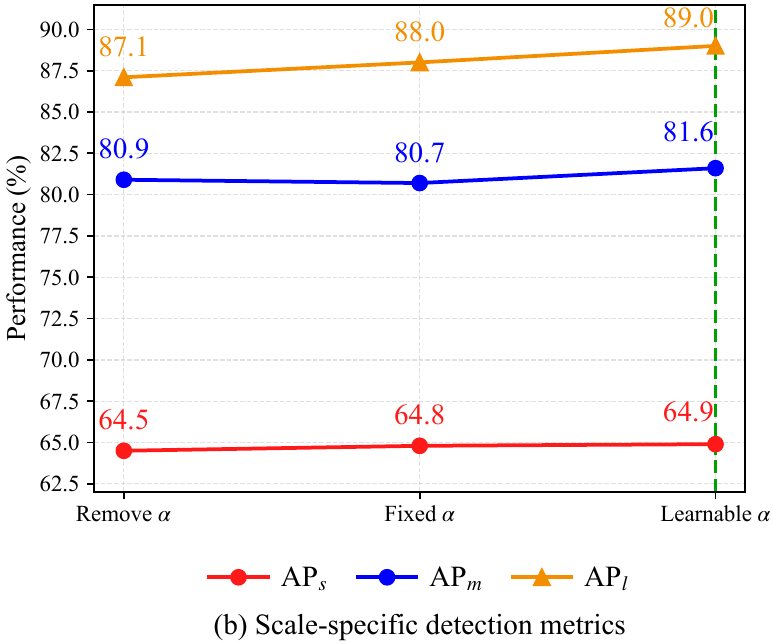}
\caption{Comparison of different fusion strategies for the modulation coefficient $\alpha$. The green dashed line indicates the learnable-$\alpha$ setting.} 
    \label{fig:fusion_strategy}
\end{figure}

\subsection{Ablation on Saliency-Guided Token Routing}  
We further investigate the contribution of saliency-guided token routing to detection performance. Table~\ref{tab:saliency_impact} compares four configurations on NWPU VHR-10-Split: the DINO baseline, DINO equipped with saliency-guided token routing (DINO-Saliency), DINO coupled with the latent restoration branch (DINO-SR Branch), and the complete CoLR-Det framework. Compared with the baseline, DINO-Saliency improves the overall AP from 0.766 to 0.776, but slightly decreases AP$_s$ from 0.487 to 0.474. This result suggests that saliency-based computation allocation alone is not uniformly beneficial to small objects, because weak object responses may occasionally receive relatively low saliency rankings. Nevertheless, these low-ranked tokens remain in the feature stream through the non-destructive bypass path rather than being permanently discarded.

A different trend is observed when latent restoration supervision is introduced. The DINO-SR Branch improves AP and AP$_s$ to 0.816 and 0.552, respectively, while incorporating saliency-guided routing further increases them to 0.836 and 0.649. The resulting gains of 2.0 percentage points in AP and 9.7 percentage points in AP$_s$ demonstrate a clear complementarity between latent restoration and saliency-guided routing. Restoration supervision helps preserve fine-grained spatial evidence, whereas saliency-guided routing prioritizes high-saliency encoder tokens for attention-based refinement and reduces the influence of restoration-enhanced background responses. Their complementary effects enable CoLR-Det to better exploit restoration cues and substantially improve the detection of small objects under low-resolution observations.

\subsection{Analysis of Dynamic Modulation Mechanism}

\begin{figure*}[!t] 
\centering 
\includegraphics[width=\textwidth]{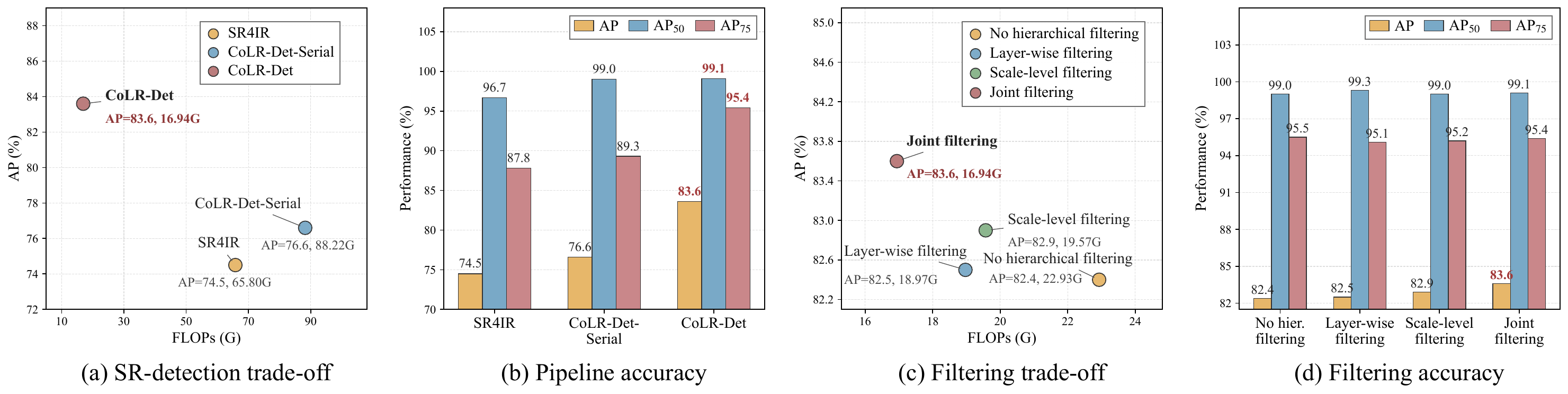} 
\caption{
\textbf{Accuracy--efficiency analysis of CoLR-Det on the NWPU VHR-10-Split dataset.}
(a) AP--FLOPs trade-off between CoLR-Det and representative two-stage SR--detection pipelines.
(b) Comparison of AP, AP$_{50}$, and AP$_{75}$ among different SR--detection pipelines.
(c) AP--FLOPs trade-off under different query-token routing strategies.
(d) Comparison of AP, AP$_{50}$, and AP$_{75}$ for different query-token routing strategies.
}
\label{fig:accuracy} 
\vspace{-12pt}
\end{figure*}

The modulation coefficient $\alpha$ (defined in Eq.~\ref{eq:saliency_propagation}) governs the intensity of saliency propagation. To determine the optimal fusion strategy, we compare three settings: removing $\alpha$ (equivalent to $\alpha=1$), fixing $\alpha=0.5$, and treating $\alpha$ as a learnable parameter. The results in Fig.~\ref{fig:fusion_strategy} clearly favor the learnable approach. The learnable $\alpha$ setting achieves the highest AP of 0.836, outperforming the fixed constant setting (0.831) and the removal setting (0.825).

This advantage comes from the ability of the learnable parameter to adaptively regulate the injection of high-level semantic priors. Unlike a fixed coefficient, which may cause insufficient background suppression or excessive smoothing of local details, a learnable $\alpha$ adjusts the strength of top-down modulation and supports effective integration between coarse semantic context and fine spatial details.

\subsection{Ablation on Training Strategy}
\label{sec:ablation_training}

To evaluate the effectiveness of the proposed detection-prioritized two-stage optimization strategy, we compare it with two single-stage joint-training baselines on the NWPU VHR-10-Split dataset. The first baseline, Single-stage (EW), jointly optimizes the SR and detection branches from the beginning using fixed equal task weights. The second baseline, Single-stage (UW), adopts the homoscedastic uncertainty weighting strategy of Kendall~\textit{et al.}~\cite{kendall2018multi} to adaptively balance the SR and detection losses. All architectural components and hyper-parameters are kept unchanged to isolate the effect of the training schedule. We also vary the Stage-One duration $T_{det} \in \{8, 12, 16, 20, 24\}$ to assess the sensitivity of the proposed strategy. The results are shown in Table~\ref{tab:training_strategy}.

As shown in Table~\ref{tab:training_strategy}, the proposed two-stage strategy generally outperforms the single-stage alternatives, especially in overall AP and small-object detection. Although the single-stage baselines achieve comparable $\text{AP}_{50}$, their lower $\text{AP}_{75}$ and $\text{AP}_{s}$ indicate weaker localization precision and less effective small-object representation. This suggests that introducing SR supervision from the beginning may disturb the formation of detection-oriented semantic features. The uncertainty-weighted baseline also fails to provide a clear advantage over equal weighting, implying that loss reweighting alone is insufficient when the SR and detection objectives are not yet well aligned.

The results with different $T_{det}$ values further show the importance of properly scheduling the two tasks. A short detection-only stage introduces SR gradients before the encoder has learned stable semantic representations, leading to limited improvement. Extending Stage One generally improves performance, but an overly long detection-only phase reduces the benefit of subsequent SR-guided refinement. Among all settings, $T_{det}=16$ achieves the best overall balance across AP, small-object detection, and large-object detection, and is therefore adopted as the default configuration.

\subsection{Computational Efficiency Analysis}

\label{subsec:model_efficiency}

\subsubsection{Comparison with Two-Stage SR--Detection Pipelines}

A critical limitation of traditional ``SR--detection'' serial pipelines is the objective misalignment, where super-resolution reconstruction may conflict with detection-oriented semantics. As demonstrated in Fig.~\ref{fig:accuracy}(a)-(b), we compare CoLR-Det against a state-of-the-art serial method (SR4IR~\cite{Kim2024}) and a baseline serial implementation (CoLR-Det-Serial). Specifically, CoLR-Det-Serial adopts the identical Swin-T backbone, SR decoder, and DINO detection head as CoLR-Det, but arranges them in a sequential SR-then-detection pipeline trained under the same hyperparameters. The serial approach suffers from high computational redundancy (88.22G FLOPs) and suboptimal accuracy (0.766 AP). This is mainly because the serial design performs costly image-level SR reconstruction before detection, introducing redundant feature decoding while optimizing pixel fidelity rather than preserving detection-discriminative semantics. In contrast, our parallel, unified framework reduces the computational load to 16.94G FLOPs while achieving a superior AP of 0.836. This confirms that collaborative optimization in a shared feature space is far more effective than sequential processing for LR inputs.

\subsubsection{Effect of Hierarchical Query Token Selection on Efficiency}

Finally, we analyze the effect of the hierarchical query token selection strategy on inference efficiency. Fig.~\ref{fig:accuracy}(c)-(d) reports an ablation in which tokens are routed at different stages. Retaining all tokens (``No hierarchical routing'') leads to high computational overhead without producing the best accuracy (0.824 AP), because redundant background tokens dilute attention. With joint layer-wise and scale-level routing, FLOPs are reduced to 16.94G while AP increases to 0.836. The joint mechanism acts as a saliency-guided routing mechanism that routes approximately 60\% of background-dominated queries through the bypass path according to saliency rankings. This concentrates computational resources on regions with high object probability and achieves a favorable balance between detection precision and model efficiency.

\section{CONCLUSION}
\label{sec:conclusion}

In this paper, we propose CoLR-Det, a saliency-driven collaborative framework for improving small object detection in low-resolution remote sensing imagery. By moving beyond the conventional serial paradigm, CoLR-Det mitigates objective misalignment and spatial attention diffusion in multi-task pipelines. The proposed method enables semantically consistent interaction between super-resolution and detection through implicit feature sharing, and the Detection-Prioritized Two-Stage Optimization Strategy balances reconstruction and detection supervision more effectively. In addition, the saliency-driven mechanism helps the model focus on critical object regions and reduce the influence of background-dominated responses on detection-oriented token refinement. Experimental results on multiple datasets show that CoLR-Det significantly outperforms state-of-the-art methods. Although this work narrows the gap between pixel-level reconstruction and semantic discrimination, future research will explore finer-grained bidirectional guidance mechanisms and adaptive task-weighting schemes to further improve multi-task synergy in complex Earth observation scenarios.

\ifCLASSOPTIONcaptionsoff
  \newpage
\fi
\vspace{-7pt}
\bibliographystyle{IEEEtran}
\bibliography{references}

@article{Nikouei2025,
  author  = {M. Nikouei and B. Baroutian and S. Nabavi and F. Taraghi and A. Aghaei and A. Sajedi and M. E. Moghaddam},
  title   = {Small object detection: A comprehensive survey on challenges, techniques and real-world applications},
  journal = {Intell. Syst. Appl.},
  volume  = {27},
  month   = {Sep.},
  year    = {2025},
  note    = {{Art. no.} 200561}
}

@article{Gui2024,
  author  = {S. Gui and S. Song and R. Qin and Y. Tang},
  title   = {Remote sensing object detection in the deep learning era---A review},
  journal = {Remote Sens.},
  volume  = {16},
  number  = {2},
  month   = {Jan.},
  year    = {2024},
  note    = {{Art. no.} 327}
}

@article{Su2025,
  author  = {H. Su and Y. Li and Y. Xu and X. Fu and S. Liu},
  title   = {A review of deep-learning-based super-resolution: From methods to applications},
  journal = {Pattern Recognit.},
  volume  = {157},
  month   = {Jan.},
  year    = {2025},
  note    = {{Art. no.} 110935}
}

@article{Lepcha2023,
  author  = {D. C. Lepcha and B. Goyal and A. Dogra and V. Goyal},
  title   = {Image super-resolution: A comprehensive review, recent trends, challenges and applications},
  journal = {Inf. Fusion},
  volume  = {91},
  pages   = {230--260},
  month   = {Mar.},
  year    = {2023}
}

@article{Wang2025,
  author  = {Y. Wang and Z. Shao and T. Lu and X. Huang and J. Wang and Z. Zhang and X. Zuo},
  title   = {Lightweight remote sensing super-resolution with multi-scale graph attention network},
  journal = {Pattern Recognit.},
  volume  = {160},
  month   = {Apr.},
  year    = {2025},
  note    = {{Art. no.} 111178}
}

@article{Alsaedi2025,
  author  = {M. Asif and M. Abrar and F. Ullah and A. Salam and F. Amin and I. de la Torre and M. G. Villar and H. Garay and G. S. Choi},
  title   = {A novel hybrid deep learning approach for super-resolution and objects detection in remote sensing},
  journal = {Sci. Rep.},
  volume  = {15},
  number  = {1},
  month   = {May},
  year    = {2025},
  note    = {{Art. no.} 17221}
}

@article{Cheng2016,
  author  = {G. Cheng and P. Zhou and J. Han},
  title   = {Learning rotation-invariant convolutional neural networks for object detection in {VHR} optical remote sensing images},
  journal = {IEEE Trans. Geosci. Remote Sens.},
  volume  = {54},
  number  = {12},
  pages   = {7405--7415},
  month   = {Dec.},
  year    = {2016}
}

@inproceedings{DOTA2018,
  author    = {G.-S. Xia and X. Bai and J. Ding and Z. Zhu and S. Belongie and J. Luo and M. Datcu and M. Pelillo and L. Zhang},
  title     = {{DOTA}: A large-scale dataset for object detection in aerial images},
  booktitle = {Proc. IEEE/CVF Conf. Comput. Vis. Pattern Recognit. (CVPR)},
  month     = {Jun.},
  year      = {2018},
  pages     = {3974--3983}
}

@article{Zhang2019,
  author  = {Y. Zhang and Y. Yuan and Y. Feng and X. Lu},
  title   = {Hierarchical and robust convolutional neural network for very high-resolution remote sensing object detection},
  journal = {IEEE Trans. Geosci. Remote Sens.},
  volume  = {57},
  number  = {8},
  pages   = {5535--5548},
  month   = {Aug.},
  year    = {2019}
}

@article{Dong2016,
  author  = {C. Dong and C. C. Loy and K. He and X. Tang},
  title   = {Image super-resolution using deep convolutional networks},
  journal = {IEEE Trans. Pattern Anal. Mach. Intell.},
  volume  = {38},
  number  = {2},
  pages   = {295--307},
  month   = {Feb.},
  year    = {2016}
}

@article{Lei2017,
  author  = {S. Lei and Z. Shi and Z. Zou},
  title   = {Super-resolution for remote sensing images via local--global combined network},
  journal = {IEEE Geosci. Remote Sens. Lett.},
  volume  = {14},
  number  = {8},
  pages   = {1243--1247},
  month   = {Aug.},
  year    = {2017}
}

@article{Jiang2018a,
  author  = {K. Jiang and Z. Wang and P. Yi and J. Jiang and J. Xiao and Y. Yao},
  title   = {Deep distillation recursive network for remote sensing imagery super-resolution},
  journal = {Remote Sens.},
  volume  = {10},
  number  = {11},
  month   = {Oct.},
  year    = {2018},
  note    = {{Art. no.} 1700}
}

@article{Jiang2019,
  author  = {K. Jiang and Z. Wang and P. Yi and G. Wang and T. Lu and J. Jiang},
  title   = {Edge-enhanced {GAN} for remote sensing image superresolution},
  journal = {IEEE Trans. Geosci. Remote Sens.},
  volume  = {57},
  number  = {8},
  pages   = {5799--5812},
  month   = {Aug.},
  year    = {2019}
}

@article{Xiao2023,
  author  = {Y. Xiao and Q. Yuan and K. Jiang and J. He and Y. Wang and L. Zhang},
  title   = {From degrade to upgrade: Learning a self-supervised degradation guided adaptive network for blind remote sensing image super-resolution},
  journal = {Inf. Fusion},
  volume  = {96},
  pages   = {297--311},
  month   = {Aug.},
  year    = {2023}
}

@inproceedings{Zhang2018,
  author    = {Y. Zhang and K. Li and K. Li and L. Wang and B. Zhong and Y. Fu},
  title     = {Image super-resolution using very deep residual channel attention networks},
  booktitle = {Proc. Eur. Conf. Comput. Vis. (ECCV)},
  month     = {Sep.},
  year      = {2018},
  pages     = {286--301}
}

@inproceedings{Niu2020,
  author    = {B. Niu and W. Wen and W. Ren and X. Zhang and L. Yang and S. Wang and K. Zhang and X. Cao and H. Shen},
  title     = {Single image super-resolution via a holistic attention network},
  booktitle = {Proc. Eur. Conf. Comput. Vis. (ECCV)},
  month     = {Aug.},
  year      = {2020},
  pages     = {191--207}
}

@article{Lei2022,
  author  = {S. Lei and Z. Shi},
  title   = {Hybrid-scale self-similarity exploitation for remote sensing image super-resolution},
  journal = {IEEE Trans. Geosci. Remote Sens.},
  volume  = {60},
  pages   = {1-10},
  month   = {Apr.},
  year    = {2022}
}

@inproceedings{Mei2021,
  author    = {Y. Mei and Y. Fan and Y. Zhou},
  title     = {Image super-resolution with non-local sparse attention},
  booktitle = {Proc. IEEE/CVF Conf. Comput. Vis. Pattern Recognit. (CVPR)},
  month     = {Jun.},
  year      = {2021},
  pages     = {3517--3526}
}

@article{Kim2023,
  author  = {B. Kim and J. Kim and J. C. Ye},
  title   = {Task-agnostic vision transformer for distributed learning of image processing},
  journal = {IEEE Trans. Image Process.},
  volume  = {32},
  pages   = {203--218},
  month   = {Dec.},
  year    = {2022}
}

@article{He2022,
  author  = {J. He and Q. Yuan and J. Li and Y. Xiao and X. Liu and Y. Zou},
  title   = {{DsTer}: A dense spectral transformer for remote sensing spectral super-resolution},
  journal = {Int. J. Appl. Earth Obs. Geoinf.},
  volume  = {109},
  month   = {May},
  year    = {2022},
  note    = {{Art. no.} 102773}
}

@article{Xiao2023b,
  author  = {Y. Xiao and Q. Yuan and K. Jiang and X. Jin and J. He and L. Zhang and C.-W. Lin},
  title   = {Local-global temporal difference learning for satellite video super-resolution},
  journal = {IEEE Trans. Circuits Syst. Video Technol.},
  volume  = {34},
  number  = {4},
  pages   = {2789--2802},
  month   = {Apr.},
  year    = {2024}
}

@inproceedings{Liang2021,
  author    = {J. Liang and J. Cao and G. Sun and K. Zhang and L. {Van Gool} and R. Timofte},
  title     = {{SwinIR}: Image restoration using {Swin Transformer}},
  booktitle = {Proc. IEEE/CVF Int. Conf. Comput. Vis. Workshops (ICCVW)},
  month     = {Oct.},
  year      = {2021},
  pages     = {1833--1844}
}

@article{Xiao2024,
  author  = {Y. Xiao and Q. Yuan and K. Jiang and J. He and C.-W. Lin and L. Zhang},
  title   = {{TTST}: A top-k token selective transformer for remote sensing image super-resolution},
  journal = {IEEE Trans. Image Process.},
  volume  = {33},
  pages   = {738--752},
  month   = {Jan.},
  year    = {2024}
}

@article{Kang2024,
  author  = {X. Kang and P. Duan and J. Li and S. Li},
  title   = {Efficient {Swin Transformer} for remote sensing image super-resolution},
  journal = {IEEE Trans. Image Process.},
  volume  = {33},
  pages   = {6367--6379},
  month   = {Nov.},
  year    = {2024}
}

@article{Ren2015,
  author  = {S. Ren and K. He and R. Girshick and J. Sun},
  title   = {Faster {R-CNN}: Towards real-time object detection with region proposal networks},
  journal = {IEEE Trans. Pattern Anal. Mach. Intell.},
  volume  = {39},
  number  = {6},
  pages   = {1137--1149},
  month   = {Jun.},
  year    = {2017}
}

@inproceedings{Redmon2017,
  author    = {J. Redmon and A. Farhadi},
  title     = {{YOLO9000}: Better, faster, stronger},
  booktitle = {Proc. IEEE Conf. Comput. Vis. Pattern Recognit. (CVPR)},
  month     = {Jul.},
  year      = {2017},
  pages     = {7263--7271}
}

@article{MSOPN2019,
  author  = {Z. Deng and H. Sun and S. Zhou and J. Zhao and L. Lei and H. Zou},
  title   = {Multi-scale object detection in remote sensing imagery with convolutional neural networks},
  journal = {ISPRS J. Photogramm. Remote Sens.},
  volume  = {145},
  pages   = {3--22},
  month   = {Nov.},
  year    = {2018}
}

@article{FSANet2023,
  author  = {J. Wu and Z. Pan and B. Lei and Y. Hu},
  title   = {{FSANet}: Feature-and-spatial-aligned network for tiny object detection in remote sensing images},
  journal = {IEEE Trans. Geosci. Remote Sens.},
  volume  = {60},
  month   = {Sep.},
  year    = {2022},
  note    = {{Art. no.} 5630717}
}

@article{FECenterNet2023,
  author  = {T. Shi and J. Gong and J. Hu and X. Zhi and W. Zhang and Y. Zhang and P. Zhang and G. Bao},
  title   = {Feature-enhanced {CenterNet} for small object detection in remote sensing images},
  journal = {Remote Sens.},
  volume  = {14},
  number  = {21},
  month   = {Oct.},
  year    = {2022},
  note    = {{Art. no.} 5488}
}

@article{FFCAYOLO2024,
  author  = {Y. Zhang and M. Ye and G. Zhu and Y. Liu and P. Guo and J. Yan},
  title   = {{FFCA-YOLO} for small object detection in remote sensing images},
  journal = {IEEE Trans. Geosci. Remote Sens.},
  volume  = {62},
  month   = {Feb.},
  year    = {2024},
  note    = {{Art. no.} 5611215}
}

@article{LSODNet2024,
  author  = {K. Jin and W. Du and M. Tang and W. Liang and K. Li and A.-S. K. Pathan},
  title   = {{LSODNet}: A lightweight and efficient detector for small object detection in remote sensing images},
  journal = {IEEE J. Sel. Top. Appl. Earth Obs. Remote Sens.},
  volume  = {18},
  pages   = {24816--24828},
  month   = {Sep.},
  year    = {2025}
}

@inproceedings{LEGNet2025,
  author    = {W. Lu and S.-B. Chen and H.-D. Li and Q.-L. Shu and C. H. Q. Ding and J. Tang and B. Luo},
  title     = {{LEGNet}: A lightweight edge-{Gaussian} network for low-quality remote sensing image object detection},
  booktitle = {Proc. IEEE/CVF Int. Conf. Comput. Vis. Workshops (ICCVW)},
  month     = {Oct.},
  year      = {2025},
  pages     = {2844--2853}
}

@article{PRDETR2024,
  author  = {Y. Chen and B. Liu and L. Yuan},
  title   = {{PR-Deformable DETR}: {DETR} for remote sensing object detection},
  journal = {IEEE Geosci. Remote Sens. Lett.},
  volume  = {21},
  month   = {May},
  year    = {2024},
  note    = {{Art. no.} 2506105}
}

@article{DroneDETR2024,
  author  = {Y. Kong and X. Shang and S. Jia},
  title   = {{Drone-DETR}: Efficient small object detection for remote sensing image using enhanced {RT-DETR} model},
  journal = {Sensors},
  volume  = {24},
  number  = {17},
  month   = {Aug.},
  year    = {2024},
  note    = {{Art. no.} 5496}
}

@inproceedings{Shermeyer2019,
  author    = {J. Shermeyer and A. {Van Etten}},
  title     = {The effects of super-resolution on object detection performance in satellite imagery},
  booktitle = {Proc. IEEE/CVF Conf. Comput. Vis. Pattern Recognit. Workshops (CVPRW)},
  month     = {Jun.},
  year      = {2019},
  pages     = {1432--1441}
}

@article{Ji2019,
  author  = {H. Ji and Z. Gao and T. Mei and B. Ramesh},
  title   = {Vehicle detection in remote sensing images leveraging on simultaneous super-resolution},
  journal = {IEEE Geosci. Remote Sens. Lett.},
  volume  = {17},
  number  = {4},
  pages   = {676--680},
  month   = {Apr.},
  year    = {2020}
}

@article{Rabbi2020,
  author  = {J. Rabbi and N. Ray and M. Schubert and S. Chowdhury and D. Chao},
  title   = {Small-object detection in remote sensing images with end-to-end edge-enhanced {GAN} and object detector network},
  journal = {Remote Sens.},
  volume  = {12},
  number  = {9},
  month   = {Apr.},
  year    = {2020},
  note    = {{Art. no.} 1432}
}

@inproceedings{Kim2024,
  author    = {J. Kim and J. Oh and K. M. Lee},
  title     = {Beyond image super-resolution for image recognition with task-driven perceptual loss},
  booktitle = {Proc. IEEE/CVF Conf. Comput. Vis. Pattern Recognit. (CVPR)},
  month     = {Jun.},
  year      = {2024},
  pages     = {2651--2661}
}

@article{Wu2021,
  author  = {J. Wu and S. Xu},
  title   = {From point to region: Accurate and efficient hierarchical small object detection in low-resolution remote sensing images},
  journal = {Remote Sens.},
  volume  = {13},
  number  = {13},
  month   = {Jul.},
  year    = {2021},
  note    = {{Art. no.} 2620}
}

@inproceedings{Liu2021,
  author    = {Z. Liu and Y. Lin and Y. Cao and H. Hu and Y. Wei and Z. Zhang and S. Lin and B. Guo},
  title     = {{Swin Transformer}: Hierarchical vision transformer using shifted windows},
  booktitle = {Proc. IEEE/CVF Int. Conf. Comput. Vis. (ICCV)},
  month     = {Oct.},
  year      = {2021},
  pages     = {10012--10022}
}

@inproceedings{UNet2015,
  author    = {O. Ronneberger and P. Fischer and T. Brox},
  title     = {{U-Net}: Convolutional networks for biomedical image segmentation},
  booktitle = {Proc. Int. Conf. Med. Image Comput. Comput.-Assist. Interv. (MICCAI)},
  month     = {Oct.},
  year      = {2015},
  pages     = {234--241}
}

@inproceedings{FocalLoss2017,
  author    = {T.-Y. Lin and P. Goyal and R. Girshick and K. He and P. Doll{\'a}r},
  title     = {Focal loss for dense object detection},
  booktitle = {Proc. IEEE Int. Conf. Comput. Vis. (ICCV)},
  month     = {Oct.},
  year      = {2017},
  pages     = {2980--2988}
}

@inproceedings{CascadeRCNN2018,
  author    = {Z. Cai and N. Vasconcelos},
  title     = {Cascade {R-CNN}: Delving into high quality object detection},
  booktitle = {Proc. IEEE/CVF Conf. Comput. Vis. Pattern Recognit. (CVPR)},
  month     = {Jun.},
  year      = {2018},
  pages     = {6154--6162}
}

@misc{YOLOX2021,
  author       = {Z. Ge and S. Liu and F. Wang and Z. Li and J. Sun},
  title        = {{YOLOX}: Exceeding {YOLO} series in 2021},
  howpublished = {arXiv:2107.08430},
  month        = {Jul.},
  year         = {2021}
}

@article{SparseRCNN2023,
  author  = {P. Sun and R. Zhang and Y. Jiang and T. Kong and C. Xu and W. Zhan and M. Tomizuka and Z. Yuan and P. Luo},
  title   = {Sparse {R-CNN}: An end-to-end framework for object detection},
  journal = {IEEE Trans. Pattern Anal. Mach. Intell.},
  volume  = {45},
  number  = {12},
  pages   = {15650--15664},
  month   = {Dec.},
  year    = {2023}
}

@inproceedings{peng2025dfine,
  author    = {Y. Peng and H. Li and P. Wu and Y. Zhang and X. Sun and F. Wu},
  title     = {{D-FINE}: Redefine regression task of {DETRs} as fine-grained distribution refinement},
  booktitle = {Proc. Int. Conf. Learn. Represent. (ICLR)},
  month     = {Apr.},
  year      = {2025}
}

@inproceedings{DABDETR2022,
  author    = {S. Liu and F. Li and H. Zhang and X. Yang and X. Qi and H. Su and J. Zhu and L. Zhang},
  title     = {{DAB-DETR}: Dynamic anchor boxes are better queries for {DETR}},
  booktitle = {Proc. Int. Conf. Learn. Represent. (ICLR)},
  month     = {Apr.},
  year      = {2022}
}

@inproceedings{DNDETR2022,
  author    = {F. Li and H. Zhang and S. Liu and J. Guo and L. M. Ni and L. Zhang},
  title     = {{DN-DETR}: Accelerate {DETR} training by introducing query denoising},
  booktitle = {Proc. IEEE/CVF Conf. Comput. Vis. Pattern Recognit. (CVPR)},
  month     = {Jun.},
  year      = {2022},
  pages     = {13619--13627}
}

@inproceedings{Zhang2022,
  author    = {H. Zhang and F. Li and S. Liu and L. Zhang and H. Su and J. Zhu and L. M. Ni and H.-Y. Shum},
  title     = {{DINO}: {DETR} with improved denoising anchor boxes for end-to-end object detection},
  booktitle = {Proc. Int. Conf. Learn. Represent. (ICLR)},
  month     = {May},
  year      = {2023}
}

@inproceedings{DeformableDETR2020,
  author    = {X. Zhu and W. Su and L. Lu and B. Li and X. Wang and J. Dai},
  title     = {Deformable {DETR}: Deformable transformers for end-to-end object detection},
  booktitle = {Proc. Int. Conf. Learn. Represent. (ICLR)},
  month     = {May},
  year      = {2021}
}

@article{SuperYOLO2023,
  author  = {J. Zhang and J. Lei and W. Xie and Z. Fang and Y. Li and Q. Du},
  title   = {{SuperYOLO}: Super resolution assisted object detection in multimodal remote sensing imagery},
  journal = {IEEE Trans. Geosci. Remote Sens.},
  volume  = {61},
  pages   = {1-15},
  month   = {Mar.},
  year    = {2023}
}

@article{MGAM2025,
  author  = {S. Shi and Q. Fang and X. Xu and D. Dong},
  title   = {Multiscale {Gaussian} attention mechanism for tiny-object detection in remote sensing images},
  journal = {IEEE Trans. Geosci. Remote Sens.},
  volume  = {63},
  month   = {Jan.},
  year    = {2025},
  note    = {{Art. no.} 5635216}
}

@article{yuan2026strip,
  author  = {X. Yuan and Z. Zheng and Y. Li and X. Liu and L. Liu and X. Li and Q. Hou and M.-M. Cheng},
  title   = {Strip {R-CNN}: Large strip convolution for remote sensing object detection},
  journal = {Proc. AAAI Conf. Artif. Intell.},
  volume  = {40},
  number  = {15},
  pages   = {12259--12267},
  month   = {Jan.},
  year    = {2026}
}

@inproceedings{kendall2018multi,
  author    = {A. Kendall and Y. Gal and R. Cipolla},
  title     = {Multi-task learning using uncertainty to weigh losses for scene geometry and semantics},
  booktitle = {Proc. IEEE/CVF Conf. Comput. Vis. Pattern Recognit. (CVPR)},
  month     = {Jun.},
  year      = {2018},
  pages     = {7482--7491}
}

\raggedbottom
\makeatletter
\def\@IEEEBIOskipN{1.0\baselineskip} 
\makeatother

\begin{IEEEbiography}
[{\includegraphics[width=1in,height=1.25in,clip,keepaspectratio]{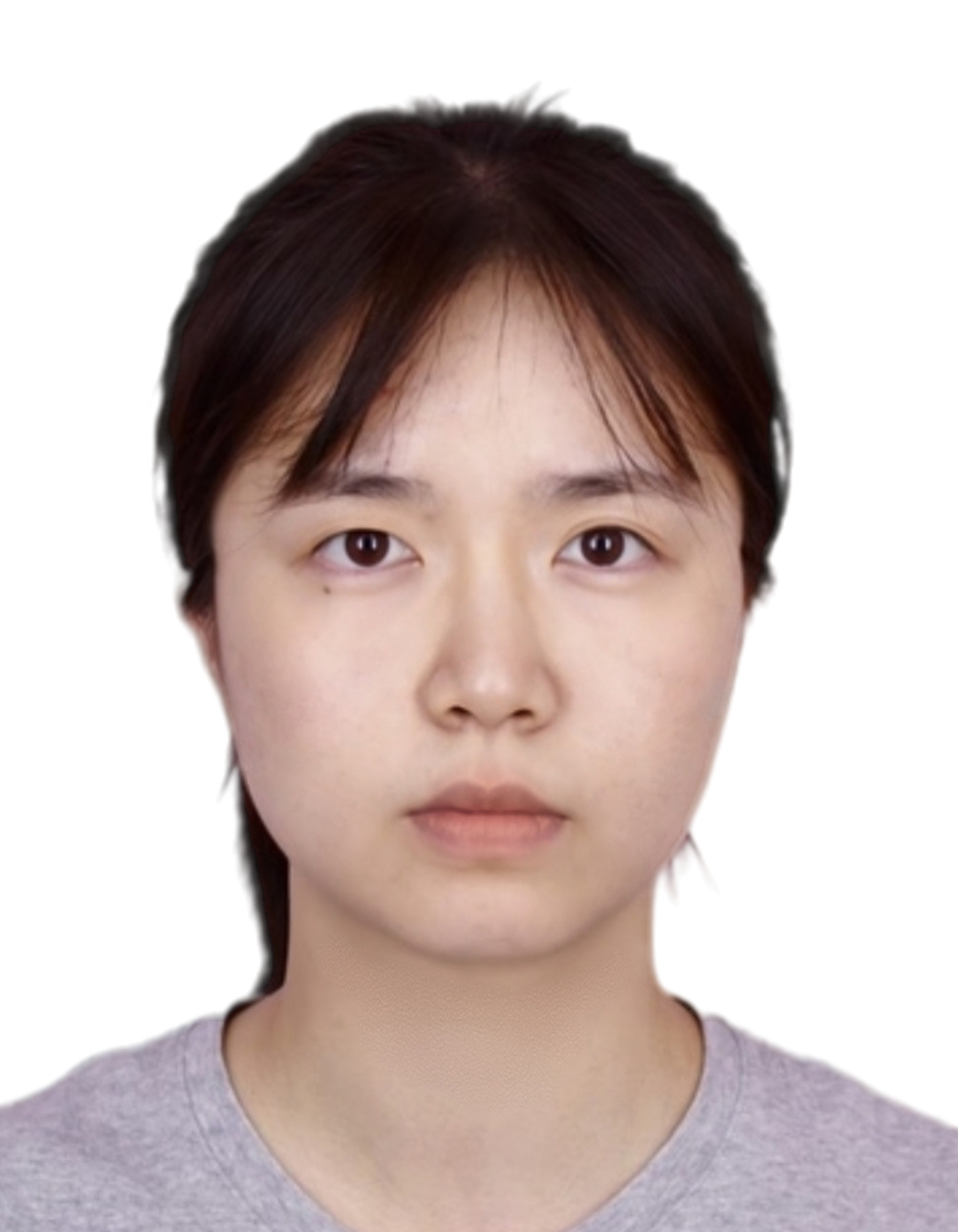}}]{Ruo Qi}
received the M.E. degree in electronic information from Harbin University of Science and Technology, Harbin, China, in 2023. She is currently pursuing the D.Eng. degree in electronic information with Shenzhen University, Shenzhen, China. Her research interests include remote sensing image processing and object detection.
\end{IEEEbiography}

\begin{IEEEbiography}
[{\includegraphics[width=1in,height=1.25in,clip,keepaspectratio]{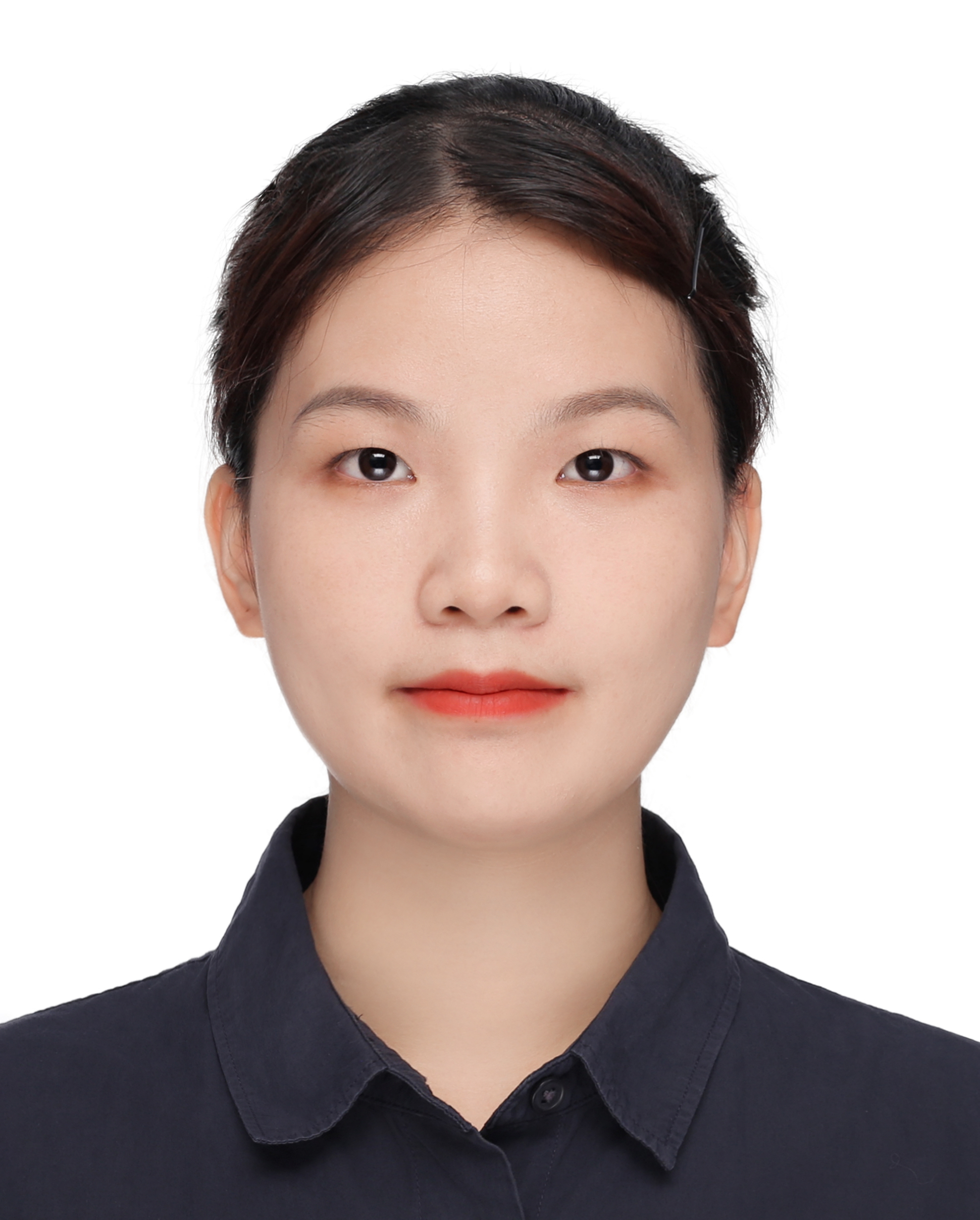}}]{Linhui Dai}(Corresponding author) 
received the Ph.D. degree from the School of Computer Science, Peking University (PKU), China, in 2024, under the supervision of Prof. Hong Liu. She currently serves as an Assistant Professor at the College of Electronics and Information Engineering, Shenzhen University, Shenzhen, China. She has authored or coauthored papers in TMM, PR, T-CSVT, CAAI TRIT, and others. Her research interests include open world object detection, underwater object detection, and salient object detection.
\end{IEEEbiography}

\begin{IEEEbiography}
[{\includegraphics[width=1in,height=1.25in,clip,keepaspectratio]{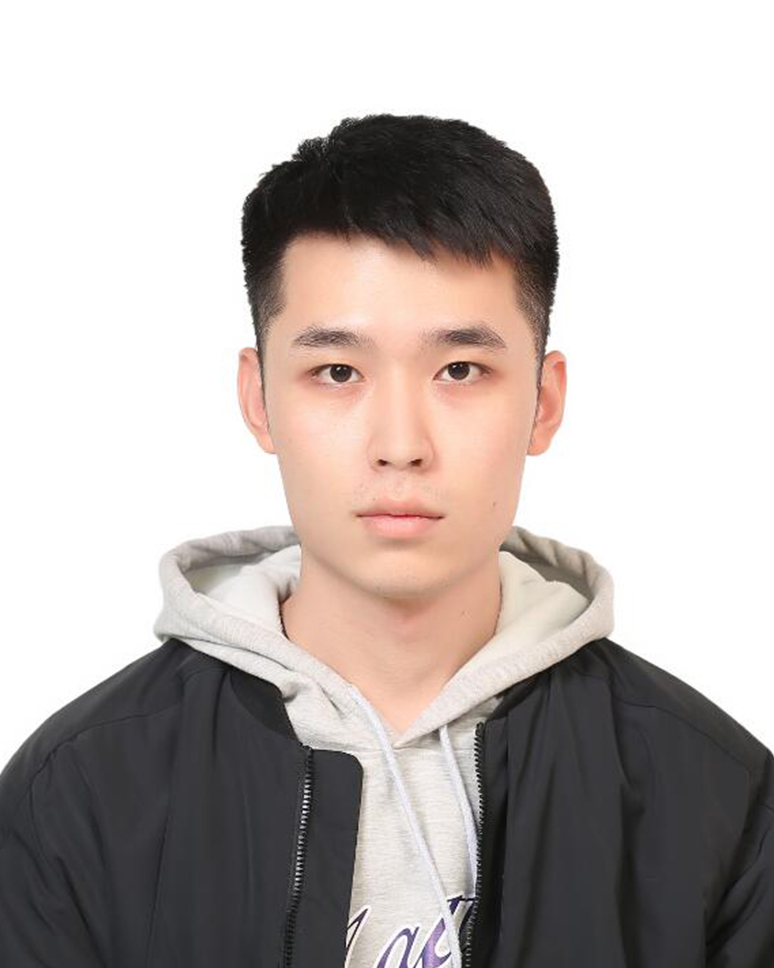}}]{Yusong Qin}
is currently pursuing the Ph.D. degree in information and communication engineering with the College of Electronic and Information Engineering, Shenzhen University, Shenzhen, China. His research interests include image processing, computer vision, and explainability of deep networks.
\end{IEEEbiography}

\begin{IEEEbiography}
[{\includegraphics[width=1in,height=1.25in,clip,keepaspectratio]{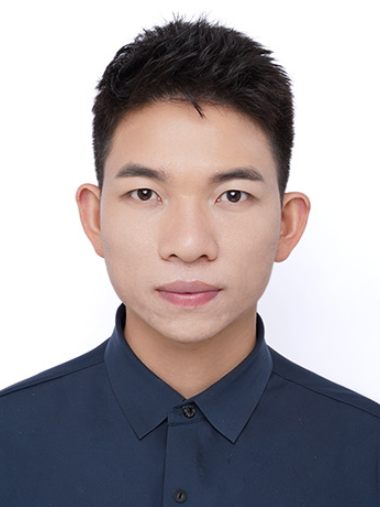}}]{Chaolei Yang}
is currently pursuing the Ph.D. degree in information and communication engineering with the College of Electronic and Information Engineering, Shenzhen University, Shenzhen, China. His research interests include image/video processing, computer vision, and person ReID.
\end{IEEEbiography}

\begin{IEEEbiography}
[{\includegraphics[width=1in,height=1.25in,clip,keepaspectratio]{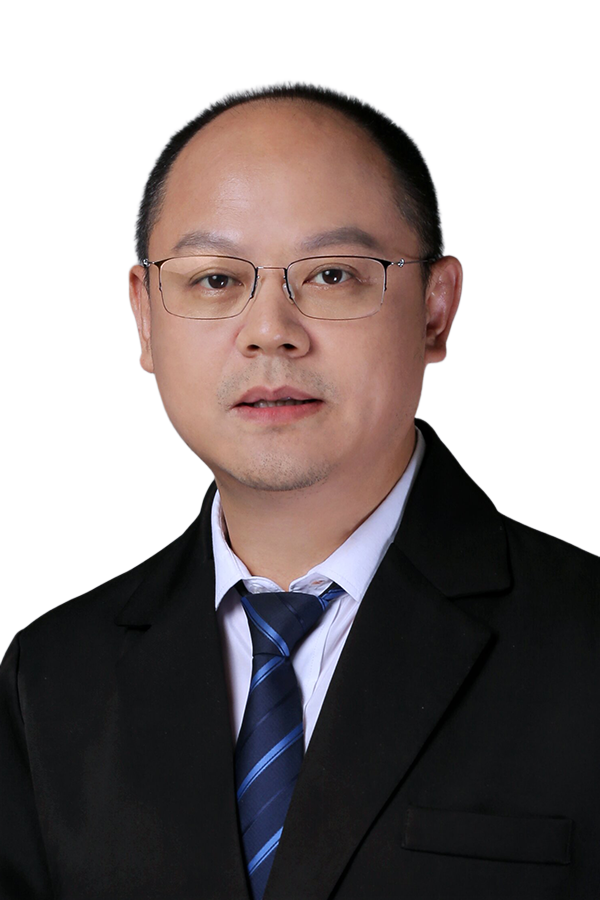}}]{Yanshan Li}
received the Ph.D. degree from South China University of Technology. He is currently a Researcher and Doctoral Supervisor with the Institute of Intelligent Information Processing and Guangdong Key Laboratory of Intelligent Information Processing, Shenzhen University. His research interests include computer vision, machine learning, and image analysis.
\end{IEEEbiography}

\end{document}